\documentclass[10pt,journal,compsoc]{IEEEtran}
\usepackage{tabularx}
\usepackage{indentfirst}
\ifCLASSOPTIONcompsoc
  \usepackage[nocompress]{cite}
\else
  \usepackage{cite}
\fi
\ifCLASSINFOpdf
\else
\fi

\usepackage{epsfig}
\usepackage{graphicx}
\usepackage{amsmath}
\usepackage{amssymb}
\usepackage{enumitem}
\usepackage{tikz}
\usetikzlibrary{shapes.geometric, arrows.meta, shapes.arrows, positioning}

\usepackage[breaklinks=true,colorlinks,bookmarks=false,urlcolor=red,citecolor=cyan]{hyperref}

\usepackage{gensymb,graphics,subfigure,inputenc}
\usepackage[linesnumbered,algo2e,boxed]{algorithm2e}
\graphicspath{{Figure/}}
\usepackage[figuresright]{rotating}
\usepackage{amsmath,amssymb,textcomp,subfigure,multirow,upgreek}
\usepackage{booktabs}
\usepackage{color,mdwlist}

\usepackage{todonotes}

\usepackage{float}

\usepackage{etoolbox}

\usepackage{ragged2e}

\usepackage{review}
\usepackage{longtable}

\usepackage{graphicx,fontawesome}
\graphicspath{{Figures/}}

\usepackage{stfloats}
\usepackage{multicol}
\usepackage{caption}
\usepackage{wrapfig}
\usepackage{booktabs}
\usepackage{enumitem}
\setitemize{noitemsep,topsep=0pt,parsep=0pt,partopsep=0pt}

\begin{document}
\title{Multimodal Large Language Models for Medicine: A Comprehensive Survey}
\author{Jiarui Ye  \quad Hao Tang$^*$ 
	\IEEEcompsocitemizethanks{
   \IEEEcompsocthanksitem Jiarui Ye is with the School of Computer Science and Engineering, Nanjing University of Science and Technology, Nanjing 210094, China. E-mail: jiaruiye@njust.edu.cn \protect
   \IEEEcompsocthanksitem Hao Tang is with the School of Computer Science, Peking University, Beijing 100871, China. E-mail: haotang@pku.edu.cn \protect
        }% <-this % stops an unwanted space
\thanks{$^*$Corresponding author: Hao Tang.}
}


\markboth{IEEE Transactions on Pattern Analysis and Machine Intelligence}%
{Shell \MakeLowercase{\textit{et al.}}: Bare Demo of IEEEtran.cls for Computer Society Journals}



\IEEEtitleabstractindextext{%
%\begin{abstract}
%The abstract goes here.
%\end{abstract}
\justify
\begin{abstract}
% MLLMs have recently become a focal point in the field of artificial intelligence research. Based on the strong ability of LLMs, MLLMs are adept at addressing complex multi-modal tasks. With the release of gpt-4, MLLMs have gained substantial attention from different domains. Researchers has begun to explore the potential of the MLLMs in the medical healthcare domain. In this paper, we first introduce the background and the fundamental concepts related to LMMs and MLLMs, while emphasizing the working principle of MLLMs. Subsequently, we summarize three main directions of application within medical healthcare: medical reporting, medical diagnosis, and medical treatment. We illustrate the remarkable capabilities of MLLMs in these domains by providing specific exam1ples.   For data, We point out six mainstream modes of data along with their corresponding evaluation benchmark. In the end of the survey, we discuss the challenges that faced by MLLMs in medical healthcare domain and propose the feasible methods to mitigate or overcome these issues.    

MLLMs have recently become a focal point in the field of artificial intelligence research. Building on the strong capabilities of LLMs, MLLMs are adept at addressing complex multi-modal tasks. With the release of GPT-4, MLLMs have gained substantial attention from different domains. Researchers have begun to explore the potential of MLLMs in the medical and healthcare domain. In this paper, we first introduce the background and fundamental concepts related to LLMs and MLLMs, while emphasizing the working principles of MLLMs. Subsequently, we summarize three main directions of application within healthcare: medical reporting, medical diagnosis, and medical treatment.
Our findings are based on a comprehensive review of 330 recent papers in this area. 
We illustrate the remarkable capabilities of MLLMs in these domains by providing specific examples. For data, we present six mainstream modes of data along with their corresponding evaluation benchmarks. At the end of the survey, we discuss the challenges faced by MLLMs in the medical and healthcare domain and propose feasible methods to mitigate or overcome these issues.

\end{abstract}

}

\maketitle

\IEEEdisplaynontitleabstractindextext

% \fi
\IEEEpeerreviewmaketitle


\section{Introduction}

Language models play an important role in NLP (natural language processing) tasks. By comprehending and generating text, these models are able to perform various language-related tasks, such as text extraction, sentiment analysis, and so on. In the development of language models, Transformer, which was published by Google in 2017, is a significant milestone\cite{1}. It is a deep learning architecture that relies on the self-attention mechanism, thus enhancing efficiency through parallel computing. The model assigns different attention weights to different parts of the input, thereby enhancing its ability to grasp the meaning of the text. With the release of Transformer, the scale and number of parameters in the models have gradually expanded.

This marks the beginning of the era of LLMs. A series of LLMs have been introduced. Among these models, BERT\cite{2}, which is based on Transformer, remarkably grasps the meaning of the context through pre-training tasks such as masked language modeling and next sentence prediction tasks. Furthermore, several open-source LLMs such as Flan-T5\cite{3}, Vicuna\cite{4}, and LLaMA\cite{5} have made substantial progress and contributed to the field of LLMs.

In the field of medical healthcare, LLMs have been used to improve the quality of medical work. LLMs have played a crucial role in supporting specific applications, such as generating brief and accurate reports based on EHR, progress notes, doctor-patient conversations, and other medical text formats. Although the medical domain exists with multiple data modalities such as text, images, videos, audio, and omics, using these types of data efficiently and combining them in appropriate ways has become a trend in addressing complex healthcare tasks.

Recently, MLLMs, which are based on LLMs and are capable of solving multimodal tasks, have entered the public eye. Most current MLLMs have similar structures. They put LLMs at their core, while incorporating an encoder and a diffusion generative model at the input and output, respectively. Some modules designed for handling multimodal tasks have been progressively refined, such as CLIP\cite{6}, BLIP\cite{7}, BLIP-2\cite{8}, and Flamingo\cite{9}, which is used for few-shot learning. To deal with the various modalities of medical data, MLLMs have been introduced into the field of medicine. For example, MLLMs have mainly been utilized to address medical image-text tasks, such as generating diagnostic reports based on textual knowledge and image data, such as CT scans.

\newcommand{\inserttimeline}[7]{
    \node[timelabelstyle] (timeLabel) at (#2) {\textbf{#1}};
    \node[inner sep=0pt, right=0.05cm of timeLabel.east] (stick) {\includegraphics[width=0.05cm, height=0.7cm]{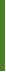}};
    \node[modelstyle] (modelName) at ([xshift=0.4cm, yshift=-0.1cm] timeLabel.south) {#3};
    \node[inner sep=0pt, below=0.08cm of modelName, xshift=#6, yshift=#7] (modelImage) {\includegraphics[width=#5]{images/figure2/#4}};
}

\newcommand{\insertangle}[5]{
    \node[inner sep=0pt] (angleImage) at (#1) {\includegraphics[width=#4, height=#5]{images/figure2/#2}};
    \node[align=center, textstyle, right=1pt, yshift=-1pt] at (angleImage.east) {#3};
}

\newcommand{\insertarrow}[4]{
    \node[arrowstyle] (#1) at (#2) {\includegraphics[width=3.5cm]{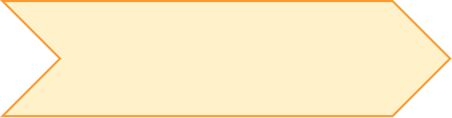}};
    \node[labelstyle] at (#1.center) {#3};
    \node[descstyle] at (#1.north) {#4};
}

\newcommand{\insertarrowsecond}[4]{
    \node[arrowstyle1] (#1) at (#2) {\includegraphics[width=7.5cm]{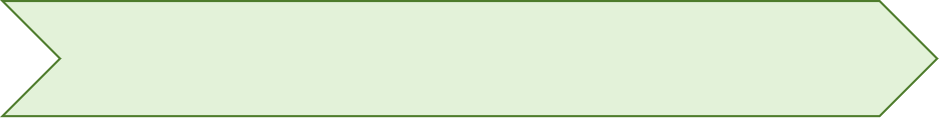}};
    \node[labelstyle] at (#1.center) {#3};
    \node[descstyle] at (#1.north) {#4};
}

\newcommand{\insertmedrepresent}[7]{
    \node[timelabelstyle] (timeLabelmed) at (#2) {\textbf{#1}};
    \node[inner sep=0pt, below=0.1cm of timeLabelmed, xshift=#6, yshift=#7] (modelmedImage) {\includegraphics[width=#5]{images/figure2/#4}};
    \node[modelstyle] (modelmedName) at ([xshift=0cm, yshift=-0.1cm] modelmedImage.south) {#3};
}

\begin{figure*}[htbp]
\centering
\begin{tikzpicture}
\tikzset{
        arrowstyle/.style={inner sep=0pt, anchor=center},
        arrowstyle1/.style={inner sep=0pt, anchor=center},
        labelstyle/.style={align=center, font=\bfseries\small, text=black, anchor=center},
        descstyle/.style={align=center, font=\fontsize{7pt}{10pt}\selectfont, text=black, anchor=south, xshift=-5pt, yshift=1pt},
        timelabelstyle/.style={align=center, font=\fontsize{7.6pt}{1.2pt}\selectfont, text=black, anchor=center},
        modelstyle/.style={align=center, font=\fontsize{7pt}{7pt}\selectfont, text=black, anchor=north},
        textstyle/.style={anchor=west, align=left, font=\small,anchor=center}
    }

    \node[inner sep=0pt](growtharrow) at (9.8cm, 1.8cm) {\includegraphics[width=0.85cm]{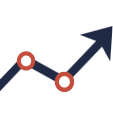}};
    \node[align=center, font=\bfseries\itshape\fontsize{9pt}{10pt}\selectfont, text=black, anchor=east] at ([xshift=-5pt, yshift=-5pt] growtharrow.west) {massive parameter growth};
    \node[inner sep=0pt] (bluearrow) at (8cm, 1.2cm) {\includegraphics[width=4cm]{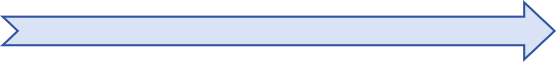}};
    \pgfdeclarelayer{background} 
    \pgfsetlayers{background,main} 
    
    \begin{pgfonlayer}{background} 
        \node[inner sep=0pt](penetratearrow) at (6.8cm, -2.35cm) {\includegraphics[width=18cm]{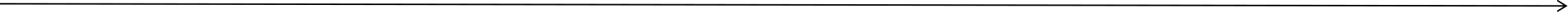}};
    \node[inner sep=0pt](angle1) at (9.55cm, -3.1cm) {\includegraphics[width=1cm,height= 0.5cm]{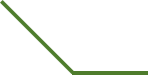}};
    \node[anchor=north west, xshift=1pt, yshift=-1pt,font=\scriptsize] at (angle1.east) {Inception of LLMs};
    \node[inner sep=0pt](angle2) at (12.55cm, -3.1cm) {\includegraphics[width=1cm,height= 0.5cm]{images/figure2/angle.png}};
    \node[anchor=north west, xshift=1pt, yshift=-1pt,font=\scriptsize] at (angle2.east) {Inception of MLLMs};
    \end{pgfonlayer} ;
    \insertarrow{arrow1}{0,0}{SLM}{Statistical Language Model}
    \insertarrow{arrow2}{3.3cm,0}{NLM}{Neural Language Model}
    \insertarrow{arrow3}{6.6cm,0}{PLM}{Pre-trained Language Model}
    \insertarrow{arrow4}{9.9cm,0}{LLM}{Large Language Model}
    \insertarrow{arrow5}{13.2cm,0}{MLLM}{Multimodal\\ Large Language Model}
    
    \insertarrowsecond{arrow6}{3.1cm,-5cm}{LLM}{Medical Large Language Model}
    \insertarrowsecond{arrow6}{10.4cm,-5cm}{MLLM}{Medical Multimodal Large Language Model}
    
    \inserttimeline{1970s}{-1.50,-1}{HMM\cite{hmmllm}}{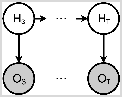}{1.0cm}{4.2pt}{-3.3pt}
    \inserttimeline{1990s}{0.24,-1}{n-gram\cite{ngramllm}}{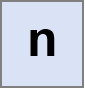}{0.6cm}{4pt}{-3.5pt}
    \inserttimeline{2010}{2.74,-1}{RNN\cite{14}}{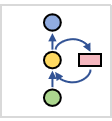}{0.9cm}{1.4pt}{0pt}
    \inserttimeline{2017}{5.10,-1}{Transformer\cite{1}}{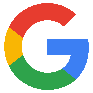}{0.8cm}{2pt}{-2pt}
    \inserttimeline{2018}{6.80,-1}{BERT\cite{2}}{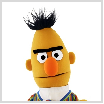}{0.85cm}{2pt}{-1pt}
    \inserttimeline{2020}{8.50,-1}{GPT-3\cite{gpt3}}{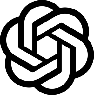}{0.8cm}{2pt}{-1.7pt}
    \inserttimeline{2022}{10.00,-1}{PaLM\cite{palm}}{google.png}{0.8cm}{0pt}{-2pt}
    \inserttimeline{2022}{11.50,-1}{Flamingo\cite{9}}{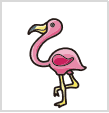}{0.8cm}{2pt}{0pt}
    \inserttimeline{2023}{13.00,-1}{GPT-4\cite{gpt4}}{gpt.png}{0.8cm}{2pt}{-2pt}
    \inserttimeline{2024}{14.30,-1}{MM1\cite{mm1}}{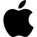}{0.8cm}{2pt}{0pt}
    \node[inner sep=0pt](divideline) at (6.75cm, -6.82cm) {\includegraphics[width=0.6cm,height= 0.43cm]{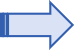}};
    \insertmedrepresent{2022}{0.1,-6}{Med-Palm\cite{medpalm}}{google.png}{0.8cm}{0}{0}
    \insertmedrepresent{2022}{2,-6}{ChatDoctor\cite{chatdoctor}}{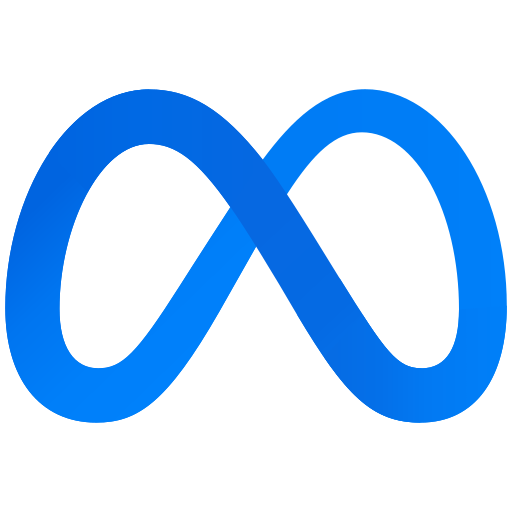}{0.8cm}{0}{0}
    \insertmedrepresent{2022}{4,-6}{HuatuoGPT\cite{huatuo})}{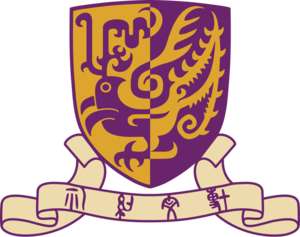}{1.03cm}{0.7pt}{0}
    \insertmedrepresent{2022}{5.7,-6}{AMIE\cite{amie}}{google.png}{0.8cm}{0}{0}
    \insertmedrepresent{2023}{7.8,-6}{Med-Flamingo\cite{medflamingo}}{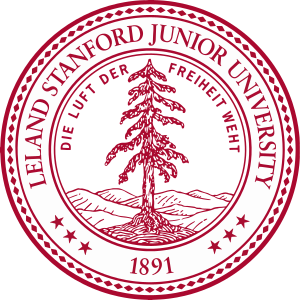}{0.82cm}{0}{1pt}
    \insertmedrepresent{2023}{9.9,-6}{LLaVA-Med\cite{llavamed}}{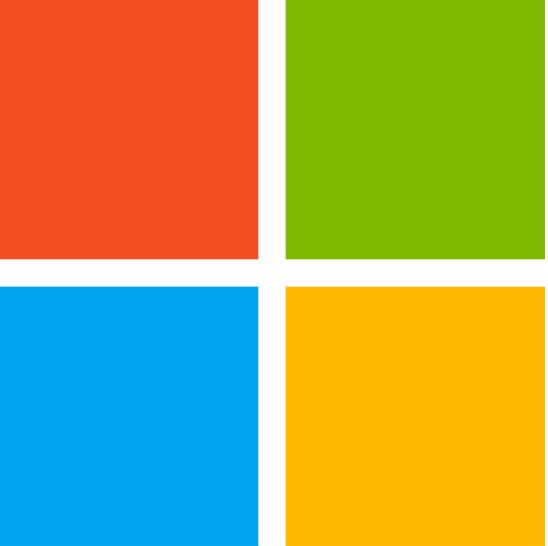}{0.77cm}{0.2pt}{-0.6pt}
    \insertmedrepresent{2023}{11.9,-6}{Med-Palm M\cite{medpalmm}}{google.png}{0.77cm}{0}{0}
    \insertmedrepresent{2024}{13.8,-6}{SkinGPT-4\cite{skingpt}}{kaust.png}{0.85cm}{0}{0}

\end{tikzpicture}
\vspace{8pt}
\caption{The Evolutionary Pathway of Multimodal Large Language Models with Medical Applications Highlighted.}
\label{fig:development}
\end{figure*}

While considering the unsteady accuracy and doubtful professionalism, the medical community concerns whether MLLMs are qualified for clinical application. To implement MLLMs in clinical healthcare, we believe MLLMs should meet professional requirements. Based on our investigation, we summarize several requirements. These requirements include various aspects, including but not limited to professionalism, accuracy, hallucination, and fairness. In the meantime, the corresponding evaluation benchmarks are mentioned. According to the concerns of the medical community and the requirements, we summarize a series of challenges faced by MLLMs in the context of medical healthcare. Feasible work has also been mentioned subsequently. By improving the evaluation benchmarks and making efforts to overcome the challenges, we are optimistic about the application of MLLMs in clinical settings.

\section{Preliminaries}
\graphicspath{{images/}}

\subsection{Large Language Models} 

% \subsubsection{Development} 

The language model is capable of understanding and generating human language. The development of the language model can be divided into four stages: SLM, NLM, PLM, and LLM. Representative models for each stage are illustrated in Figure~\ref{fig:development}.

Large language models excel by undergoing extensive pre-training on vast datasets and leveraging the Transformer architecture to achieve remarkable performance. Notable models, such as the GPT series and LLaMA, utilize the Transformer as their core architecture. The self-attention mechanisms and parallel computing capabilities within Transformers allow for efficient processing of long text sequences, thereby facilitating the handling of large-scale data and enabling substantial model parameter scaling. 
This structured approach to model architecture not only facilitates comprehensive learning, but also significantly enhances the applicability of the model to a range of tasks, such as text generation, translation, and complex question-answering.

\subsection{Multimodal Large Language Models}

In order to address the expansion of data modalities, which now include text, images, audio, and more, multiple modal large language models (MLLMs) have emerged as a focal point in AI research. Using multimodal data, tasks can be fully understood and executed. MLLMs are primarily composed of a large language model (LLM) that processes textual data and other modality encoders that process other modalities of data. Between the encoder of the Large Scale Language Model (LLM) and the other modalities, there is an alignment module that aligns the text input with the input of the other modalities into the feature space. By selecting the appropriate pre-trained LLM and modality encoder according to the specific task requirements, MLLMs have achieved significant breakthroughs in several fields. In this section, we will introduce the main components of MLLMs and several classical pre-training tasks for MLLMs.

When constructing application models, it is a way to flexibly combine multiple pre-trained tasks based on specific requirements. For instance, in the medical healthcare domain, Med-MLLM \cite{medmllm}—a medical multimodal large language model for future pandemics, has been designed for COVID-19 reporting, diagnosis, and prognosis by employing a three-tiered approach to pre-training tasks. During the pre-training phase, the model initially utilized contrastive learning methods for image module training. By leveraging patient-level comparisons—such as different types of medical images (e.g., X-rays and CT) of the chest from the same patient as inputs and incorporating image augmentation and regularization techniques, the model was trained to minimize losses between various images from the same patient. This process enhances the model's understanding of unique physiological characteristics. At the aspect of the language module, typical text pre-training tasks such as Masked Language Modeling (MLM), sentence reconstruction, and outcome-impression alignment were employed to boost the model’s performance in text comprehension and generation. In terms of multimodal pre-training involving both images and text, the method adopts a contrastive learning approach similar to that used in CLIP for image-text pairing while also improving upon it through integration with UMLS \cite{umls} knowledge bases and predefined objectives. Through these three aspects of pre-training, Med-MLLM is capable of addressing various applications including COVID-19 reporting (i.e., medical report generation), diagnosis (i.e., disease classification), and prognosis (i.e., outcome prediction).
\section{Applications}\label{sec:chapter_three}
\graphicspath{{images/}}

\begin{figure*}
    \centering
    \includegraphics[width=0.85\textwidth, keepaspectratio]{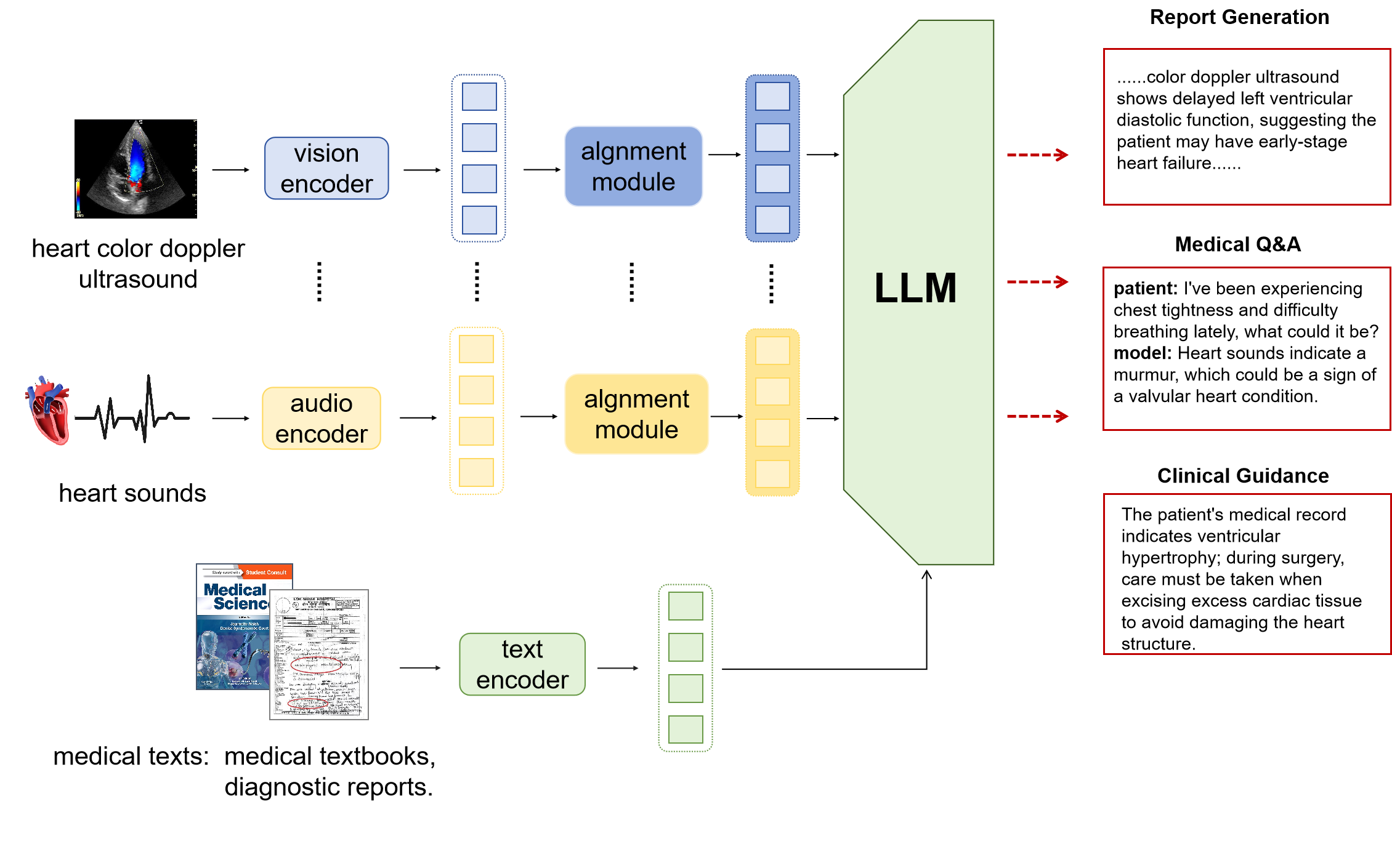} 
    \caption{Training Workflow for Medical Multimodal Large Language Models (MLLMs) in Application Tasks.}
    \label{fig:mllmstasks}
    \vspace{-0.4cm}
\end{figure*}

By leveraging various modalities of medical data, such as medical images, textual medical records, textbooks, and audio files, the multimodal large language model in healthcare can achieve a comprehensive understanding of the tasks and requirements at hand. This enables the model to perform medical tasks effectively, as illustrated in Figure~\ref{fig:mllmstasks}.

In this section, we will introduce the primary applications of multimodal large language models in healthcare, covering areas such as medical report generation, clinical communication, and clinical guidance.

\subsection{Medical Report Generation }\label{sec:medical_report}
Medical reports assist doctors in making diagnoses and formulating treatment plans. In addition, medical reports serve as a medium for transferring medical information, enabling medical personnel to track the disease. Every day, a large number of medical reports must be written by experienced radiologists or experts. The process costs significant time and human resources. In addition to the consumption of medical resources, the content of the report may be erroneous due to human error, leading to delays in treatment and misdiagnoses. To solve the problem, using artificial intelligence to produce accurate medical reports effectively has emerged as a promising task.

Through fine-tuning on medical healthcare data, LLMs have been proven to have great ability and potential in processing medical text\cite{ab1,ab2,ab3}. Leveraging this capability, LLMs cooperate with various modality modules to generate medical reports based on data from different modalities.

According to this perspective, the primary method at present is to utilize MLLMs to generate medical reports. The train of thought is to leverage medical images, such as X-ray \cite{xraygpt}, CT\cite{ct1,ct2}, MRIs \cite{brain}, and even 3D scan images\cite{3d}, as input data and combine the images with corresponding captions to form image-text pairs that serve as inputs to train the medical MLLMs. Specifically, X-ray GPT inputs X-ray images into a frozen vision encoder to extract image features. Subsequently, it utilizes the alignment module, which is a learnable linear transformation layer, to align the extracted image features with the caption text data. The aligned features are then input into a medical LLM, which is based on Vicuna and fine-tuned on medical data. Finally, the model will be guided to generate image captions via prompts such as ``What are the main findings and impressions of the given X-ray?'' The procedures allow the model to effectively comprehend and generate medical text.

Pre-training is a crucial phase for models that are designed for medical report generation, as it determines how effectively image and text data can be utilized. Moreover, the quality of pre-trained models is directly influenced by the effectiveness of medical report data. Research indicates that although the content of clinical medical reports adheres to the format of ``finding-impression'', the overall structure is generally disordered and does not follow a fixed regulation \cite{noorder}. Critical information often includes redundant content that is not conducive to model training. Currently, various methods have been proposed to address these issues. One strategy is to standardize or regenerate the format of medical report texts, helping the model focus on key content \cite{noorder,huatuo}. Although MedKLIP \cite{threespot} employs a Triplet Extraction module, which transforms the whole caption into a concise segment of text that includes only three key points: appearance, location, existence, to avoid the complexity of comprehending the content. MLLMs for the generation of medical reports have also been observed to primarily focus on ``impression'' and lack inferential capabilities for deep analysis. The phenomenon has led to the issue that some models have demonstrated excellent capability in assessing benchmarks, but have poor performance in clinical cases \cite{gpt4poor}. To enable models to possess inferential capability, \cite{gptformat,findingimpression} emphasizes the need for models to be trained in texts that exhibit a complete reasoning process. For instance, texts that adhere to the 'finding-impression' structure.

It should be noted that, unlike typical text generation tasks, medical reports have a strict logical structure. They generally consist of detailed observations and a corresponding summary \cite{structure}. Moreover, the generated text must meet the standards for medical terms. To achieve the objective of generating professional medical text, specific pre-training tasks can be implemented, such as SR (sentence reconstruction) \cite{medmllm} and MLM (masked language modeling). These tasks help the model study the writing styles employed by doctors and experts when composing reports. ChatCAD+ \cite{chatcad} is also employed to address the discrepancies in writing styles between experts and LLMs. In addition, some obscure terms in the medical field may be incorrectly separated into distinct words. For instance, 'cardiomegaly' might be divided into 'card-io-me-gal-y'. Establishing a specific dictionary and updating it regularly can help mitigate the problem \cite{dictionary}.

Preparatory tasks can also be efficiently completed by MLLMs to further automate the entire process. By employing MLLMs to record and summarize doctors' dictation \cite{record1,record2}, work stress can be significantly alleviated. Regarding imaging, the application of MLLMs for medical segmentation enables models to concentrate on the key part of the image \cite{segment1,segment2}, facilitating the execution of commands based on weak supervision.

\begin{table*}
    \vspace{0.4cm}
    \centering
    \caption{Details of Medical Multimodal Large Language Models.}
    \label{tab:multimodallargelanguagemodel1}
    \resizebox{1\linewidth}{!}{%
    \begin{tabular}{p{1.5cm} p{3cm} p{4cm} p{11cm}}
    \toprule
    \textbf{Year} & \textbf{Model} & \textbf{Base Model} & \textbf{Description} \\
    \midrule
        2023/01 & OphGLM\cite{meddialog1}& Chatglm-6b\cite{chatglm} & By fine-tuning the ophthalmology dialogue dataset and leveraging ChatGLM technology, aim to enhance the accuracy and interactivity of ophthalmic diagnoses. \\\\
        2023/07 & Med-Flamingo\cite{medflamingo}& LLaMA-7B\cite{5}& Undergo continuous pre-training with images and texts sourced from medical textbooks, journals, and authoritative medical publications, along with a newly created multimodal medical question-answering dataset styled after the USMLE, integrate medical imaging, clinical scenarios, and laboratory results.  Significant potential is demonstrated for extensive applications in clinical practice and education.\\\\
        2023/09 & Radiology-Llama2\cite{radiologyllama2} & LLaMA2\cite{llama2}& Specifically optimized for radiology and trained using the MIMIC-CXR and OpenI datasets. Incorporating instruction fine-tuning techniques, it demonstrates exceptional performance in generating radiology reports and providing diagnostic support.\\\\
        
        2023/11 & Qilin-Med-VL\cite{qilinmedvl}& Llama2-Chinese-13b-Chat\cite{llama2}& The first large-scale visual language model in the Chinese medical field enhances capabilities in generating medical captions and responding to complex medical inquiries.\\\\
        
        2024/10 & SigPhi-Med\cite{sigphimed}& phi-2\cite{phi2} &A lightweight biomedical multimodal language model with optimized visual encoder and compact language model components. Reduces the volume of training data while achieving superior performance compared to similar large models on tasks such as VQA-RAD, SLAKE, and Path-VQA.\\\\
        
        2023/02 & ChatCAD\cite{chatcad} & GPT-3(text-davinci-003)\cite{chatdavinci}& Integrate CAD to convert vectors or segmentation mask results from training the CAD network on medical imaging data into natural language text.  Combine this with the text generation and reasoning capabilities of LLMs to produce high-quality medical reports and provide interactive medical recommendations.\\\\
        2023/05 & MedBLIP\cite{medblip}& BioMedLM\cite{biomedlm}&Use a lightweight CAD system with a MedQFormer module to bridge the gap between 3D medical imaging, 2D pre-trained image encoders, and language models. Achieve state-of-the-art (SOTA) performance in zero-shot classification among subjects with healthy cognition, mild cognitive impairment (MCI), and Alzheimer's disease (AD). Demonstrate exceptional capabilities in medical image classification and visual question answering.\\\\
        2023/06 & PCLmed\cite{pclmed}& ChatGLM\cite{chatglm}&A medical report generation model based on the BLIP-2 framework has demonstrated exceptional performance in the task of generating medical reports.\\\\ 
        2023/05 & MedVInt\cite{pmcvqa}& PMC-LLaMA\cite{pmcllama}&Develop a generative medical visual question-answering model utilizing the diversity of the PMC-VQA\cite{pmcvqa}dataset to encompass multimodal medical scenarios. Support applications in diagnosis, education, and research, offering a flexible and efficient solution for medical AI.\\\\
        2023/05 & PathAsst\cite{pathasst}& Vicuna\cite{4}&The first multimodal generative foundational model specifically optimized for the field of pathology integrates 142,000 image-text data pairs and instruction fine-tuning techniques. It efficiently processes both pathological images and text tasks and demonstrates exceptional capabilities in pathological diagnosis, predictive analysis, and medical education.\\\\
        2023/06 & LLaVA-Med\cite{llavamed}& Vicuna\cite{4}&The training data for the model consists of 600,000 biomedical image-text pairs extracted from PubMed Central, along with multi-turn dialogue data generated using GPT-4.  Feature alignment of biomedical concepts is performed initially, followed by open-ended instruction fine-tuning.  The model primarily serves as a multimodal dialogue assistant, addressing open research questions related to biomedical images.\\\\
        2023/06 & XrayGPT\cite{xraygpt}& LLaMA2\cite{llama2}& Designed for chest X-ray analysis and related inquiries, the model leverages a large corpus of high-quality radiology report summaries.  Undergoing a two-stage training optimization process, it evolves into a conversational medical visual language model capable of analyzing chest X-rays and addressing open-ended questions effectively.\\\\
        2023/06 & ChatCAD+\cite{chatcadplus}& ChatGPT\cite{chatgpt}& Integrate various CAD models with large language models, utilizing hierarchical contextual learning and information from authoritative medical websites to generate high-quality medical reports and provide reliable healthcare recommendations.\\\\
        2023/07 & Med-PaLM M\cite{medpalmm}& PaLM\cite{palm}& Utilize professional medical examination questions, medical research literature, and consumer health inquiries to provide accurate and helpful answers to medical questions. Deliver precise and informative long-form responses to address various medical issues effectively.\\\\
        
        2023/08 & RadFM\cite{radfm}& MedLLAMA-13B\cite{pmcllama}&Integrate 2D or 3D medical scans with textual input to generate responses for various radiological tasks.  Achieve this through fine-tuning on a curated subset of 3 million visual-language pairs in radiology.  Handle multiple radiological tasks, including modality recognition, disease diagnosis, visual question answering, report generation, and diagnostic reasoning.\\\\
        2023/09 & R2GenGPT\cite{r2gengpt}&LLaMA2\cite{llama2}&Achieve optimization of specific domains by training only a lightweight visual alignment module while keeping all LLM parameters frozen. Focus on processing radiological images, such as chest X-rays, to generate comprehensive radiology reports with detailed descriptions of medical images.\\\\
   
        2023/10 & CXR-LLAVA\cite{cxrllava}&LLaMA2\cite{llama2}&Utilize annotated public chest X-ray images to pre-train a visual Transformer, which is then integrated with a large language model (LLM) and fine-tuned using freely available radiology report data. Develop an open-source multimodal large model for interpreting chest X-rays, simulating the diagnostic capabilities of radiologists.\\\\

        2023/11 & MAIRA-1\cite{maira1}&LLaMA2\cite{llama2}& Integrate the chest X-ray (CXR)-specific image encoder RAD-DINO and fine-tune using a dataset comprising CXRs and their corresponding reports. Focus on generating high-quality radiology reports from CXRs, with particular emphasis on the "Findings" section.\\\\    

       \bottomrule
    \end{tabular}}
    \vspace{-0.4cm}
\end{table*}

\begin{table*}
    \centering
    \label{tab:multimodallargelanguagemodel2}
    \resizebox{1\linewidth}{!}{%
    \begin{tabular}{p{1.5cm} p{3cm} p{4cm} p{11cm}}
    \toprule
    \textbf{Year} & \textbf{Model} & \textbf{Base Model} & \textbf{Description} \\
\midrule
        2024/01 & PeFoMed\cite{pefomed}& MiniGPT-v2\cite{minigptv2}&Designed for medical visual question answering (Med-VQA) and medical report generation (MRG) tasks, enabling accurate interpretation of medical images and comprehensive report generation.\\\\
        2024/03 & M3D-LaMed\cite{m3d}&LLaMA2\cite{llama2}&Aim to achieve direct understanding and reasoning of 3D medical images by integrating a pre-trained 3D visual encoder with an efficient 3D spatial pooling perceptron. Enable the execution of various tasks, including image-text retrieval, report generation, visual question answering, localization, and segmentation.\\\\
        2024/04 & MoE-TinyMed \cite{moetinymed}&LLaMA2\cite{llama2}&Employ a sparse activation mechanism to dynamically select expert sub-models for processing input.  Align multimodal medical features, fine-tune with medical instructions, and refine with expert guidance.  Ensure efficient and accurate performance in medical visual question answering, suitable for resource-constrained healthcare environments.\\\\

        2024/05 & Dia-LLaMA\cite{ctreport} & llama2-7B\cite{llama2}& Utilize chest CT images and reports with a disease prototype memory bank and a disease perception attention mechanism to enhance anomaly detection and report generation. Automatically produce high-quality chest CT reports.\\\\
        2024/06 & MAIRA-2\cite{maira2}&LLaMA2\cite{llama2}&Based on MAIRA-1
        \cite{maira1}, enhanced support for multiple input data types, improved image-text correspondence mechanisms, and strengthened visualization and interpretability features of reports.\\\\
        2024/07 & HuatuoGPT-Vision\cite{huatuo}&LLaMA2\cite{llama2}& Apply GPT-4V denoising and reformatting techniques to construct a high-quality PubMedVision dataset with 1.3 million medical VQA samples.  Significantly enhance model performance in medical multimodal tasks.\\\\
        2024/07 & miniGPT-Med\cite{minigptmed}&LLaMA2\cite{llama2}&Possess exceptional adaptability across various imaging modalities, including X-rays, CT scans, and MRIs.  Enable tasks such as medical report generation, visual question answering (VQA), and disease identification within medical images.\\\\
        2024/08 & SkinGPT-4\cite{skingpt}& LLaMA\cite{5}&Based on MiniGPT-4, fine-tuned with 52,929 skin disease images from public and proprietary datasets, accompanied by clinical concepts and physician notes. Efficiently address skin disease diagnostic tasks and provide interactive treatment recommendations. Support local deployment to ensure robust user privacy protection.\\\\
        2024/08 & LLaVA-Surg\cite{surgicalapp1}& LLaMA3-70B\cite{llama3}& Fine-tune using the Surg-QA dataset, which contains structured surgical knowledge extracted from publicly available surgical lecture videos, including pairs of surgical videos and corresponding instructions. Serve as a multimodal dialogue assistant for surgery, capable of answering open-ended questions related to surgical videos.\\\\
        2024/11 & HIST-AID \cite{histaid}& Aquila2\cite{aquila2}& A multimodal automated diagnostic framework leveraging patients' historical reports and imaging data to significantly enhance the accuracy of abnormality detection in chest X-rays.\\\\
        2025/01 & Vision -BioLLM \cite{visionbiollm}& openbiollms\cite{openbiollms}&A large-scale visual language model for biomedical visual dialogue, specifically addressing visual dialogue tasks in biomedical images.\\
  \bottomrule
    \end{tabular}}
    \vspace{-0.4cm}
\end{table*}

\subsection{Professional and Compassionate Medical Communication}
In recent years, chatbots have gained significant popularity. Especially in the medical healthcare field, chatbots have demonstrated promising prospects. Prior studies mainly focus on addressing single-modal data, such as text, by fine-tuning models based on datasets of doctor-patient conversations \cite{baize,chatdoctor}, and medical VQA \cite{medpalm}, and have achieved significant results.

Recently, the rapid development of MLLMs capable of handling multi-modal tasks opens up more possibilities for chatbots. Patients can upload medical images and videos for professional medical answers from chatbots. For example, when fine-tuned on a large amount of skin data, Skin-GPT4 \cite{skingpt} is able to offer professional medical advice on skin diseases. LLaVA-Med \cite{llavamed} is capable of processing medical images to perform VQA tasks on X-rays, CT scans, and MRIs, and has achieved SOTA performance on several closed datasets for medical VQA.

Although these MLLMs show outstanding performance on various datasets and assessment benchmarks, their excessive dependence on the impressions may undermine their inferential capabilities, leading to poor performance in clinical cases \cite{gpt4vpoor}.

Besides, research indicates that people still prefer to receive medical services from humans. Interactivity and affinity are the main reasons behind this preference \cite{facetoface,preferhuman}. For patients, feeling empathy and being understood are as important as diagnostic accuracy. At the same time, the fast-paced lifestyle and the prevalence of social media have significantly raised the demand for psychological services \cite{mentalhealth1,mentalhealth2}. In psychological treatment, besides the execution of the treatment plan, the therapeutic effects of conversations between experts and patients should not be underestimated \cite{consel1}. In the realm of psychological therapy, chatbots have demonstrated their potential \cite{chatboteffect1,chatboteffect2}, helping alleviate the pressure on healthcare capacity and reduce treatment costs. Also, because patients can express themselves more freely without embarrassment, chatbots are preferable to humans \cite{patientrelax1,patientrelax2}.

Psychological consultation chatbots based on LLMs have been introduced \cite{llmchatbot1,llmchatbot2,llmchatbot3}. They mainly focus on the information contained in patients' statements, such as sentiment, degree of cooperation, and communication habits, enabling LLMs to respond empathetically—through questioning, comforting, affirming, listening, trusting, and other forms of emotional support. Additionally, they employ relevant benchmarks for evaluation and optimization purposes \cite{chatevaluate1,chatevaluate2}. However, the information expressed through text still has limitations. For example, the sentence “That's great” can convey very different emotions depending on facial expressions or tone of voice, such as rolling eyes or speaking with irony. This may lead the model to misinterpret the emotion. Therefore, MLLMs have been introduced to address this issue. By extracting facial motion, body motion, eye movement, as well as the rhythm and tone of speech, the patient’s state can be analyzed comprehensively \cite{audiosentiment,allsentiment}. By attaining deeper understanding of the patient, a face-to-face diagnostic effect can be achieved. The problem of empathy deficiency can thus be alleviated \cite{llmchatbot1}.

Chatbots in the medical healthcare realm have the characteristics of timeliness, low cost, and high efficiency, which the current medical system pursues. With these advantages combined, patients will be more willing to visit clinics. Related legislation and quality surveillance work should be implemented before widespread adoption \cite{regulate}.

\subsection{Clinical Surgery Assistance}
Due to the lack of professional medical knowledge,  patients have to rely on experts or doctors for detailed explanations and analysis of the surgery.  Even doctors with  limited clinical experience still need to consult expert specialists for help.  However,  high-level specialists have a huge workload to manage every day.  It's difficult to care for every case within a limited  time.  To mitigate the issue,  computer has already been introduced\cite{computerassist1,computerassist2}.  While the techniques still rely on experts to  answer specific clinical questions. 

Some MLLMs have been proposed in the field of medical surgery to assist with or replace the work of experts.  Introducing  Surgery VQA data \cite{surgicalvqa} and training MLLMs on surgical video data to enable them to answer questions in the  surgical setting\cite{surgicalapp1,surgicalapp2}.  SurgicalGPT \cite{surgicalgpt} combines GPT with a visual encoder and is fine-tuned on endoscopic images of  the kidney.  The model demonstrates strong capabilities in this domain and achieves SOTA performance on the  EndoVis18-VQA, Cholec80-VQA, and PSI-AVA-VQA datasets. 

VQA surgical MLLMs mainly concentrate on the region of interest related to the disease,  usually overlooking the background information.  The lack of background knowledge may cause the model to have a flawed  understanding of the surgery.  It leads to potential misjudgments\cite{maylack}.   Taking into account the motion of the  surgical operation, the surgical tools, and other factors may imply background knowledge,  allowing the model to gain a more comprehensive understanding of the surgery and provide more accurate responses \cite{toolgesture1,toolgesture2}.  Expanding the data categories used by the surgical model, such as race, region, EHRs,  and medical history, is a promising work to improve the model's performance.

Moereover, based on achievements in medical report generation(refer to Section~\ref{sec:medical_report}), MLLMs are also employed to generate surgical reports and analyses \cite{report1,report2} that assist doctors in summarizing  the process that provide decision support for subsequent procedures.

Since there is no room for error in clinical surgery,  the response and guidance from the model must be exemplary. Although the related MLLMs have demonstrated impressive  capabilities on closed datasets, there is still a long way to go before their clinical application. Moreover, the related legislation should be enacted, and the responsibilities should be clarified.  Regarding data,  currently only endoscopic data is ample,  while other types are scarcely available. Expanding the category of data is the premise for surgery MLLMs to be widely  used. 

\section{Data}

\begin{figure}
    \centering
    \includegraphics[width=0.49\textwidth, keepaspectratio]{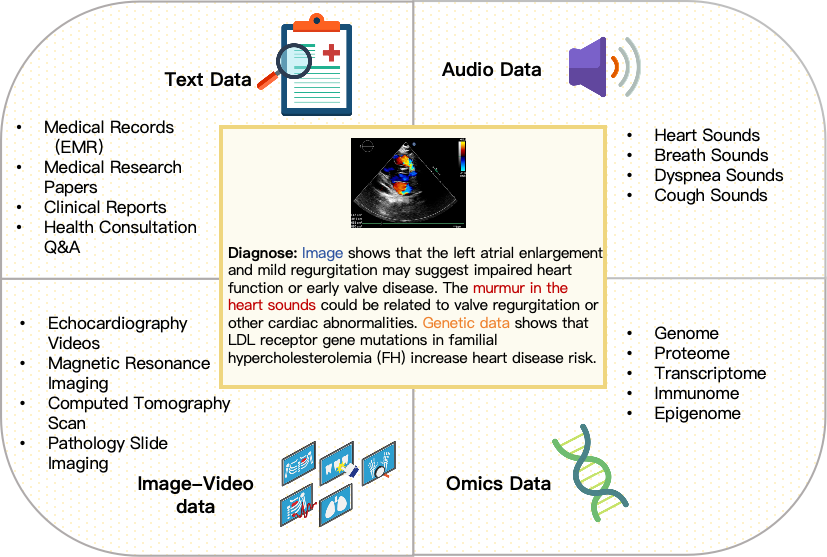} 
    \caption{Comprehensive Diagnosis Based on Various multimodal Data.}
    \label{fig:datatraits}
    \vspace{-0.4cm}
\end{figure}

In this section, we will explore the various types of multimodal data within the healthcare domain, as shown in Figure \ref{fig:datatraits}. We will analyze how different structures, types, and categories of data can enhance the potential of models in performing diverse tasks and fostering professional competencies. Various forms of medical datasets across different modalities for training purposes have been compiled in Table~\ref{tab:training_data_1} and Table~\ref{tab:training_data_2}. We will enumerate and introduce several existing datasets utilized for training multimodal large language models Table~\ref{tab:mllmtrainingdata1}and large language models Table~\ref{tab:llmtrainingdata1}. Furthermore, due to concerns regarding privacy and security, challenges associated with data collection have led to a relative scarcity of data for medical multimodal large language models. To address this issue, we will discuss effective solutions from two perspectives: model optimization and data augmentation.

\begin{table} \small
    \centering
    \caption{Categories of Training Data for MLLMs.}
    \label{tab:training_data_1}
    \renewcommand{\arraystretch}{1.2} % 设置表格行距为原来的1.2倍
    \begin{tabular}{p{2.6cm} p{5.8cm}}
    \toprule
    \textbf{Type} & \textbf{Description} \\
    \midrule
    Image-Report & Medical images (e.g., X-rays, CT scans) paired with professional reports or natural language explanations. \\
    Image-Caption & Medical images paired with textual descriptions or labels, used in annotation, segmentation, tracking, and multimodal learning. \\
    QA & Medical visual question answering data combining images with related questions and answers. \\
    Scientific Literature & Biomedical or medical literature used in vision-language pretraining. \\
    Instruction Following & Medical images with instructions for executing tasks based on visual understanding. \\
    Hybrid & Integrated data combining multiple modalities or formats. \\
    \bottomrule
    \end{tabular}
    \renewcommand{\arraystretch}{1.0} % 恢复默认值
\end{table}

\begin{table} \small
    \centering
    \caption{Categories of Training Data for LLMs.}
    \label{tab:training_data_2}
    \renewcommand{\arraystretch}{1.2} % 设置表格行距为原来的1.5倍
    \begin{tabular}{p{2.6cm} p{5.8cm}}
    \toprule
    \textbf{Type} & \textbf{Description} \\
    \midrule
    QA & Medical question-and-answer pairs for natural language understanding and reasoning. \\
    EHR & Clinical treatment data and structured/unstructured patient health records. \\
    Dialogue & Multi-turn doctor-patient conversations used for training dialogue systems. \\
    Medical Question Summarization & Medical problems and their summaries, used in summarization tasks. \\
    Knowledge Base & Medical entities and relations for constructing knowledge graphs. \\
    Scientific Literature & Biomedical research literature for evidence-based reasoning. \\
    Web Data & Online texts annotated with medical entities, including social media and domain articles. \\
    \bottomrule
    \end{tabular}
    \renewcommand{\arraystretch}{1.0} % 恢复默认值，防止影响其他表格
\end{table}

\subsection{Image}
Medical imaging, which can clearly reveal the structure of the body and organs, is a crucial component of medical analysis. Through various forms of imaging, such as radiological and photographic images, experts can comprehensively extract information for analysis and make medical treatment plans. To effectively process medical images, deep learning has been introduced as an aid \cite{cnnxray1,cnnsegment,dldiagnose}. Recently, MLLMs, known for their advantages in processing multimodal tasks, have been utilized to address tasks involving images. It has been demonstrated that MLLMs have achieved success across various forms of medical imaging.

\subsubsection{Radiological Images}
MRI is a non-invasive medical imaging technique that uses magnetic fields, radio waves, and computer technology to produce detailed images of the internal structures of the human body. It is suitable for soft tissue. MLLMs have been applied to process relevant segments such as the brain \cite{brainmri}, articulations \cite{mribone}, and others. Moreover, MRI techniques feature various sequences that focus on different components. For example, T1-weighted sequences highlight the signal of fat, while T2-weighted sequences are sensitive to water.
Medical staff select different sequences of MRIs to obtain specific clinical information based on different demands. At the same time, various sequences can be combined to obtain comprehensive disease information \cite{zhoutao}.

Based on the varying extents of absorption due to differences in density and thickness, X-rays are leveraged to create medical images. MLLMs have been utilized to process images of human body parts captured by X-rays, such as the chest \cite{xraychest1} \cite{xraychest2} and bones \cite{bonexray}. CT imaging, also produced by X-rays, performs cross-sectional scans of body parts. This technique provides clear anatomical images by virtue of its excellent resolution. CT images of the chest \cite{ctreport}, brain \cite{brainct1} \cite{brainct2}, and bones \cite{bonect} have been proven to be effectively processed by MLLMs for medical examinations. Considering the high dimensionality of CT images, MLLMs must overcome the instability and inaccuracy that arise from higher dimensions. To address this, for example, a pre-trained specific encoder (ViT3D) combined with a perceiver is used to extract visual information from 3D images \cite{ctreport}. Additionally, the way to overcome the challenges of generating captions from 3D images, which is more difficult and complex than from 2D images, is crucial \cite{brainct2}.

\subsubsection{Photographic Images}

Photographic images refer to those captured using optical devices, typically representing the surface or interior structures of the body directly. Common applications include skin imaging and endoscopic imaging. By processing photos directly captured by optical devices for the eyes \cite{eyephoto} and skin \cite{skingpt}, and by taking endoscopic photos of structures such as polyps and the rectum for diagnosis \cite{endoscopy}, MLLMs can help relieve the workload of experts.

\subsection{Text}
Medical text is a critical carrier of medical information and knowledge, clarifying the accordance, processes, and consequences of medical analysis. There are many forms of medical text, including medical reports, dialogues between doctors and their patients, electronic health records, professional textbooks, and more.
Some progress has been made in processing text tasks by single-modal LLMs. By extracting descriptions of disease symptoms, these models can diagnose and make treatment plans automatically and in a timely manner \cite{oldconsciousllm,llmsingle2,singlellm3}.

In the development of MLLMs, text data keep serving as a crucial role. Through different forms of data, can provide MLLMs abilities for various text tasks. MedDialog \cite{meddialog} is a large-scale medical dialogue dataset, available in both English and Chinese, including over 100 disease specialties. By fine-tuned with dialogue data, MLLMs can serve outstandingly as doctors in medical dialogue tasks, due to improved professionalism and friendliness \cite{meddialog1,medrecommend}.

To enhance the capabilities of vision question answering tasks, VQA data can be leveraged. By using VQA text data to fine-tune the model and combining this text with previously processed image features, medical MLLMs can more effectively handle medical multimodal tasks and provide medical assistance \cite{vqamodel2,huatuo,surgicalgpt,vqamodel}.

In addition to datasets specifically constructed for training, existing data, such as medical reports and electronic health records (EHRs), also possess great value for model training. Extracting logical information, such as ``finding-impression'' data, to train models enhances the ability to infer \cite{reportimage1} \cite{reportimage2} \cite{reportimage3}. Alternatively, directly combining historical medical reports with medical images can aid in diagnosing the condition of a disease \cite{historyreportcombine}.

Moreover, EHRs, which have a discrete structure of patient information, have also been proven to facilitate the training of clinical LLMs \cite{ehrproof}. So far, the use of EHR data for training models is still limited. Training models with EHR data is a promising approach due to its volume and high informativeness.

\vspace{-10pt}
\subsection{Data with Potential}
\subsubsection{Audio}

Besides visual forms, a patient's body can also convey signals of health status through auditory information. For instance, the sound of a patient's cough, breathing, heartbeat, and other audio data are valuable for medical analysis. Using MLLMs to capture the details of audio information can provide the necessary data for diagnosis.
Specifically, RespLLM \cite{audiomulti} transforms the sound signals of coughing and breathing into spectrograms. Subsequently, it performs segmentation and reflection to convert these spectrograms into embeddings that align with text data. In the realm of psychological healthcare, the tone, tempo, and other characteristics of a patient's expression hold significant value as diagnostic references. These aspects are crucial for understanding the patient well.

Furthermore, audio data can also be considered the raw form of text data. When generating medical reports, the dictated notes from medical experts and the self-descriptions from patients usually form the basis of the diagnosis and the core structure of the medical report \cite{audiorecord1,audiorecord2}. In the aspect of medical education, transforming experts' clinical dictations and medical podcasts into text allows these resources to be leveraged as materials for medical teaching \cite{audioedu}. Improving the transformation of audio to text can enhance the automation and intelligence of medical services.

\subsubsection{Omics}
With the significant development of sequencing techniques and the associated continuously decreasing costs, omics data have been widely applied in medical diagnosis and clinical research and have achieved some success \cite{omicsresearch1,omicsresearch2}. Common omics data include the genome, proteome, transcriptome, immunome, epigenome, metabolome, and microbiome. Using these data, personalized medical services can be offered to improve the quality of medical care.

Given the complexity and variety of omics data, the key focus of current research is to effectively integrate and combine various modalities of omics data. LLMs have been shown to have the capability to predict and analyze components or elements of omics data \cite{omicllm1,omicllm2}.
Combining omics data with other modalities of medical data helps models understand the complex interactions between the human body, disease biological entities, and clinical information. This integration is key to advancing medical MLLMs toward greater precision and personalization, which holds vast potential for application.

\subsection{Data Scarcity}
As mentioned above, it has been proven that medical MLLMs can be applied to achieve favorable outcomes across various types and aspects of diseases. However, the potential of medical MLLMs has not been fully exploited. The performance of these models is not satisfactory, and there is still a long way to go before they can be applied in clinical applications. Meanwhile, research has shown that medical models perform well in specific body parts and types of medical data forms, but their performance is generally inferior to that of general models in average medical tasks \cite{inferior}.
The scarcity of data is attributed to the main reason for the poor performance and limited adaptability of the models. Due to legal and ethical restrictions, encountering resistance to obtaining medical data is inevitable. This issue primarily appears in text data. However, the rapid evolution of medical knowledge and the emergence of rare diseases demand robust MLLMs, which require substantial amounts of high-quality data for support. In response to this challenge, various strategies have been proposed to address it.

\noindent\textbf{Models’work}: It is the consensus that fine-tuning is the primary value of pre-trained models, and also of LLMs. By fine-tuning foundational or pre-trained models, they already possess the basic capabilities required for specific general tasks.

Introduced by Google in 2022, PaLM possesses the ability to address complex language tasks. By fine-tuning it with medical data, Med-PaLM \cite{medpalm} has been proposed for medical applications, expanding the application settings, enabling the model to excellently process and generate medical text. Similarly, LLaVA-Med \cite{llavamed} is an improvement on LLaVA, achieved by fine-tuning with a combination of medical images and dialogue. This enhancement can improve performance in medical conversations and question-answering tasks. PMC-LLaMA \cite{pmcllama} is enhanced with the help of LLaMA, which possesses strong capabilities in natural language comprehension. It has been fine-tuned using 4.8 million biomedical academic papers and 30,000 medical textbooks, demonstrating impressive performance in public medical question-answering tasks.Furthermore, enhancing the inference capabilities of models can improve their generalization abilities. By focusing on inference rather than memorizing impressions, this approach reduces dependency on large-scale data.

For instance, Med-MLLM \cite{medmllm}has enhanced logical inference capabilities through pre-training. Even when dealing with rare diseases, the model still maintains high performance.

\noindent\textbf{Data’s Work}: Besides improving adaptability to scarce data, expanding the data scale is a direct strategy to address this issue. Research indicates that AI-generated data can be effectively utilized for fine-tuning downstream tasks \cite{proveaigenerate}. Various forms of text data generated by ChatGPT have been utilized for data expansion or augmentation. For instance, \cite{gptgenerate1} utilizes GPT-4 by programming it to generate questions based on medical textbooks. This approach reduces the occurrence of misleading phenomena and ensures that the generated questions and answers are consistent with the textbook content. \cite{gptformat} uses a retriever to obtain possible descriptions from textbooks and medical databases for caption-lacked medical images. It then leverages GPT-4 to generate captions based on the retrieved text or to format existing captions for better training outcomes.

If the data possessed are of high quality, GPT can be used to generate text with similar meanings and combined with image enhancement techniques to achieve data expansion \cite{similar1}.

\section{Traits}
\subsection{Professionalism}
\graphicspath{{images/}}
Unlike routine MLLMs, medical MLLMs are required to exhibit strict professionalism. To put it directly, medical MLLMs should possess the expertise of specialists in the corresponding fields of tasks. Currently, it is still easy to recognize that the level of specialization in AI remains inferior to human expertise \cite{61eye,accudown}. To meet the standards of clinical application and professional demands, aligning the model's performance with the abilities of real experts is a viable strategy.

Grasping precise medical knowledge is the foundation for providing medical services. Medical knowledge typically uses text as its medium. By utilizing specific medical texts combined with other data modalities to fine-tune the model, MLLMs gain a robust capability to learn and apply knowledge effectively.

In Section~\ref{sec:chapter_three}, we have introduced routine methods to help models acquire capabilities in the medical domain. In addition to utilizing appropriate forms or content of data, constructing a professional medical dictionary from clinical notes and clinical reports helps separate rare medical terms, aiding models in capturing the crucial information within the text \cite{dictionary1,dictionary2}. These preparatory works serve an important role, especially for rare diseases \cite{raredictionary}.

% \begin{figure*}
%     \centering
%     \vspace{10pt} 
%     \includegraphics[width=1.00\textwidth, keepaspectratio]{evaluate1.jpg} 

%     \label{fig:evaluate1}

%     \includegraphics[width=1.00\textwidth, keepaspectratio]{evaluate2.jpg} 
%     \label{fig:evaluate2}
%     \caption{GPT as a Medical Expert: Evaluating the Quality of Medical MLLMs' Outputs to Analyze Model Performance.}   
% \end{figure*}

\noindent\textbf{Evaluate Methods}.
Various aspects can be undertaken to evaluate the professionalism of MLLMs. Generally, feedback from models, such as the generation of medical reports and responses in conversations, provides valuable guidance.

In terms of surface form, models' expressions should conform to those of medical experts. By comparing the feedback generated in text form with clinical text, similarity is considered as the evaluation benchmark. Common benchmarks for natural language generation \cite{bleu,rouge,cider} are utilized to evaluate the lexical, semantic, structural, and content salience between the text being evaluated and the reference text to determine similarity \cite{medmllm,rougesingle}. By improving the models according to the benchmarks, the models can express themselves with a professional and strict structural format.

Besides aligning with the expressions of experts, the logic and precision of the content are the core of professionalism. The general assessment method is to evaluate the accuracy, details, features, logic, and other characteristics of models' outputs. The evaluation requires a high level of comprehension, which cannot be measured by a fixed benchmark. Commonly, evaluation methods can be conducted manually or by AI.

\setlength{\parskip}{1em}  
\noindent\textbf{Manual Evaluation}: The standard process requires experts to rate different characteristics on a scale as a basis for evaluation \cite{huatuo,platformrate,artificialassess1}. The score reflects the level of professionalism to some degree. Specifically, \cite{huatuo} invites experts to rate on four dimensions: (1) accuracy of description; (2) level of detail; (3) consideration of overall characteristics; (4) usefulness to application.

\setlength{\parskip}{1em}  
\noindent\textbf{Automatic Evaluation}: We categorize automatic evaluation into two main cases: (1) Models perform tasks with an evaluative nature; (2) Utilize AI to evaluate the models' outputs in an expert-like manner.

Various dataset benchmarks are created for specific tasks, but they cannot provide authority. Similar to human examinations, utilizing models to participate in professional medical assessments is a way to evaluate their professionalism.USMLE, which stands for United States Medical Licensing Examination, is the medical licensing examination for the United States. This authoritative examination has been used to evaluate the performance of GPT series models, Med-PaLM, and other models \cite{usml1,usml2,usml3}. Although the evaluation results are satisfactory, with GPT-4 and Med-PaLM achieving over 86 percent accuracy, rivaling human experts \cite{usml1}, models still perform unsatisfactorily in clinical settings. Over-dependence on memorization leads to a lack of inference ability, preventing the models from adapting to clinical settings \cite{gptformat}.

Utilizing AI for analysis is similar to manual evaluation, where large language models are instructed to play the role of an expert and rate MLLM outputs across different dimensions \cite{gptformat} \cite{beyond}, as illustrated in Figure~\ref{fig:evaluate}. Specifically, \cite{beyond} suggests that Gemini-Pro and GPT-4 play the role of experts in the medical imaging research field, scoring based on different assessment criteria.

Though evaluating MLLMs based on those benchmarks, we can directly observe if the surface form and insight comprehension conform to medical professionalism.

\subsection{Hallucination }

At present, feedback provided by medical MLLMs is still met with skepticism in the medical field. In addition to accuracy, the credibility of the information generated plays a critical role in the evaluation process. Hallucination is the primary obstacle for this reason.

% \begin{figure*}
%     \centering
%     \vspace{10pt} 
%     \includegraphics[width=0.75\textwidth, keepaspectratio]{hallucination1.jpg} 
%     \label{fig:hallucination1}
%     \caption{Hallucination due to object misidentification.}   

%     \vspace{10pt} 
%     \includegraphics[width=0.75\textwidth, keepaspectratio]{hallucination2.jpg}
%     \vspace{10pt} 
%     \label{fig:hallucination2}
%     \caption{hallucination due to misperceived object relationships}   
% \end{figure*}

Hallucination refers to instances where LLMs generate plausible but incorrect or fabricated facts \cite{hdefinition1,hdefinition2,hdefinition3,hdefinition4}. It means that although models may perform precise and logical inferences, the direction of the content for diagnosis could be entirely incorrect. The misleading information may lead to serious consequences in the medical field. Research has identified the reasons for this phenomenon, including poor quality of instruction \cite{reasoninstruct} and inadequate data quality and quantity, which exacerbate the issue. Rapidly updated data in the medical field will hinder models that rely on memory and have weak inference capabilities \cite{dependent}.

To mitigate the issue, leveraging high-quality medical data is an effective strategy. As \cite{textbooksolve} proposes, generating responses based strictly on professional textbooks is a key approach. Alternatively, designing a self-check structure could help improve LLMs' responses by avoiding hallucinations \cite{check1,correction1}.

When it comes to MLLMs, additional problems have been introduced \cite{metoo1,metoo2}. First, similar to LLMs, insufficient data will be a problem. The scarcity of data will lead to the issue where different modal data cannot be aligned adequately, thereby causing hallucinations \cite{align1}. Misidentification of the object has also been found to be the main cause \cite{objectmistake}. Research has found that existing large language models focused on vision often overlook image information, relying excessively on the pre-existing knowledge of LLMs to make predictions \cite{evaluatehallucination1}. Based on experience, hallucinations caused by misidentification can be categorized into three types: type, property, and relationship \cite{hallucination1}.

\begin{figure}
    \centering
    \begin{minipage}{0.45\textwidth} 
        \centering
        \includegraphics[width=\linewidth, keepaspectratio]{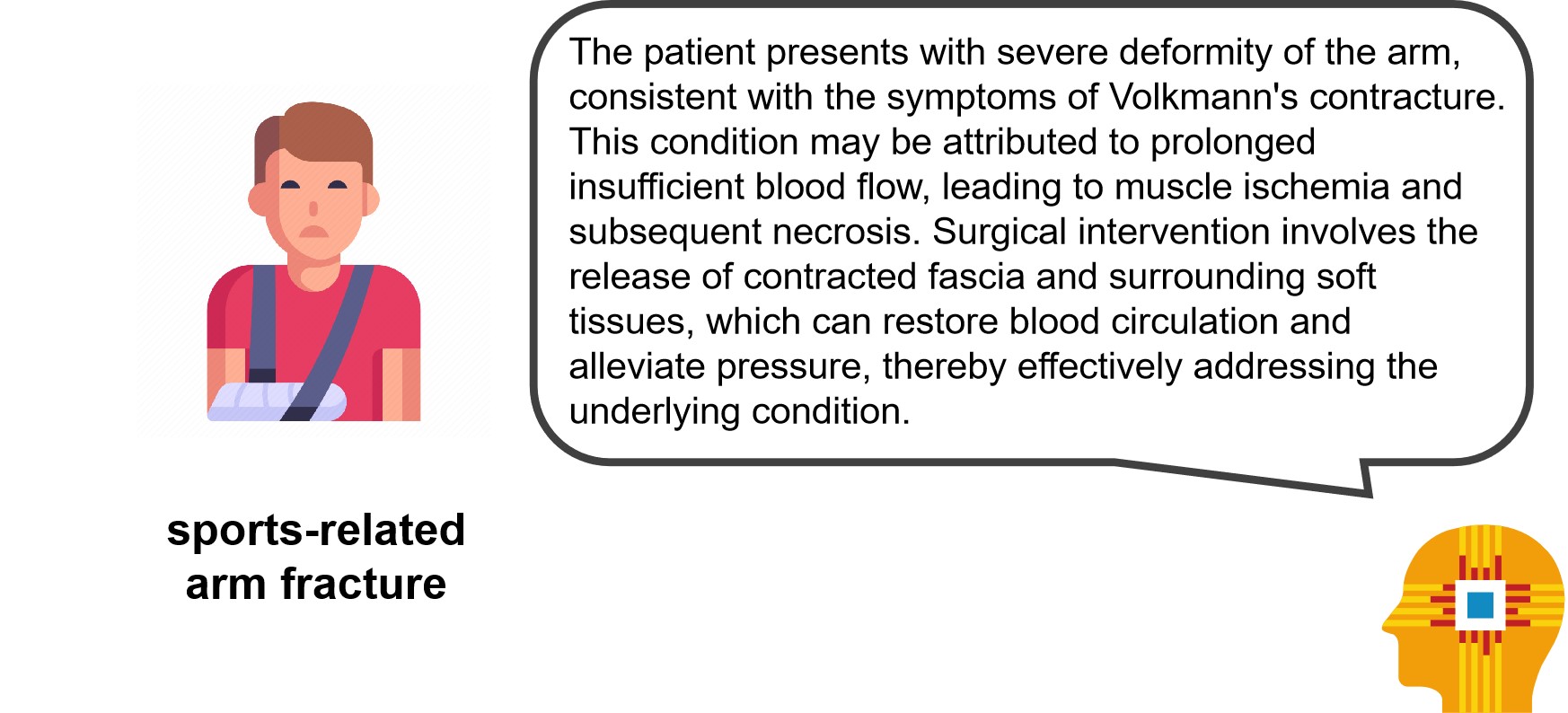}
        \caption{Hallucination due to object misidentification.}
        \label{fig:hallucination1}
    \end{minipage}%
\vspace{0.6cm}
    \begin{minipage}{0.45\textwidth} 
        \centering
        \includegraphics[width=\linewidth, keepaspectratio]{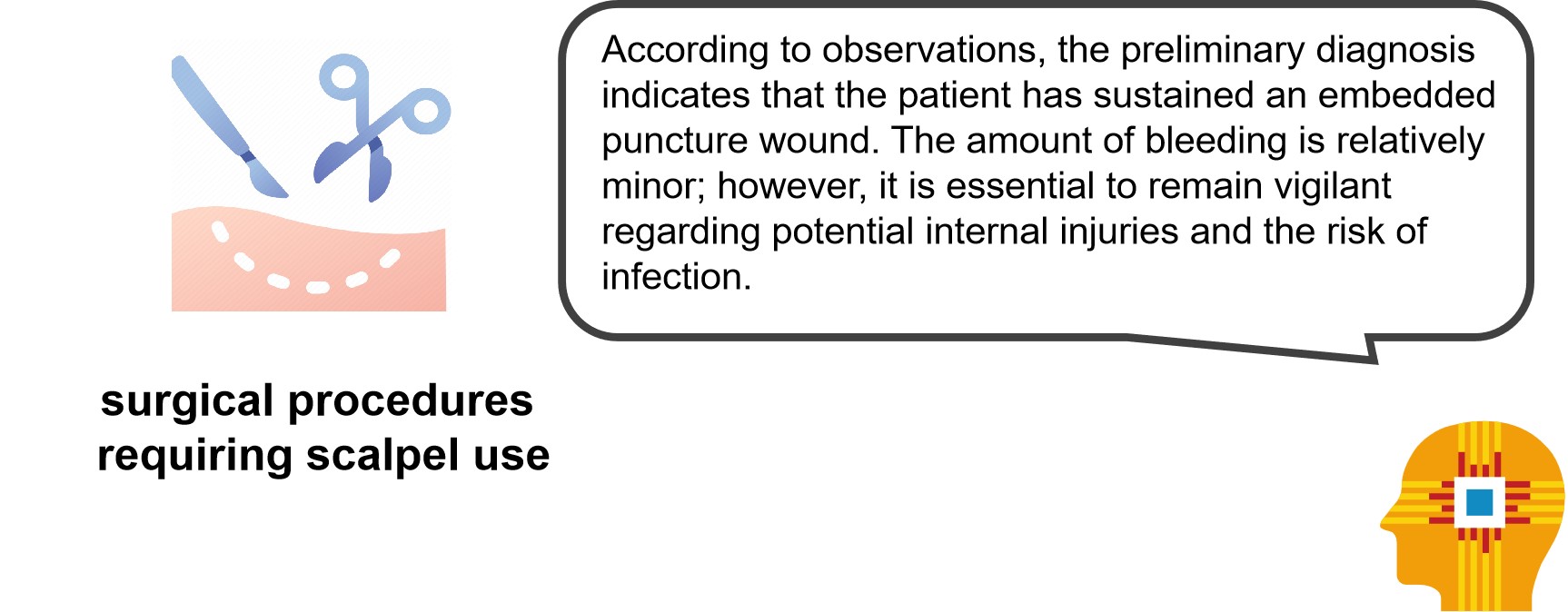}
        \caption{Hallucination due to misperceived object relationships.}
        \label{fig:hallucination2}
    \end{minipage}
\end{figure}

The main reason for the first two categories is that models cannot identify and focus on the key parts of the image, leading to hallucinations, as illustrated in Figure~\ref{fig:hallucination1}. Besides the lack of data, other key reasons may include insufficient parameters in the vision and alignment modules, or inadequate resolution in the visual encoder \cite{xiangsuheqformer}. These shortcomings lead to the issue of information loss. To solve this issue, in addition to adjusting the module specifications, techniques like image segmentation can be leveraged in order to highlight the key part \cite{misobject1,misobject2}.

The third point, the abstract relationship, must also be addressed. Relationship hallucinations can be categorized into co-occurrence, counterfactual, and illusion relationship hallucinations \cite{evaluatehallucination1}.

These issues are mainly caused by an overdependence on stereotypes. The diagram in Figure~\ref{fig:hallucination2} illustrates how the inherent perception of the relationship between the tool and the wound can lead to counterfactual illusions, resulting in misjudgments by the model.
Besides improving the models' ability to focus, leveraging additional boundary-accessorizing techniques can help the model concentrate on the parts that highlight relationships \cite{evaluatehallucination1}. A hallucinatory confrontational test can help prevent the model from being misled by misleading images \cite{against1,against2}.

\noindent\textbf{Evaluate Methods}. The evaluation of hallucination is similar to that of accuracy. By transforming the evaluation into binary questions focused on hallucination-related issues, the performance can be assessed based on the model's responses \cite{qassess1,qassess2,qassess3}.
It is undoubted that MLLMs have great potential to serve as an information source in the medical education and research realms, but this potential can only be realized if high-quality supervision and stringent quality control are ensured.

\subsection{Fairness and Bias}
Models' accuracy and professionalism may also be threatened by issues of fairness and bias. The bias includes race, social role, region, and so on. Data for AI algorithms and MLLMs primarily come from the internet, where they may be influenced by various opinions, predominantly mainstream prejudices or stereotypes. This may cause the models to seriously misjudge or produce inappropriate statements. In the context of medical scenarios, specifically, \cite{whiteman} found that the model tends to predict higher costs and longer hospital stays for white populations. Furthermore, in healthcare situations where the expected survival rates are already high, the model exhibits a tendency to display more optimistic projections. \cite{blackman} took the cost of medical expenses into consideration and reduced the treatment rates for minority groups based on stereotypes. Additionally, research that excessively focuses on the same population leads to unstable datasets, which is one of the main reasons for prejudice \cite{europegene}. For example, disadvantaged groups that are not covered by medical insurance often yield inaccurate results \cite{lessdataresult}. \cite{medmllm} indicates that ethnic groups from different regions have poor adaptability under cross-validation. Medical data combined from diverse populations demands high transferability and stability from models.

Resolution can be considered from different perspectives. In terms of datasets, filtering and averaging representative data is an effective approach to prevent bias from the majority class \cite{against1}. \cite{antifactbias} achieves de-biasing effects by leveraging counterfactual data and resampling underrepresented data that may otherwise be overlooked. Regarding study strategies, RLHF \cite{rlhf} can be employed to align models with human values. At the same time, measures must be implemented to prevent the model from generating potentially racist statements. Enhancing model empathy has proven to be an effective approach to making them more friendly \cite{empathy}.

\noindent\textbf{Evaluate Methods}.
Related evaluation benchmarks primarily combine patient information with classic data types, such as question-answering and medical reports, to conduct assessments.

For example, based on question-answering data, Harvard-FairVLMed \cite{luobasedata} incorporates additional metadata such as race, gender, and ethnicity. FMBench is introduced as an evaluation benchmark, built upon Harvard-FairVLMed by incorporating additional metadata, such as ethnicity, race, and gender, to evaluate responses based on questions from various populations.

\section{Future Works}
\subsection{Rapidly Updated Data}

With the development of medical research and sudden discoveries in the biological field, unprecedented medical concepts are constantly being proposed, such as new diseases and new treatments. The paradox is that MLLMs are generally trained on inherent data, while in clinical scenarios they often have to confront unknown knowledge. This threatens the professionalism and accuracy of the model.
Some solutions have been proposed to mitigate the situation, such as constructing a specific vocabulary dictionary for the medical domain to help the model understand the meaning of new words and concepts, thus preventing the model from misinterpreting obscure terms \cite{dictionary1,dictionary2}.

In terms of improvement during the training phase, a dynamic training strategy is employed to address the problem of knowledge lag. By continuously inputting new data and knowledge through fine-tuning, the model can maintain the freshness and relevance of its domain-specific knowledge \cite{dynamic,keepft}.

However, continuous fine-tuning with large-scale data input may lead to catastrophic forgetting, causing the model to forget previous training data and resulting in performance lower than that of the pre-trained model. To prevent disaster, for example, \cite{reviewdisaster} proposed methods to retain old knowledge and conduct regular reviews. \cite{segmentreview} introduces a technique, EMT (Evaluating MulTimodality), that leverages MLLMs as image classifiers to mitigate the situation and improve the alignment performance between text and visual data to maintain the performance. When maintaining the timeliness of the model's knowledge, the related issues that arise must also be addressed.

\begin{figure}[H]
    \centering
    \includegraphics[width=0.48\textwidth, keepaspectratio]{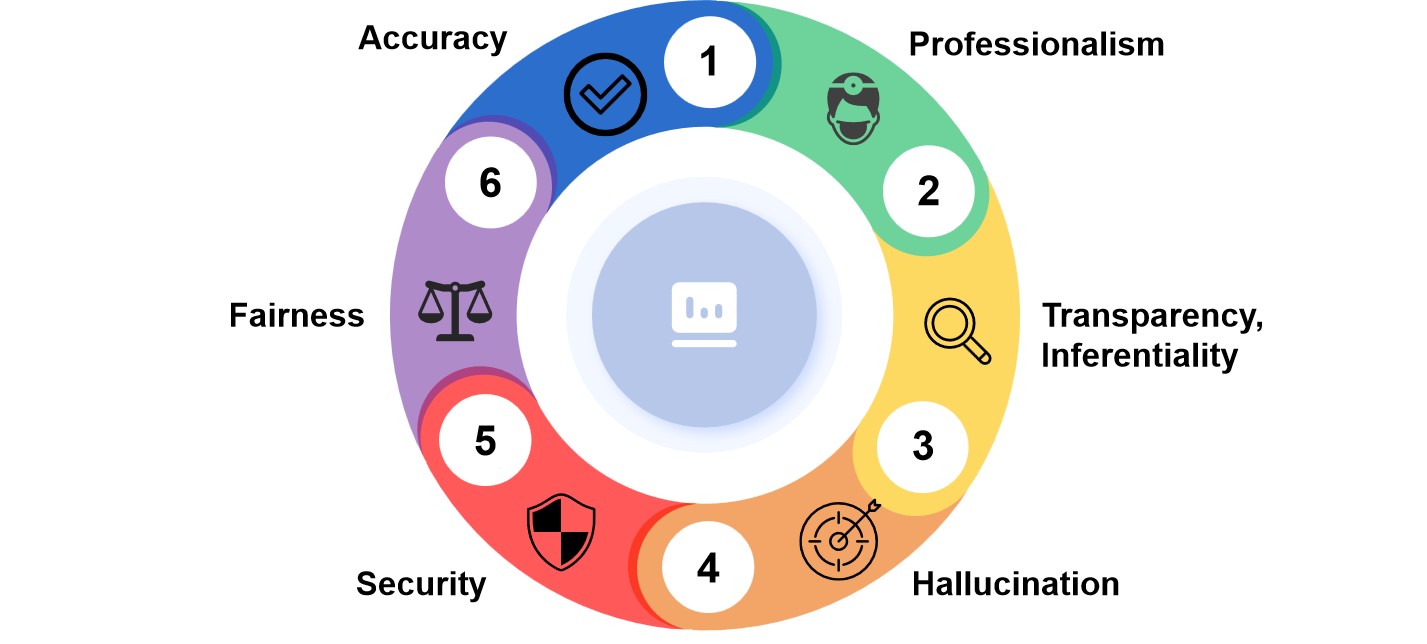} 
    \caption{Ideal Benchmarking Guidelines for Assessing Key Attributes of Large Medical Models.}    
    \label{fig:traits}
    \vspace{-0.4cm}
\end{figure}

\subsection{Deployment}
Medical MLLMs, with their automatic and low-cost characteristics, have been considered a solution to address concerns over the scarcity of medical resources. Ironically, regions with a scarcity of medical resources often also lack technical resources. This means that the deployment of such models may be hindered by the limited computational resources available. The demand for high-quality networks to ensure timely clinical applications also results in high costs. These requirements result in regions with the greatest need being unable to access the necessary resources.

To popularize the deployment of medical MLLMs, instead of deploying on high-computational devices, edge deployment has become the emerging solution. Edge devices such as smartphones, smart wearables, IoT devices, and others, typically have smaller computational capabilities, limited storage capacity, and lower bandwidth.
In order to run the model on these edge devices, methods have been proposed to reduce the model parameters, addressing the high computational consumption and the large memory demands \cite{minimize1,minimize2,mobilevlm,xmodelllava}. At the same time, a Lightweight Downsampling Projector can be leveraged to further optimize the model's efficiency. However, under resource constraints, it should be noted that this can lead to a sharp decline in performance. \cite{mobilevlm} and \cite{xmodelllava} ensure the quality of alignment in various modalities, maintaining strong performance in edge devices with an acceptable loss of information.

Highly effective parameter update techniques can be leveraged to support edge deployment, achieving performance comparable to full parameter updates. Specifically, LoRA \cite{lora} significantly reduces the number of parameters that need updating. Meanwhile, Q-LoRA \cite{qlora} incorporates quantitative techniques into LoRA to achieve further optimization.

In addition, a lightweight framework can be another research direction. Med-MoE \cite{sixformat} leverages domain-specific ``expert systems'' for model optimization by selectively activating particular experts to handle different input modalities, thus reducing the use of model parameters.

These techniques and formats still have considerable room for growth. We anticipate more effective and extensive solutions that will allow medical MLLMs to be used in a wider range of environments.

Efficiently mining and analyzing these data can serve as the foundation for model training to understand healthcare features. By combining personalized and dynamic data, medical research can progress further and drive the development of precise, individualized medical strategies.

\subsection{Privacy and Security}
Due to the legislation of related laws, privacy concerns are the main reason why the quantity of medical data available for training does not meet the requirements \cite{legislation1,legislation2,legislation3,attackexpose1}. Besides the fact of illness, the data may also contain personal information. For example, conversations between experts and patients may include private and sensitive details.
It has been found that adversarial attacks and indications can be leveraged to extract or infer personal information from LLMs \cite{attackexpose1,identify1,inferidentity2,extract1}. Specifically, even with limited background information, privacy-sensitive details can be inferred and extracted by models \cite{netflixidentify}.

The response measure is that medical data is typically anonymized or personal information is removed before being used to prevent intrusions into data privacy \cite{chatdoctor,monotonous2,delete2}. Several strategies have also been suggested.

Differential privacy introduces noise to the training data, blurring personal information and effectively minimizing the risk of privacy leakage \cite{chafen1,chafen2,chafen3}.Federated learning enables MLLMs to be trained on distributed data, without the need to transmit personal information to a central server, thus ensuring data privacy and security \cite{algorism2} \cite{lianbang}.Homomorphic encryption allows computations on encrypted data, enabling model weights to be shared without leaking any sensitive information \cite{samecode}.

However, research indicates that the larger the language model, the higher the likelihood of information leakage. The practice of leveraging large-scale internet data to train MLLMs makes this issue not only inevitable but also irreversible \cite{bigmistake1}. To mitigate the problem as much as possible, the corresponding medical service providers must establish robust data regulations and policies, strictly adhering to the relevant legislation.

\section{Conclusion}
The rise of MLLMs has brought numerous possibilities to artificial intelligence applications. In this paper, we catalog MLLM applications, data modalities, essential model characteristics, and future work, conducting a comprehensive review of medical MLLMs. The potential of MLLMs in medical settings has been demonstrated, and their capabilities are promising. Before full clinical implementation, in addition to enhancing the performance of models, the corresponding procedures must be strictly implemented. A comprehensive assessment benchmark (as shown in Figure~\ref{fig:traits}) and corresponding legislation urgently need to be formulated to enforce stringent oversight of models. We expect that the full potential of MLLMs can be realized once these obstacles are overcome.

{\small
\bibliographystyle{IEEEtran}
\bibliography{ref}

\begin{thebibliography}{100}
\providecommand{\url}[1]{#1}
\csname url@samestyle\endcsname
\providecommand{\newblock}{\relax}
\providecommand{\bibinfo}[2]{#2}
\providecommand{\BIBentrySTDinterwordspacing}{\spaceskip=0pt\relax}
\providecommand{\BIBentryALTinterwordstretchfactor}{4}
\providecommand{\BIBentryALTinterwordspacing}{\spaceskip=\fontdimen2\font plus
\BIBentryALTinterwordstretchfactor\fontdimen3\font minus \fontdimen4\font\relax}
\providecommand{\BIBforeignlanguage}[2]{{%
\expandafter\ifx\csname l@#1\endcsname\relax
\typeout{** WARNING: IEEEtran.bst: No hyphenation pattern has been}%
\typeout{** loaded for the language `#1'. Using the pattern for}%
\typeout{** the default language instead.}%
\else
\language=\csname l@#1\endcsname
\fi
#2}}
\providecommand{\BIBdecl}{\relax}
\BIBdecl

\bibitem{1}
A.~Vaswani, ``Attention is all you need,'' \emph{NeurIPS}, 2017.

\bibitem{2}
J.~Devlin, ``Bert: Pre-training of deep bidirectional transformers for language understanding,'' \emph{arXiv preprint arXiv:1810.04805}, 2018.

\bibitem{3}
H.~W. Chung, L.~Hou, S.~Longpre, B.~Zoph, Y.~Tay, W.~Fedus, Y.~Li, X.~Wang, M.~Dehghani, S.~Brahma \emph{et~al.}, ``Scaling instruction-finetuned language models,'' \emph{Journal of Machine Learning Research}, vol.~25, no.~70, pp. 1--53, 2024.

\bibitem{4}
W.-L. Chiang, Z.~Li, Z.~Lin, Y.~Sheng, Z.~Wu, H.~Zhang, L.~Zheng, S.~Zhuang, Y.~Zhuang, J.~E. Gonzalez \emph{et~al.}, ``Vicuna: An open-source chatbot impressing gpt-4 with 90\%* chatgpt quality,'' \emph{See https://vicuna. lmsys. org (accessed 14 April 2023)}, vol.~2, no.~3, p.~6, 2023.

\bibitem{5}
H.~Touvron, T.~Lavril, G.~Izacard, X.~Martinet, M.-A. Lachaux, T.~Lacroix, B.~Rozi{\`e}re, N.~Goyal, E.~Hambro, F.~Azhar \emph{et~al.}, ``Llama: Open and efficient foundation language models,'' \emph{arXiv preprint arXiv:2302.13971}, 2023.

\bibitem{6}
A.~Radford, J.~W. Kim, C.~Hallacy, A.~Ramesh, G.~Goh, S.~Agarwal, G.~Sastry, A.~Askell, P.~Mishkin, J.~Clark \emph{et~al.}, ``Learning transferable visual models from natural language supervision,'' in \emph{International conference on machine learning}.\hskip 1em plus 0.5em minus 0.4em\relax PMLR, 2021, pp. 8748--8763.

\bibitem{7}
J.~Li, D.~Li, C.~Xiong, and S.~Hoi, ``Blip: Bootstrapping language-image pre-training for unified vision-language understanding and generation,'' in \emph{International conference on machine learning}.\hskip 1em plus 0.5em minus 0.4em\relax PMLR, 2022, pp. 12\,888--12\,900.

\bibitem{8}
J.~Li, D.~Li, S.~Savarese, and S.~Hoi, ``Blip-2: Bootstrapping language-image pre-training with frozen image encoders and large language models,'' in \emph{International conference on machine learning}.\hskip 1em plus 0.5em minus 0.4em\relax PMLR, 2023, pp. 19\,730--19\,742.

\bibitem{9}
J.-B. Alayrac, J.~Donahue, P.~Luc, A.~Miech, I.~Barr, Y.~Hasson, K.~Lenc, A.~Mensch, K.~Millican, M.~Reynolds \emph{et~al.}, ``Flamingo: a visual language model for few-shot learning,'' \emph{NeurIPS}, 2022.

\bibitem{hmmllm}
L.~R. Bahl, F.~Jelinek, and R.~L. Mercer, ``A maximum likelihood approach to continuous speech recognition,'' \emph{IEEE transactions on pattern analysis and machine intelligence}, no.~2, pp. 179--190, 1983.

\bibitem{ngramllm}
F.~Jelinek, ``Interpolated estimation of markov source parameters from sparse data,'' in \emph{Proc. Workshop on Pattern Recognition in Practice, 1980}, 1980.

\bibitem{14}
T.~Mikolov, M.~Karafi{\'a}t, L.~Burget, J.~Cernock{\`y}, and S.~Khudanpur, ``Recurrent neural network based language model.'' in \emph{Interspeech}, vol.~2, no.~3.\hskip 1em plus 0.5em minus 0.4em\relax Makuhari, 2010, pp. 1045--1048.

\bibitem{gpt3}
T.~B. Brown, ``Language models are few-shot learners,'' \emph{arXiv preprint arXiv:2005.14165}, 2020.

\bibitem{palm}
A.~Chowdhery, S.~Narang, J.~Devlin, M.~Bosma, G.~Mishra, A.~Roberts, P.~Barham, H.~W. Chung, C.~Sutton, S.~Gehrmann \emph{et~al.}, ``Palm: Scaling language modeling with pathways,'' \emph{Journal of Machine Learning Research}, vol.~24, no. 240, pp. 1--113, 2023.

\bibitem{gpt4}
J.~Achiam, S.~Adler, S.~Agarwal, L.~Ahmad, I.~Akkaya, F.~L. Aleman, D.~Almeida, J.~Altenschmidt, S.~Altman, S.~Anadkat \emph{et~al.}, ``Gpt-4 technical report,'' \emph{arXiv preprint arXiv:2303.08774}, 2023.

\bibitem{mm1}
B.~McKinzie, Z.~Gan, J.-P. Fauconnier, S.~Dodge, B.~Zhang, P.~Dufter, D.~Shah, X.~Du, F.~Peng, A.~Belyi \emph{et~al.}, ``Mm1: methods, analysis and insights from multimodal llm pre-training,'' in \emph{European Conference on Computer Vision}.\hskip 1em plus 0.5em minus 0.4em\relax Springer, 2025, pp. 304--323.

\bibitem{medpalm}
K.~Singhal, S.~Azizi, T.~Tu, S.~S. Mahdavi, J.~Wei, H.~W. Chung, N.~Scales, A.~Tanwani, H.~Cole-Lewis, S.~Pfohl \emph{et~al.}, ``Large language models encode clinical knowledge,'' \emph{Nature}, vol. 620, no. 7972, pp. 172--180, 2023.

\bibitem{chatdoctor}
Y.~Li, Z.~Li, K.~Zhang, R.~Dan, S.~Jiang, and Y.~Zhang, ``Chatdoctor: A medical chat model fine-tuned on a large language model meta-ai (llama) using medical domain knowledge,'' \emph{Cureus}, vol.~15, no.~6, 2023.

\bibitem{huatuo}
J.~Chen, C.~Gui, R.~Ouyang, A.~Gao, S.~Chen, G.~H. Chen, X.~Wang, R.~Zhang, Z.~Cai, K.~Ji \emph{et~al.}, ``Huatuogpt-vision, towards injecting medical visual knowledge into multimodal llms at scale,'' \emph{arXiv preprint arXiv:2406.19280}, 2024.

\bibitem{amie}
T.~Tu, A.~Palepu, M.~Schaekermann, K.~Saab, J.~Freyberg, R.~Tanno, A.~Wang, B.~Li, M.~Amin, N.~Tomasev \emph{et~al.}, ``Towards conversational diagnostic ai,'' \emph{arXiv preprint arXiv:2401.05654}, 2024.

\bibitem{medflamingo}
M.~Moor, Q.~Huang, S.~Wu, M.~Yasunaga, Y.~Dalmia, J.~Leskovec, C.~Zakka, E.~P. Reis, and P.~Rajpurkar, ``Med-flamingo: a multimodal medical few-shot learner,'' in \emph{Machine Learning for Health (ML4H)}.\hskip 1em plus 0.5em minus 0.4em\relax PMLR, 2023, pp. 353--367.

\bibitem{llavamed}
C.~Li, C.~Wong, S.~Zhang, N.~Usuyama, H.~Liu, J.~Yang, T.~Naumann, H.~Poon, and J.~Gao, ``Llava-med: Training a large language-and-vision assistant for biomedicine in one day,'' in \emph{NeurIPS}, 2024.

\bibitem{medpalmm}
T.~Tu, S.~Azizi, D.~Driess, M.~Schaekermann, M.~Amin, P.-C. Chang, A.~Carroll, C.~Lau, R.~Tanno, I.~Ktena \emph{et~al.}, ``Towards generalist biomedical ai,'' \emph{NEJM AI}, vol.~1, no.~3, p. AIoa2300138, 2024.

\bibitem{skingpt}
J.~Zhou, X.~He, L.~Sun, J.~Xu, X.~Chen, Y.~Chu, L.~Zhou, X.~Liao, B.~Zhang, and X.~Gao, ``Skingpt-4: an interactive dermatology diagnostic system with visual large language model,'' \emph{arXiv preprint arXiv:2304.10691}, 2023.

\bibitem{medmllm}
F.~Liu, T.~Zhu, X.~Wu, B.~Yang, C.~You, C.~Wang, L.~Lu, Z.~Liu, Y.~Zheng, X.~Sun \emph{et~al.}, ``A medical multimodal large language model for future pandemics,'' \emph{NPJ Digital Medicine}, vol.~6, no.~1, p. 226, 2023.

\bibitem{umls}
O.~Bodenreider, ``The unified medical language system (umls): integrating biomedical terminology,'' \emph{Nucleic acids research}, vol.~32, no. suppl\_1, pp. D267--D270, 2004.

\bibitem{ab1}
M.~Wornow, Y.~Xu, R.~Thapa, B.~Patel, E.~Steinberg, S.~Fleming, M.~A. Pfeffer, J.~Fries, and N.~H. Shah, ``The shaky foundations of large language models and foundation models for electronic health records,'' \emph{npj Digital Medicine}, vol.~6, no.~1, p. 135, 2023.

\bibitem{ab2}
A.~J. Thirunavukarasu, D.~S.~J. Ting, K.~Elangovan, L.~Gutierrez, T.~F. Tan, and D.~S.~W. Ting, ``Large language models in medicine,'' \emph{Nature medicine}, vol.~29, no.~8, pp. 1930--1940, 2023.

\bibitem{ab3}
D.~Van~Veen, C.~Van~Uden, L.~Blankemeier, J.-B. Delbrouck, A.~Aali, C.~Bluethgen, A.~Pareek, M.~Polacin, E.~P. Reis, A.~Seehofnerova \emph{et~al.}, ``Clinical text summarization: adapting large language models can outperform human experts,'' \emph{Research Square}, 2023.

\bibitem{xraygpt}
O.~C. Thawakar, A.~M. Shaker, S.~S. Mullappilly, H.~Cholakkal, R.~M. Anwer, S.~Khan, J.~Laaksonen, and F.~Khan, ``Xraygpt: Chest radiographs summarization using large medical vision-language models,'' in \emph{Proceedings of the 23rd workshop on biomedical natural language processing}, 2024, pp. 440--448.

\bibitem{ct1}
Z.~Chen, L.~Luo, Y.~Bie, and H.~Chen, ``Dia-llama: Towards large language model-driven ct report generation,'' \emph{arXiv preprint arXiv:2403.16386}, 2024.

\bibitem{ct2}
{\c{S}}.-V. Voinea, M.~M{\u{a}}muleanu, R.~V. Teic{\u{a}}, L.~M. Florescu, D.~Seli{\c{s}}teanu, and I.~A. Gheonea, ``Gpt-driven radiology report generation with fine-tuned llama 3,'' \emph{Bioengineering}, vol.~11, no.~10, p. 1043, 2024.

\bibitem{brain}
J.~Lei, X.~Zhang, C.~Wu, L.~Dai, Y.~Zhang, Y.~Zhang, Y.~Wang, W.~Xie, and Y.~Li, ``Autorg-brain: Grounded report generation for brain mri,'' \emph{arXiv preprint arXiv:2407.16684}, 2024.

\bibitem{3d}
C.~Wu, X.~Zhang, Y.~Zhang, Y.~Wang, and W.~Xie, ``Towards generalist foundation model for radiology,'' \emph{arXiv preprint arXiv:2308.02463}, 2023.

\bibitem{noorder}
A.~Gupta, H.~Malhotra, A.~K. Garg, and K.~Rangarajan, ``Enhancing radiological reporting in head and neck cancer: Converting free-text ct scan reports to structured reports using large language models,'' \emph{Indian Journal of Radiology and Imaging}, 2024.

\bibitem{threespot}
C.~Wu, X.~Zhang, Y.~Zhang, Y.~Wang, and W.~Xie, ``Medklip: Medical knowledge enhanced language-image pre-training in radiology,'' \emph{arXiv preprint arXiv:2301.02228}, 2023.

\bibitem{gpt4poor}
S.~Senkaiahliyan, A.~Toma, J.~Ma, A.-W. Chan, A.~Ha, K.~R. An, H.~Suresh, B.~Rubin, and B.~Wang, ``Gpt-4v (ision) unsuitable for clinical care and education: a clinician-evaluated assessment,'' \emph{medRxiv}, pp. 2023--11, 2023.

\bibitem{gptformat}
J.~Wang, Y.~Ting, E.~Z. Chen, H.~Tran, H.~Yu, W.~Huang, and T.~Chen, ``Semihvision: Enhancing medical multimodal models with a semi-human annotated dataset and fine-tuned instruction generation,'' \emph{arXiv preprint arXiv:2410.14948}, 2024.

\bibitem{findingimpression}
C.~Liu, S.~Cheng, M.~Shi, A.~Shah, W.~Bai, and R.~Arcucci, ``Imitate: Clinical prior guided hierarchical vision-language pre-training,'' \emph{IEEE TMI}, 2024.

\bibitem{structure}
A.~Wallis and P.~McCoubrie, ``The radiology report—are we getting the message across?'' \emph{Clinical radiology}, vol.~66, no.~11, pp. 1015--1022, 2011.

\bibitem{chatcad}
S.~Wang, Z.~Zhao, X.~Ouyang, Q.~Wang, and D.~Shen, ``Chatcad: Interactive computer-aided diagnosis on medical image using large language models,'' \emph{arXiv preprint arXiv:2302.07257}, 2023.

\bibitem{dictionary}
B.~Boecking, N.~Usuyama, S.~Bannur, D.~C. Castro, A.~Schwaighofer, S.~Hyland, M.~Wetscherek, T.~Naumann, A.~Nori, J.~Alvarez-Valle \emph{et~al.}, ``Making the most of text semantics to improve biomedical vision--language processing,'' in \emph{ECCV}, 2022.

\bibitem{record1}
Y.~Kumar, A.~Koul, and S.~Mahajan, ``A deep learning approaches and fastai text classification to predict 25 medical diseases from medical speech utterances, transcription and intent,'' \emph{Springer Soft computing}, vol.~26, no.~17, pp. 8253--8272, 2022.

\bibitem{record2}
F.~S. Falcetta, F.~K. De~Almeida, J.~C.~S. Lemos, J.~R. Goldim, and C.~A. Da~Costa, ``Automatic documentation of professional health interactions: A systematic review,'' \emph{Elsevier Artificial Intelligence in Medicine}, vol. 137, p. 102487, 2023.

\bibitem{segment1}
R.~Zhao, X.~Wang, H.~Dai, P.~Gao, and P.~Li, ``Medical report generation based on segment-enhanced contrastive representation learning,'' in \emph{CCF International Conference on Natural Language Processing and Chinese Computing}.\hskip 1em plus 0.5em minus 0.4em\relax Springer, 2023, pp. 838--849.

\bibitem{segment2}
H.~Yang, T.~Zhou, Y.~Zhou, Y.~Zhang, and H.~Fu, ``Flexible fusion network for multi-modal brain tumor segmentation,'' \emph{IEEE Journal of Biomedical and Health Informatics}, vol.~27, no.~7, pp. 3349--3359, 2023.

\bibitem{meddialog1}
W.~Gao, Z.~Deng, Z.~Niu, F.~Rong, C.~Chen, Z.~Gong, W.~Zhang, D.~Xiao, F.~Li, Z.~Cao \emph{et~al.}, ``Ophglm: Training an ophthalmology large language-and-vision assistant based on instructions and dialogue,'' \emph{arXiv preprint arXiv:2306.12174}, 2023.

\bibitem{chatglm}
T.~GLM, A.~Zeng, B.~Xu, B.~Wang, C.~Zhang, D.~Yin, D.~Zhang, D.~Rojas, G.~Feng, H.~Zhao \emph{et~al.}, ``Chatglm: A family of large language models from glm-130b to glm-4 all tools,'' \emph{arXiv preprint arXiv:2406.12793}, 2024.

\bibitem{radiologyllama2}
Z.~Liu, Y.~Li, P.~Shu, A.~Zhong, L.~Yang, C.~Ju, Z.~Wu, C.~Ma, J.~Luo, C.~Chen \emph{et~al.}, ``Radiology-llama2: Best-in-class large language model for radiology,'' \emph{arXiv preprint arXiv:2309.06419}, 2023.

\bibitem{llama2}
H.~Touvron, L.~Martin, K.~Stone, P.~Albert, A.~Almahairi, Y.~Babaei, N.~Bashlykov, S.~Batra, P.~Bhargava, S.~Bhosale \emph{et~al.}, ``Llama 2: Open foundation and fine-tuned chat models,'' \emph{arXiv preprint arXiv:2307.09288}, 2023.

\bibitem{qilinmedvl}
J.~Liu, Z.~Wang, Q.~Ye, D.~Chong, P.~Zhou, and Y.~Hua, ``Qilin-med-vl: Towards chinese large vision-language model for general healthcare,'' \emph{arXiv preprint arXiv:2310.17956}, 2023.

\bibitem{sigphimed}
F.~Zhou, X.~Liu, Q.~Zeng, Z.~Li, and H.~Xiao, ``Sigphi-med: A lightweight vision-language assistant for biomedicine,'' \emph{Available at SSRN 4988925}, 2024.

\bibitem{phi2}
M.~Javaheripi, S.~Bubeck, M.~Abdin, J.~Aneja, S.~Bubeck, C.~C.~T. Mendes, W.~Chen, A.~Del~Giorno, R.~Eldan, S.~Gopi \emph{et~al.}, ``Phi-2: The surprising power of small language models,'' \emph{Microsoft Research Blog}, vol.~1, no.~3, p.~3, 2023.

\bibitem{chatdavinci}
L.~Ouyang, J.~Wu, X.~Jiang, D.~Almeida, C.~Wainwright, P.~Mishkin, C.~Zhang, S.~Agarwal, K.~Slama, A.~Ray \emph{et~al.}, ``Training language models to follow instructions with human feedback,'' \emph{Advances in neural information processing systems}, vol.~35, pp. 27\,730--27\,744, 2022.

\bibitem{medblip}
Q.~Chen and Y.~Hong, ``Medblip: Bootstrapping language-image pre-training from 3d medical images and texts,'' in \emph{Proceedings of the Asian Conference on Computer Vision}, 2024, pp. 2404--2420.

\bibitem{biomedlm}
E.~Bolton, A.~Venigalla, M.~Yasunaga, D.~Hall, B.~Xiong, T.~Lee, R.~Daneshjou, J.~Frankle, P.~Liang, M.~Carbin \emph{et~al.}, ``Biomedlm: A 2.7 b parameter language model trained on biomedical text,'' \emph{arXiv preprint arXiv:2403.18421}, 2024.

\bibitem{pclmed}
B.~Yang, A.~Raza, Y.~Zou, and T.~Zhang, ``Pclmed at imageclefmedical 2023: Customizing general-purpose foundation models for medical report generation.'' in \emph{CLEF (Working Notes)}, 2023, pp. 1754--1766.

\bibitem{pmcvqa}
X.~Zhang, C.~Wu, Z.~Zhao, W.~Lin, Y.~Zhang, Y.~Wang, and W.~Xie, ``Pmc-vqa: Visual instruction tuning for medical visual question answering,'' \emph{arXiv preprint arXiv:2305.10415}, 2023.

\bibitem{pmcllama}
C.~Wu, W.~Lin, X.~Zhang, Y.~Zhang, W.~Xie, and Y.~Wang, ``Pmc-llama: toward building open-source language models for medicine,'' \emph{Journal of the American Medical Informatics Association}, p. ocae045, 2024.

\bibitem{pathasst}
Y.~Sun, C.~Zhu, S.~Zheng, K.~Zhang, L.~Sun, Z.~Shui, Y.~Zhang, H.~Li, and L.~Yang, ``Pathasst: A generative foundation ai assistant towards artificial general intelligence of pathology,'' in \emph{Proceedings of the AAAI Conference on Artificial Intelligence}, vol.~38, no.~5, 2024, pp. 5034--5042.

\bibitem{chatcadplus}
Z.~Zhao, S.~Wang, J.~Gu, Y.~Zhu, L.~Mei, Z.~Zhuang, Z.~Cui, Q.~Wang, and D.~Shen, ``Chatcad+: Towards a universal and reliable interactive cad using llms,'' \emph{IEEE TMI}, 2024.

\bibitem{chatgpt}
{OpenAI}, ``Chatgpt: Optimizing language models for dialogue,'' \url{https://openai.com/blog/chatgpt/}, 2023, accessed: 2023-01-08.

\bibitem{radfm}
C.~Wu, X.~Zhang, Y.~Zhang, Y.~Wang, and W.~Xie, ``Towards generalist foundation model for radiology,'' \emph{arXiv preprint arXiv:2308.02463}, 2023.

\bibitem{r2gengpt}
Z.~Wang, L.~Liu, L.~Wang, and L.~Zhou, ``R2gengpt: Radiology report generation with frozen llms,'' \emph{Meta-Radiology}, vol.~1, no.~3, p. 100033, 2023.

\bibitem{cxrllava}
S.~Lee, J.~Youn, H.~Kim, M.~Kim, and S.~H. Yoon, ``Cxr-llava: A multimodal large language model for interpreting chest x-ray images,'' \emph{European Radiology}, pp. 1--13, 2025.

\bibitem{maira1}
S.~L. Hyland, S.~Bannur, K.~Bouzid, D.~C. Castro, M.~Ranjit, A.~Schwaighofer, F.~P{\'e}rez-Garc{\'\i}a, V.~Salvatelli, S.~Srivastav, A.~Thieme \emph{et~al.}, ``Maira-1: A specialised large multimodal model for radiology report generation,'' \emph{arXiv preprint arXiv:2311.13668}, 2023.

\bibitem{pefomed}
G.~Liu, J.~He, P.~Li, G.~He, Z.~Chen, and S.~Zhong, ``Pefomed: Parameter efficient fine-tuning of multimodal large language models for medical imaging,'' \emph{arXiv preprint arXiv:2401.02797}, 2024.

\bibitem{minigptv2}
J.~Chen, D.~Zhu, X.~Shen, X.~Li, Z.~Liu, P.~Zhang, R.~Krishnamoorthi, V.~Chandra, Y.~Xiong, and M.~Elhoseiny, ``Minigpt-v2: large language model as a unified interface for vision-language multi-task learning,'' \emph{arXiv preprint arXiv:2310.09478}, 2023.

\bibitem{m3d}
F.~Bai, Y.~Du, T.~Huang, M.~Q.-H. Meng, and B.~Zhao, ``M3d: Advancing 3d medical image analysis with multi-modal large language models,'' \emph{arXiv preprint arXiv:2404.00578}, 2024.

\bibitem{moetinymed}
S.~Jiang, T.~Zheng, Y.~Zhang, Y.~Jin, and Z.~Liu, ``Moe-tinymed: Mixture of experts for tiny medical large vision-language models,'' \emph{arXiv preprint arXiv:2404.10237}, 2024.

\bibitem{ctreport}
Z.~Chen, L.~Luo, Y.~Bie, and H.~Chen, ``Dia-llama: Towards large language model-driven ct report generation,'' \emph{arXiv preprint arXiv:2403.16386}, 2024.

\bibitem{maira2}
S.~Bannur, K.~Bouzid, D.~C. Castro, A.~Schwaighofer, A.~Thieme, S.~Bond-Taylor, M.~Ilse, F.~P{\'e}rez-Garc{\'\i}a, V.~Salvatelli, H.~Sharma \emph{et~al.}, ``Maira-2: Grounded radiology report generation,'' \emph{arXiv preprint arXiv:2406.04449}, 2024.

\bibitem{minigptmed}
A.~Alkhaldi, R.~Alnajim, L.~Alabdullatef, R.~Alyahya, J.~Chen, D.~Zhu, A.~Alsinan, and M.~Elhoseiny, ``Minigpt-med: Large language model as a general interface for radiology diagnosis,'' \emph{arXiv preprint arXiv:2407.04106}, 2024.

\bibitem{surgicalapp1}
J.~Li, G.~Skinner, G.~Yang, B.~R. Quaranto, S.~D. Schwaitzberg, P.~C. Kim, and J.~Xiong, ``Llava-surg: Towards multimodal surgical assistant via structured surgical video learning,'' \emph{arXiv preprint arXiv:2408.07981}, 2024.

\bibitem{llama3}
A.~Dubey, A.~Jauhri, A.~Pandey, A.~Kadian, A.~Al-Dahle, A.~Letman, A.~Mathur, A.~Schelten, A.~Yang, A.~Fan \emph{et~al.}, ``The llama 3 herd of models,'' \emph{arXiv preprint arXiv:2407.21783}, 2024.

\bibitem{histaid}
H.~Huang, C.~M. Deniz, K.~Cho, S.~Chopra, and D.~Madaan, ``Hist-aid: Leveraging historical patient reports for enhanced multi-modal automatic diagnosis,'' \emph{arXiv preprint arXiv:2411.10684}, 2024.

\bibitem{aquila2}
B.-W. Zhang, L.~Wang, J.~Li, S.~Gu, X.~Wu, Z.~Zhang, B.~Gao, Y.~Ao, and G.~Liu, ``Aquila2 technical report,'' \emph{arXiv preprint arXiv:2408.07410}, 2024.

\bibitem{visionbiollm}
A.~AlShibli, Y.~Bazi, M.~M. Al~Rahhal, and M.~Zuair, ``Vision-biollm: Large vision language model for visual dialogue in biomedical imagery,'' \emph{Biomedical Signal Processing and Control}, vol. 103, p. 107437, 2025.

\bibitem{openbiollms}
A.~Pal and M.~Sankarasubbu, ``Openbiollms: Advancing open-source large language models for healthcare and life sciences,'' 2024.

\bibitem{baize}
C.~Xu, D.~Guo, N.~Duan, and J.~McAuley, ``Baize: An open-source chat model with parameter-efficient tuning on self-chat data,'' \emph{arXiv preprint arXiv:2304.01196}, 2023.

\bibitem{gpt4vpoor}
S.~Senkaiahliyan, A.~Toma, J.~Ma, A.-W. Chan, A.~Ha, K.~R. An, H.~Suresh, B.~Rubin, and B.~Wang, ``Gpt-4v (ision) unsuitable for clinical care and education: a clinician-evaluated assessment,'' \emph{medRxiv}, pp. 2023--11, 2023.

\bibitem{facetoface}
T.~Nadarzynski, J.~Bayley, C.~Llewellyn, S.~Kidsley, and C.~A. Graham, ``Acceptability of artificial intelligence (ai)-enabled chatbots, video consultations and live webchats as online platforms for sexual health advice,'' \emph{BMJ sexual \& reproductive health}, vol.~46, no.~3, pp. 210--217, 2020.

\bibitem{preferhuman}
S.~Hussain, M.~Alherz, E.~Albazee, H.~Almhanedi, J.~Hayat, M.~Lari, and A.~Lari, ``Investigating public perception on use of chatgpt in initial consultations prior to healthcare provider consultations,'' \emph{Annals of Medicine and Surgery}, pp. 10--1097, 2024.

\bibitem{mentalhealth1}
K.~Olga and Z.~Xuehan, ``Mental health problems of contemporary youth,'' \emph{Human Health (Zdorov'e cheloveka), Theory and Methodology of Physical Culture and Sports}, no. 4 (15), pp. 45--49, 2019, in Russian.

\bibitem{mentalhealth2}
M.~Prince, V.~Patel, S.~Saxena, M.~Maj, J.~Maselko, M.~R. Phillips, and A.~Rahman, ``No health without mental health,'' \emph{The lancet}, vol. 370, no. 9590, pp. 859--877, 2007.

\bibitem{consel1}
F.~Vescovelli, P.~Melani, C.~Ruini, P.~E. Ricci~Bitti, and F.~Monti, ``University counseling service for improving students’ mental health.'' \emph{Psychological services}, vol.~14, no.~4, p. 470, 2017.

\bibitem{chatboteffect1}
A.~Schick, J.~Feine, S.~Morana, A.~Maedche, and U.~Reininghaus, ``Validity of chatbot use for mental health assessment: experimental study,'' \emph{JMIR mHealth and uHealth}, vol.~10, no.~10, p. e28082, 2022.

\bibitem{chatboteffect2}
A.~A. Abd-Alrazaq, A.~Rababeh, M.~Alajlani, B.~M. Bewick, and M.~Househ, ``Effectiveness and safety of using chatbots to improve mental health: systematic review and meta-analysis,'' \emph{Journal of medical Internet research}, vol.~22, no.~7, p. e16021, 2020.

\bibitem{patientrelax1}
A.~Stock, S.~Schl{\"o}gl, and A.~Groth, ``Tell me, what are you most afraid of? exploring the effects of agent representation on information disclosure in human-chatbot interaction,'' in \emph{International Conference on Human-Computer Interaction}.\hskip 1em plus 0.5em minus 0.4em\relax Springer, 2023, pp. 179--191.

\bibitem{patientrelax2}
A.~P. Chaves and M.~A. Gerosa, ``How should my chatbot interact? a survey on social characteristics in human--chatbot interaction design,'' \emph{International Journal of Human--Computer Interaction}, vol.~37, no.~8, pp. 729--758, 2021.

\bibitem{llmchatbot1}
Y.~Chen, X.~Xing, J.~Lin, H.~Zheng, Z.~Wang, Q.~Liu, and X.~Xu, ``Soulchat: Improving llms’ empathy, listening, and comfort abilities through fine-tuning with multi-turn empathy conversations,'' in \emph{EMNLP}, 2023.

\bibitem{llmchatbot2}
H.~Qiu, A.~Li, L.~Ma, and Z.~Lan, ``Psychat: A client-centric dialogue system for mental health support,'' in \emph{CSCWD}, 2024.

\bibitem{llmchatbot3}
H.~Qiu, H.~He, S.~Zhang, A.~Li, and Z.~Lan, ``Smile: Single-turn to multi-turn inclusive language expansion via chatgpt for mental health support,'' \emph{arXiv preprint arXiv:2305.00450}, 2023.

\bibitem{chatevaluate1}
H.~Zhao, L.~Li, S.~Chen, S.~Kong, J.~Wang, K.~Huang, T.~Gu, Y.~Wang, W.~Jian, D.~Liang \emph{et~al.}, ``Esc-eval: Evaluating emotion support conversations in large language models,'' \emph{arXiv preprint arXiv:2406.14952}, 2024.

\bibitem{chatevaluate2}
C.~Zhang, R.~Li, M.~Tan, M.~Yang, J.~Zhu, D.~Yang, J.~Zhao, G.~Ye, C.~Li, and X.~Hu, ``Cpsycoun: A report-based multi-turn dialogue reconstruction and evaluation framework for chinese psychological counseling,'' \emph{arXiv preprint arXiv:2405.16433}, 2024.

\bibitem{audiosentiment}
Y.~Bhangdia, R.~Bhansali, N.~Chaudhari, D.~Chandnani, and M.~Dhore, ``Speech emotion recognition and sentiment analysis based therapist bot,'' in \emph{ICIRCA}, 2021.

\bibitem{allsentiment}
T.~Dong, F.~Liu, X.~Wang, Y.~Jiang, X.~Zhang, and X.~Sun, ``Emoada: A multimodal emotion interaction and psychological adaptation system,'' in \emph{International Conference on Multimedia Modeling}, 2024.

\bibitem{regulate}
Y.~Zhang, Y.~Pan, T.~Zhong, P.~Dong, K.~Xie, Y.~Liu, H.~Jiang, Z.~Wu, Z.~Liu, W.~Zhao \emph{et~al.}, ``Potential of multimodal large language models for data mining of medical images and free-text reports,'' \emph{Meta-Radiology}, vol.~2, no.~4, p. 100103, 2024.

\bibitem{computerassist1}
L.~Adams, W.~Krybus, D.~Meyer-Ebrecht, R.~Rueger, J.~M. Gilsbach, R.~Moesges, and G.~Schloendorff, ``Computer-assisted surgery,'' \emph{IEEE Computer graphics and applications}, vol.~10, no.~3, pp. 43--51, 1990.

\bibitem{computerassist2}
L.~Adams, J.~M. Gilsbach, W.~Krybus, D.~Meyer-Ebrecht, R.~M{\"o}sges, and G.~Schl{\"o}ndorff, ``Cas—a navigation support for surgery,'' in \emph{3D Imaging in Medicine: Algorithms, Systems, Applications}.\hskip 1em plus 0.5em minus 0.4em\relax Springer, 1990, pp. 411--423.

\bibitem{surgicalvqa}
L.~Seenivasan, M.~Islam, A.~K. Krishna, and H.~Ren, ``Surgical-vqa: Visual question answering in surgical scenes using transformer,'' in \emph{MICCAI}, 2022.

\bibitem{surgicalapp2}
K.~Yuan, V.~Srivastav, T.~Yu, J.~L. Lavanchy, P.~Mascagni, N.~Navab, and N.~Padoy, ``Learning multi-modal representations by watching hundreds of surgical video lectures,'' \emph{arXiv preprint arXiv:2307.15220}, 2023.

\bibitem{surgicalgpt}
L.~Seenivasan, M.~Islam, G.~Kannan, and H.~Ren, ``Surgicalgpt: end-to-end language-vision gpt for visual question answering in surgery,'' in \emph{MICCAI}, 2023.

\bibitem{maylack}
K.~Yuan, M.~Kattel, J.~L. Lavanchy, N.~Navab, V.~Srivastav, and N.~Padoy, ``Advancing surgical vqa with scene graph knowledge,'' \emph{International Journal of Computer Assisted Radiology and Surgery}, pp. 1--9, 2024.

\bibitem{toolgesture1}
L.~Seenivasan, S.~Mitheran, M.~Islam, and H.~Ren, ``Global-reasoned multi-task learning model for surgical scene understanding,'' \emph{IEEE RAL}, vol.~7, no.~2, pp. 3858--3865, 2022.

\bibitem{toolgesture2}
L.~Bai, G.~Wang, M.~Islam, L.~Seenivasan, A.~Wang, and H.~Ren, ``Surgical-vqla++: Adversarial contrastive learning for calibrated robust visual question-localized answering in robotic surgery,'' \emph{Information Fusion}, vol. 113, p. 102602, 2025.

\bibitem{report1}
H.~Wang, Y.~Jin, and L.~Zhu, ``Dynamic interactive relation capturing via scene graph learning for robotic surgical report generation,'' in \emph{ICRA}, 2023.

\bibitem{report2}
C.~Lin, S.~Zheng, Z.~Liu, Y.~Li, Z.~Zhu, and Y.~Zhao, ``Sgt: Scene graph-guided transformer for surgical report generation,'' in \emph{MICCAI}, 2022.

\bibitem{cnnxray1}
K.~Yuan, V.~Srivastav, T.~Yu, J.~L. Lavanchy, P.~Mascagni, N.~Navab, and N.~Padoy, ``Learning multi-modal representations by watching hundreds of surgical video lectures,'' \emph{arXiv preprint arXiv:2307.15220}, 2023.

\bibitem{cnnsegment}
L.~Chen, P.~Bentley, K.~Mori, K.~Misawa, M.~Fujiwara, and D.~Rueckert, ``Drinet for medical image segmentation,'' \emph{IEEE TMI}, vol.~37, no.~11, pp. 2453--2462, 2018.

\bibitem{dldiagnose}
T.~Liu, E.~Siegel, and D.~Shen, ``Deep learning and medical image analysis for covid-19 diagnosis and prediction,'' \emph{Annual review of biomedical engineering}, vol.~24, no.~1, pp. 179--201, 2022.

\bibitem{brainmri}
J.~Lei, X.~Zhang, C.~Wu, L.~Dai, Y.~Zhang, Y.~Zhang, Y.~Wang, W.~Xie, and Y.~Li, ``Autorg-brain: Grounded report generation for brain mri,'' \emph{arXiv preprint arXiv:2407.16684}, 2024.

\bibitem{mribone}
S.~Saran, K.~Shirodkar, S.~Ariyaratne, K.~Iyengar, N.~Jenko, B.~Durgaprasad, and R.~Botchu, ``Exploring chat generated pre-trained transformer-3 ability to interpret mri knee images and generate reports,'' \emph{Journal of Arthroscopic Surgery and Sports Medicine}, vol.~5, no.~2, pp. 75--80, 2024.

\bibitem{zhoutao}
H.~Yang, T.~Zhou, Y.~Zhou, Y.~Zhang, and H.~Fu, ``Flexible fusion network for multi-modal brain tumor segmentation,'' \emph{IEEE Journal of Biomedical and Health Informatics}, vol.~27, no.~7, pp. 3349--3359, 2023.

\bibitem{xraychest1}
S.~Xu, L.~Yang, C.~Kelly, M.~Sieniek, T.~Kohlberger, M.~Ma, W.-H. Weng, A.~Kiraly, S.~Kazemzadeh, Z.~Melamed \emph{et~al.}, ``Elixr: Towards a general purpose x-ray artificial intelligence system through alignment of large language models and radiology vision encoders,'' \emph{arXiv preprint arXiv:2308.01317}, 2023.

\bibitem{xraychest2}
J.~Park, S.~Kim, B.~Yoon, J.~Hyun, and K.~Choi, ``M4cxr: Exploring multi-task potentials of multi-modal large language models for chest x-ray interpretation,'' \emph{arXiv preprint arXiv:2408.16213}, 2024.

\bibitem{bonexray}
Z.~Su, Y.~Zhou, J.~Zhou, H.~Cao, and H.~Zhang, ``Boneclip-xgboost: A multimodal approach for bone fracture diagnosis,'' \emph{IEEE Access}, 2024.

\bibitem{brainct1}
C.~Zheng, J.~Ji, Y.~Shi, X.~Zhang, and L.~Qu, ``See detail say clear: Towards brain ct report generation via pathological clue-driven representation learning,'' \emph{arXiv preprint arXiv:2409.19676}, 2024.

\bibitem{brainct2}
C.-Y. Li, K.-J. Chang, C.-F. Yang, H.-Y. Wu, W.~Chen, H.~Bansal, L.~Chen, Y.-P. Yang, Y.-C. Chen, S.-P. Chen \emph{et~al.}, ``Towards a holistic framework for multimodal large language models in three-dimensional brain ct report generation,'' \emph{arXiv preprint arXiv:2407.02235}, 2024.

\bibitem{bonect}
Y.~Jin and Y.~Zhang, ``Orthodoc: Multimodal large language model for assisting diagnosis in computed tomography,'' \emph{arXiv preprint arXiv:2409.09052}, 2024.

\bibitem{eyephoto}
H.~Zhao, Q.~Ling, Y.~Pan, T.~Zhong, J.-Y. Hu, J.~Yao, F.~Xiao, Z.~Xiao, Y.~Zhang, S.-H. Xu \emph{et~al.}, ``Ophtha-llama2: A large language model for ophthalmology,'' \emph{arXiv preprint arXiv:2312.04906}, 2023.

\bibitem{endoscopy}
E.~J. Gong and C.~S. Bang, ``Revolutionizing gastrointestinal endoscopy: the emerging role of large language models,'' \emph{Clinical Endoscopy}, 2024.

\bibitem{oldconsciousllm}
X.~Du, J.~Novoa-Laurentiev, J.~M. Plasek, Y.-W. Chuang, L.~Wang, G.~A. Marshall, S.~K. Mueller, F.~Chang, S.~Datta, H.~Paek \emph{et~al.}, ``Enhancing early detection of cognitive decline in the elderly: a comparative study utilizing large language models in clinical notes,'' \emph{EBioMedicine}, vol. 109, 2024.

\bibitem{llmsingle2}
M.~Benary, X.~D. Wang, M.~Schmidt, D.~Soll, G.~Hilfenhaus, M.~Nassir, C.~Sigler, M.~Kn{\"o}dler, U.~Keller, D.~Beule \emph{et~al.}, ``Leveraging large language models for decision support in personalized oncology,'' \emph{JAMA Network Open}, vol.~6, no.~11, pp. e2\,343\,689--e2\,343\,689, 2023.

\bibitem{singlellm3}
M.~A. Fink, A.~Bischoff, C.~A. Fink, M.~Moll, J.~Kroschke, L.~Dulz, C.~P. Heu{\ss}el, H.-U. Kauczor, and T.~F. Weber, ``Potential of chatgpt and gpt-4 for data mining of free-text ct reports on lung cancer,'' \emph{Radiology}, vol. 308, no.~3, p. e231362, 2023.

\bibitem{meddialog}
G.~Zeng, W.~Yang, Z.~Ju, Y.~Yang, S.~Wang, R.~Zhang, M.~Zhou, J.~Zeng, X.~Dong, R.~Zhang \emph{et~al.}, ``Meddialog: Large-scale medical dialogue datasets,'' in \emph{EMNLP}, 2020.

\bibitem{medrecommend}
------, ``Meddialog: Large-scale medical dialogue datasets,'' in \emph{EMNLP}, 2020.

\bibitem{vqamodel2}
C.~Wu, X.~Zhang, Y.~Zhang, Y.~Wang, and W.~Xie, ``Towards generalist foundation model for radiology,'' \emph{arXiv preprint arXiv:2308.02463}, 2023.

\bibitem{vqamodel}
T.~Gu, K.~Yang, D.~Liu, and W.~Cai, ``Lapa: Latent prompt assist model for medical visual question answering,'' in \emph{Proceedings of the IEEE/CVF Conference on Computer Vision and Pattern Recognition}, 2024, pp. 4971--4980.

\bibitem{reportimage1}
L.~Xu, H.~Sun, Z.~Ni, H.~Li, and S.~Zhang, ``Medvilam: A multimodal large language model with advanced generalizability and explainability for medical data understanding and generation,'' \emph{arXiv preprint arXiv:2409.19684}, 2024.

\bibitem{reportimage2}
C.~Liu, S.~Cheng, M.~Shi, A.~Shah, W.~Bai, and R.~Arcucci, ``Imitate: Clinical prior guided hierarchical vision-language pre-training,'' \emph{IEEE TMI}, 2024.

\bibitem{reportimage3}
F.~Liu, T.~Zhu, X.~Wu, B.~Yang, C.~You, C.~Wang, L.~Lu, Z.~Liu, Y.~Zheng, X.~Sun \emph{et~al.}, ``A medical multimodal large language model for future pandemics,'' \emph{NPJ Digital Medicine}, vol.~6, no.~1, p. 226, 2023.

\bibitem{historyreportcombine}
H.~Huang, C.~M. Deniz, K.~Cho, S.~Chopra, and D.~Madaan, ``Hist-aid: Leveraging historical patient reports for enhanced multi-modal automatic diagnosis,'' \emph{arXiv preprint arXiv:2411.10684}, 2024.

\bibitem{ehrproof}
L.~Y. Jiang, X.~C. Liu, N.~P. Nejatian, M.~Nasir-Moin, D.~Wang, A.~Abidin, K.~Eaton, H.~A. Riina, I.~Laufer, P.~Punjabi \emph{et~al.}, ``Health system-scale language models are all-purpose prediction engines,'' \emph{Nature}, vol. 619, no. 7969, pp. 357--362, 2023.

\bibitem{audiomulti}
Y.~Zhang, T.~Xia, A.~Saeed, and C.~Mascolo, ``Respllm: Unifying audio and text with multimodal llms for generalized respiratory health prediction,'' \emph{arXiv preprint arXiv:2410.05361}, 2024.

\bibitem{audiorecord1}
F.~S. Falcetta, F.~K. De~Almeida, J.~C.~S. Lemos, J.~R. Goldim, and C.~A. Da~Costa, ``Automatic documentation of professional health interactions: A systematic review,'' \emph{Artificial Intelligence in Medicine}, vol. 137, p. 102487, 2023.

\bibitem{audiorecord2}
J.~T. Anibal, A.~J. Landa, N.~T. Hang, M.~J. Song, A.~K. Peltekian, A.~Shin, H.~B. Huth, L.~A. Hazen, A.~S. Christou, J.~Rivera \emph{et~al.}, ``Omicron detection with large language models and youtube audio data,'' \emph{medRxiv}, 2024.

\bibitem{audioedu}
S.~Jia, S.~Bit, E.~Searls, L.~A. Claus, P.~Fan, V.~H. Jasodanand, M.~V. Lauber, D.~Veerapaneni, W.~M. Wang, R.~Au \emph{et~al.}, ``Medpodgpt: A multilingual audio-augmented large language model for medical research and education,'' \emph{medRxiv}, 2024.

\bibitem{omicsresearch1}
R.~Duan, L.~Gao, Y.~Gao, Y.~Hu, H.~Xu, M.~Huang, K.~Song, H.~Wang, Y.~Dong, C.~Jiang \emph{et~al.}, ``Evaluation and comparison of multi-omics data integration methods for cancer subtyping,'' \emph{PLoS computational biology}, vol.~17, no.~8, p. e1009224, 2021.

\bibitem{omicsresearch2}
------, ``Evaluation and comparison of multi-omics data integration methods for cancer subtyping,'' \emph{PLoS computational biology}, vol.~17, no.~8, p. e1009224, 2021.

\bibitem{omicllm1}
V.~Fishman, Y.~Kuratov, A.~Shmelev, M.~Petrov, D.~Penzar, D.~Shepelin, N.~Chekanov, O.~Kardymon, and M.~Burtsev, ``Gena-lm: a family of open-source foundational dna language models for long sequences,'' \emph{bioRxiv}, pp. 2023--06, 2023.

\bibitem{omicllm2}
Z.~Zhou, Y.~Ji, W.~Li, P.~Dutta, R.~Davuluri, and H.~Liu, ``Dnabert-2: Efficient foundation model and benchmark for multi-species genome,'' \emph{arXiv preprint arXiv:2306.15006}, 2023.

\bibitem{inferior}
Y.~Hu, T.~Li, Q.~Lu, W.~Shao, J.~He, Y.~Qiao, and P.~Luo, ``Omnimedvqa: A new large-scale comprehensive evaluation benchmark for medical lvlm,'' in \emph{Proceedings of the IEEE/CVF Conference on Computer Vision and Pattern Recognition}, 2024, pp. 22\,170--22\,183.

\bibitem{proveaigenerate}
R.~Tang, X.~Han, X.~Jiang, and X.~Hu, ``Does synthetic data generation of llms help clinical text mining? arxiv 2023,'' \emph{arXiv preprint arXiv:2303.04360}, 2023.

\bibitem{gptgenerate1}
H.~Kim, H.~Hwang, J.~Lee, S.~Park, D.~Kim, T.~Lee, C.~Yoon, J.~Sohn, D.~Choi, and J.~Kang, ``Small language models learn enhanced reasoning skills from medical textbooks,'' \emph{arXiv preprint arXiv:2404.00376}, 2024.

\bibitem{similar1}
------, ``Small language models learn enhanced reasoning skills from medical textbooks,'' \emph{arXiv preprint arXiv:2404.00376}, 2024.

\bibitem{61eye}
I.~A. Bernstein, Y.~V. Zhang, D.~Govil, I.~Majid, R.~T. Chang, Y.~Sun, A.~Shue, J.~C. Chou, E.~Schehlein, K.~L. Christopher \emph{et~al.}, ``Comparison of ophthalmologist and large language model chatbot responses to online patient eye care questions,'' \emph{JAMA network open}, vol.~6, no.~8, pp. e2\,330\,320--e2\,330\,320, 2023.

\bibitem{accudown}
J.~Liu, C.~Wang, and S.~Liu, ``Utility of chatgpt in clinical practice,'' \emph{Journal of Medical Internet Research}, vol.~25, p. e48568, 2023.

\bibitem{dictionary1}
L.~Rasmy, Y.~Xiang, Z.~Xie, C.~Tao, and D.~Zhi, ``Med-bert: pretrained contextualized embeddings on large-scale structured electronic health records for disease prediction,'' \emph{NPJ digital medicine}, vol.~4, no.~1, p.~86, 2021.

\bibitem{dictionary2}
Y.~Gu, R.~Tinn, H.~Cheng, M.~Lucas, N.~Usuyama, X.~Liu, T.~Naumann, J.~Gao, and H.~Poon, ``Domain-specific language model pretraining for biomedical natural language processing,'' \emph{ACM Transactions on Computing for Healthcare (HEALTH)}, vol.~3, no.~1, pp. 1--23, 2021.

\bibitem{raredictionary}
F.~Liu, T.~Zhu, X.~Wu, B.~Yang, C.~You, C.~Wang, L.~Lu, Z.~Liu, Y.~Zheng, X.~Sun \emph{et~al.}, ``A medical multimodal large language model for future pandemics,'' \emph{NPJ Digital Medicine}, vol.~6, no.~1, p. 226, 2023.

\bibitem{bleu}
K.~Papineni, S.~Roukos, T.~Ward, and W.-J. Zhu, ``Bleu: a method for automatic evaluation of machine translation,'' in \emph{Proceedings of the 40th annual meeting of the Association for Computational Linguistics}, 2002, pp. 311--318.

\bibitem{rouge}
C.-Y. Lin, ``Rouge: A package for automatic evaluation of summaries,'' in \emph{Text summarization branches out}, 2004, pp. 74--81.

\bibitem{cider}
R.~Vedantam, C.~Lawrence~Zitnick, and D.~Parikh, ``Cider: Consensus-based image description evaluation,'' in \emph{Proceedings of the IEEE conference on computer vision and pattern recognition}, 2015, pp. 4566--4575.

\bibitem{rougesingle}
Y.~Zhang, Y.~Pan, T.~Zhong, P.~Dong, K.~Xie, Y.~Liu, H.~Jiang, Z.~Wu, Z.~Liu, W.~Zhao \emph{et~al.}, ``Potential of multimodal large language models for data mining of medical images and free-text reports,'' \emph{Meta-Radiology}, vol.~2, no.~4, p. 100103, 2024.

\bibitem{platformrate}
M.~Moor, Q.~Huang, S.~Wu, M.~Yasunaga, Y.~Dalmia, J.~Leskovec, C.~Zakka, E.~P. Reis, and P.~Rajpurkar, ``Med-flamingo: a multimodal medical few-shot learner,'' in \emph{Machine Learning for Health (ML4H)}.\hskip 1em plus 0.5em minus 0.4em\relax PMLR, 2023, pp. 353--367.

\bibitem{artificialassess1}
P.~Chambon, C.~Bluethgen, C.~P. Langlotz, and A.~Chaudhari, ``Adapting pretrained vision-language foundational models to medical imaging domains,'' \emph{arXiv preprint arXiv:2210.04133}, 2022.

\bibitem{usml1}
H.~Nori, N.~King, S.~M. McKinney, D.~Carignan, and E.~Horvitz, ``Capabilities of gpt-4 on medical challenge problems,'' \emph{arXiv preprint arXiv:2303.13375}, 2023.

\bibitem{usml2}
Z.~Yang, L.~Li, K.~Lin, J.~Wang, C.-C. Lin, Z.~Liu, and L.~Wang, ``The dawn of lmms: Preliminary explorations with gpt-4v (ision),'' \emph{arXiv preprint arXiv:2309.17421}, vol.~9, no.~1, p.~1, 2023.

\bibitem{usml3}
T.~H. Kung, M.~Cheatham, A.~Medenilla, C.~Sillos, L.~De~Leon, C.~Elepa{\~n}o, M.~Madriaga, R.~Aggabao, G.~Diaz-Candido, J.~Maningo \emph{et~al.}, ``Performance of chatgpt on usmle: potential for ai-assisted medical education using large language models,'' \emph{PLoS digital health}, vol.~2, no.~2, p. e0000198, 2023.

\bibitem{beyond}
Y.~Nan, H.~Zhou, X.~Xing, and G.~Yang, ``Beyond the hype: A dispassionate look at vision-language models in medical scenario,'' \emph{arXiv preprint arXiv:2408.08704}, 2024.

\bibitem{hdefinition1}
H.~Alkaissi and S.~I. McFarlane, ``Artificial hallucinations in chatgpt: implications in scientific writing,'' \emph{Cureus}, vol.~15, no.~2, 2023.

\bibitem{hdefinition2}
J.~Goddard, ``Hallucinations in chatgpt: a cautionary tale for biomedical researchers,'' \emph{The American Journal of Medicine}, vol. 136, no.~11, pp. 1059--1060, 2023.

\bibitem{hdefinition3}
V.~Rawte, A.~Sheth, and A.~Das, ``A survey of hallucination in large foundation models,'' \emph{arXiv preprint arXiv:2309.05922}, 2023.

\bibitem{hdefinition4}
Z.~Ji, N.~Lee, R.~Frieske, T.~Yu, D.~Su, Y.~Xu, E.~Ishii, Y.~J. Bang, A.~Madotto, and P.~Fung, ``Survey of hallucination in natural language generation,'' \emph{ACM Computing Surveys}, vol.~55, no.~12, pp. 1--38, 2023.

\bibitem{reasoninstruct}
J.~A. Omiye, H.~Gui, S.~J. Rezaei, J.~Zou, and R.~Daneshjou, ``Large language models in medicine: the potentials and pitfalls: a narrative review,'' \emph{Annals of Internal Medicine}, vol. 177, no.~2, pp. 210--220, 2024.

\bibitem{dependent}
N.~McKenna, T.~Li, L.~Cheng, M.~J. Hosseini, M.~Johnson, and M.~Steedman, ``Sources of hallucination by large language models on inference tasks,'' \emph{arXiv preprint arXiv:2305.14552}, 2023.

\bibitem{textbooksolve}
H.~Kim, H.~Hwang, J.~Lee, S.~Park, D.~Kim, T.~Lee, C.~Yoon, J.~Sohn, D.~Choi, and J.~Kang, ``Small language models learn enhanced reasoning skills from medical textbooks,'' \emph{arXiv preprint arXiv:2404.00376}, 2024.

\bibitem{check1}
P.~Lee, S.~Bubeck, and J.~Petro, ``Benefits, limits, and risks of gpt-4 as an ai chatbot for medicine,'' \emph{New England Journal of Medicine}, vol. 388, no.~13, pp. 1233--1239, 2023.

\bibitem{correction1}
S.~Yin, C.~Fu, S.~Zhao, T.~Xu, H.~Wang, D.~Sui, Y.~Shen, K.~Li, X.~Sun, and E.~Chen, ``Woodpecker: Hallucination correction for multimodal large language models,'' \emph{arXiv preprint arXiv:2310.16045}, 2023.

\bibitem{metoo1}
Y.~Li, Y.~Du, K.~Zhou, J.~Wang, W.~X. Zhao, and J.-R. Wen, ``Evaluating object hallucination in large vision-language models,'' \emph{arXiv preprint arXiv:2305.10355}, 2023.

\bibitem{metoo2}
Y.~Zhou, C.~Cui, J.~Yoon, L.~Zhang, Z.~Deng, C.~Finn, M.~Bansal, and H.~Yao, ``Analyzing and mitigating object hallucination in large vision-language models,'' \emph{arXiv preprint arXiv:2310.00754}, 2023.

\bibitem{align1}
------, ``Analyzing and mitigating object hallucination in large vision-language models,'' \emph{arXiv preprint arXiv:2310.00754}, 2023.

\bibitem{objectmistake}
Y.~Li, Y.~Du, K.~Zhou, J.~Wang, W.~X. Zhao, and J.-R. Wen, ``Evaluating object hallucination in large vision-language models,'' \emph{arXiv preprint arXiv:2305.10355}, 2023.

\bibitem{evaluatehallucination1}
M.~Wu, J.~Ji, O.~Huang, J.~Li, Y.~Wu, X.~Sun, and R.~Ji, ``Evaluating and analyzing relationship hallucinations in large vision-language models,'' \emph{arXiv preprint arXiv:2406.16449}, 2024.

\bibitem{hallucination1}
Z.~Bai, P.~Wang, T.~Xiao, T.~He, Z.~Han, Z.~Zhang, and M.~Z. Shou, ``Hallucination of multimodal large language models: A survey,'' \emph{arXiv preprint arXiv:2404.18930}, 2024.

\bibitem{xiangsuheqformer}
H.~Xiao, F.~Zhou, X.~Liu, T.~Liu, Z.~Li, X.~Liu, and X.~Huang, ``A comprehensive survey of large language models and multimodal large language models in medicine,'' \emph{arXiv preprint arXiv:2405.08603}, 2024.

\bibitem{misobject1}
M.~Wu, X.~Zhang, X.~Sun, Y.~Zhou, C.~Chen, J.~Gu, X.~Sun, and R.~Ji, ``Difnet: Boosting visual information flow for image captioning,'' in \emph{Proceedings of the IEEE/CVF conference on computer vision and pattern recognition}, 2022, pp. 18\,020--18\,029.

\bibitem{misobject2}
Z.~Chen, Y.~Zhu, Y.~Zhan, Z.~Li, C.~Zhao, J.~Wang, and M.~Tang, ``Mitigating hallucination in visual language models with visual supervision,'' \emph{arXiv preprint arXiv:2311.16479}, 2023.

\bibitem{against1}
E.~Ferrara, ``Should chatgpt be biased? challenges and risks of bias in large language models,'' \emph{arXiv preprint arXiv:2304.03738}, 2023.

\bibitem{against2}
V.~Rawte, A.~Sheth, and A.~Das, ``A survey of hallucination in large foundation models,'' \emph{arXiv preprint arXiv:2309.05922}, 2023.

\bibitem{qassess1}
H.~Lovenia, W.~Dai, S.~Cahyawijaya, Z.~Ji, and P.~Fung, ``Negative object presence evaluation (nope) to measure object hallucination in vision-language models,'' \emph{arXiv preprint arXiv:2310.05338}, 2023.

\bibitem{qassess2}
F.~Liu, K.~Lin, L.~Li, J.~Wang, Y.~Yacoob, and L.~Wang, ``Mitigating hallucination in large multi-modal models via robust instruction tuning,'' in \emph{The Twelfth International Conference on Learning Representations}, 2023.

\bibitem{qassess3}
H.~Hu, J.~Zhang, M.~Zhao, and Z.~Sun, ``Ciem: Contrastive instruction evaluation method for better instruction tuning,'' \emph{arXiv preprint arXiv:2309.02301}, 2023.

\bibitem{whiteman}
Y.~Yang, X.~Liu, Q.~Jin, F.~Huang, and Z.~Lu, ``Unmasking and quantifying racial bias of large language models in medical report generation,'' \emph{ArXiv}, 2024.

\bibitem{blackman}
Z.~Obermeyer, B.~Powers, C.~Vogeli, and S.~Mullainathan, ``Dissecting racial bias in an algorithm used to manage the health of populations,'' \emph{Science}, vol. 366, no. 6464, pp. 447--453, 2019.

\bibitem{europegene}
G.~Sirugo, S.~M. Williams, and S.~A. Tishkoff, ``The missing diversity in human genetic studies,'' \emph{Cell}, vol. 177, no.~1, pp. 26--31, 2019.

\bibitem{lessdataresult}
L.~Seyyed-Kalantari, H.~Zhang, M.~B. McDermott, I.~Y. Chen, and M.~Ghassemi, ``Underdiagnosis bias of artificial intelligence algorithms applied to chest radiographs in under-served patient populations,'' \emph{Nature medicine}, vol.~27, no.~12, pp. 2176--2182, 2021.

\bibitem{antifactbias}
H.~Cheng, Y.~Guo, Q.~Guo, M.~Yang, T.~Gan, and L.~Nie, ``Social debiasing for fair multi-modal llms,'' \emph{arXiv preprint arXiv:2408.06569}, 2024.

\bibitem{rlhf}
D.~M. Ziegler, N.~Stiennon, J.~Wu, T.~B. Brown, A.~Radford, D.~Amodei, P.~Christiano, and G.~Irving, ``Fine-tuning language models from human preferences,'' \emph{arXiv preprint arXiv:1909.08593}, 2019.

\bibitem{empathy}
A.~Lahnala, C.~Welch, B.~Neuendorf, and L.~Flek, ``Mitigating toxic degeneration with empathetic data: Exploring the relationship between toxicity and empathy,'' \emph{arXiv preprint arXiv:2205.07233}, 2022.

\bibitem{luobasedata}
Y.~Luo, M.~Shi, M.~O. Khan, M.~M. Afzal, H.~Huang, S.~Yuan, Y.~Tian, L.~Song, A.~Kouhana, T.~Elze \emph{et~al.}, ``Fairclip: Harnessing fairness in vision-language learning,'' in \emph{Proceedings of the IEEE/CVF Conference on Computer Vision and Pattern Recognition}, 2024, pp. 12\,289--12\,301.

\bibitem{dynamic}
S.~Amba~Hombaiah, T.~Chen, M.~Zhang, M.~Bendersky, and M.~Najork, ``Dynamic language models for continuously evolving content,'' in \emph{Proceedings of the 27th ACM SIGKDD Conference on Knowledge Discovery \& Data Mining}, 2021, pp. 2514--2524.

\bibitem{keepft}
T.~Wu, L.~Luo, Y.-F. Li, S.~Pan, T.-T. Vu, and G.~Haffari, ``Continual learning for large language models: A survey,'' \emph{arXiv preprint arXiv:2402.01364}, 2024.

\bibitem{reviewdisaster}
H.~Yi, Z.~Qin, Q.~Lao, W.~Xu, Z.~Jiang, D.~Wang, S.~Zhang, and K.~Li, ``Towards general purpose medical ai: Continual learning medical foundation model,'' \emph{arXiv preprint arXiv:2303.06580}, 2023.

\bibitem{segmentreview}
Y.~Zhai, S.~Tong, X.~Li, M.~Cai, Q.~Qu, Y.~J. Lee, and Y.~Ma, ``Investigating the catastrophic forgetting in multimodal large language models,'' \emph{arXiv preprint arXiv:2309.10313}, 2023.

\bibitem{minimize1}
Z.~Yuan, Z.~Li, W.~Huang, Y.~Ye, and L.~Sun, ``Tinygpt-v: Efficient multimodal large language model via small backbones,'' \emph{arXiv preprint arXiv:2312.16862}, 2023.

\bibitem{minimize2}
H.~Wei, L.~Kong, J.~Chen, L.~Zhao, Z.~Ge, E.~Yu, J.~Sun, C.~Han, and X.~Zhang, ``Small language model meets with reinforced vision vocabulary,'' \emph{arXiv preprint arXiv:2401.12503}, 2024.

\bibitem{mobilevlm}
X.~Chu, L.~Qiao, X.~Lin, S.~Xu, Y.~Yang, Y.~Hu, F.~Wei, X.~Zhang, B.~Zhang, X.~Wei \emph{et~al.}, ``Mobilevlm: A fast, strong and open vision language assistant for mobile devices,'' \emph{arXiv preprint arXiv:2312.16886}, 2023.

\bibitem{xmodelllava}
W.~Xu, Y.~Liu, L.~He, X.~Huang, and L.~Jiang, ``Xmodel-vlm: A simple baseline for multimodal vision language model,'' \emph{arXiv preprint arXiv:2405.09215}, 2024.

\bibitem{lora}
E.~J. Hu, Y.~Shen, P.~Wallis, Z.~Allen-Zhu, Y.~Li, S.~Wang, L.~Wang, and W.~Chen, ``Lora: Low-rank adaptation of large language models,'' \emph{arXiv preprint arXiv:2106.09685}, 2021.

\bibitem{qlora}
T.~Dettmers, A.~Pagnoni, A.~Holtzman, and L.~Zettlemoyer, ``Qlora: Efficient finetuning of quantized llms,'' in \emph{NeurIPS}, 2024.

\bibitem{sixformat}
S.~Jiang, T.~Zheng, Y.~Zhang, Y.~Jin, L.~Yuan, and Z.~Liu, ``Med-moe: Mixture of domain-specific experts for lightweight medical vision-language models,'' in \emph{Findings of the Association for Computational Linguistics: EMNLP 2024}, 2024, pp. 3843--3860.

\bibitem{legislation1}
L.~B. Harman, C.~A. Flite, and K.~Bond, ``Electronic health records: privacy, confidentiality, and security,'' \emph{AMA journal of ethics}, vol.~14, no.~9, pp. 712--719, 2012.

\bibitem{legislation2}
E.~W. Clayton, P.~J. Emb{\'\i}, and B.~A. Malin, ``Dobbs and the future of health data privacy for patients and healthcare organizations,'' \emph{Journal of the American Medical Informatics Association}, vol.~30, no.~1, pp. 155--160, 2023.

\bibitem{legislation3}
L.~Harman, C.~Flite, and K.~Bond, ``State of the art and science. electronic health records: Privacy, confidentiality, and security. am med assoc j ethics. 2012; 14 (9): 712--9.''

\bibitem{attackexpose1}
N.~Carlini, F.~Tramer, E.~Wallace, M.~Jagielski, A.~Herbert-Voss, K.~Lee, A.~Roberts, T.~Brown, D.~Song, U.~Erlingsson \emph{et~al.}, ``Extracting training data from large language models,'' in \emph{30th USENIX Security Symposium (USENIX Security 21)}, 2021, pp. 2633--2650.

\bibitem{identify1}
Y.~Erlich, T.~Shor, I.~Pe’er, and S.~Carmi, ``Identity inference of genomic data using long-range familial searches,'' \emph{Science}, vol. 362, no. 6415, pp. 690--694, 2018.

\bibitem{inferidentity2}
M.~Gymrek, A.~L. McGuire, D.~Golan, E.~Halperin, and Y.~Erlich, ``Identifying personal genomes by surname inference,'' \emph{Science}, vol. 339, no. 6117, pp. 321--324, 2013.

\bibitem{extract1}
N.~Carlini, F.~Tramer, E.~Wallace, M.~Jagielski, A.~Herbert-Voss, K.~Lee, A.~Roberts, T.~Brown, D.~Song, U.~Erlingsson \emph{et~al.}, ``Extracting training data from large language models,'' in \emph{30th USENIX Security Symposium (USENIX Security 21)}, 2021, pp. 2633--2650.

\bibitem{netflixidentify}
A.~Narayanan and V.~Shmatikov, ``Robust de-anonymization of large sparse datasets,'' in \emph{2008 IEEE Symposium on Security and Privacy (sp 2008)}.\hskip 1em plus 0.5em minus 0.4em\relax IEEE, 2008, pp. 111--125.

\bibitem{monotonous2}
J.~M. Liu, D.~Li, H.~Cao, T.~Ren, Z.~Liao, and J.~Wu, ``Chatcounselor: A large language models for mental health support,'' \emph{arXiv preprint arXiv:2309.15461}, 2023.

\bibitem{delete2}
M.~R. Kosorok and E.~B. Laber, ``Annual review of statistics and its application,'' \emph{Precis Med}, vol.~6, pp. 263--286, 2019.

\bibitem{chafen1}
B.~Mudassar, S.~Tahir, F.~Khan, S.~A. Shah, S.~I. Shah, and Q.~H. Abbasi, ``Privacy-preserving data analytics in internet of medical things,'' \emph{Future Internet}, vol.~16, no.~11, p. 407, 2024.

\bibitem{chafen2}
Y.~Sei, A.~Ohsuga, J.~A. Onesimu, and A.~L. Imoize, ``Local differential privacy for artificial intelligence of medical things,'' in \emph{Handbook of Security and Privacy of AI-Enabled Healthcare Systems and Internet of Medical Things}.\hskip 1em plus 0.5em minus 0.4em\relax CRC Press, 2024, pp. 241--270.

\bibitem{chafen3}
Z.~Guan, Z.~Lv, X.~Du, L.~Wu, and M.~Guizani, ``Achieving data utility-privacy tradeoff in internet of medical things: A machine learning approach,'' \emph{Future Generation Computer Systems}, vol.~98, pp. 60--68, 2019.

\bibitem{algorism2}
R.~Wang, J.~Lai, Z.~Zhang, X.~Li, P.~Vijayakumar, and M.~Karuppiah, ``Privacy-preserving federated learning for internet of medical things under edge computing,'' \emph{IEEE journal of biomedical and health informatics}, vol.~27, no.~2, pp. 854--865, 2022.

\bibitem{lianbang}
G.~A. Kaissis, M.~R. Makowski, D.~R{\"u}ckert, and R.~F. Braren, ``Secure, privacy-preserving and federated machine learning in medical imaging,'' \emph{Nature Machine Intelligence}, vol.~2, no.~6, pp. 305--311, 2020.

\bibitem{samecode}
A.~Wood, K.~Najarian, and D.~Kahrobaei, ``Homomorphic encryption for machine learning in medicine and bioinformatics,'' \emph{ACM Computing Surveys (CSUR)}, vol.~53, no.~4, pp. 1--35, 2020.

\bibitem{bigmistake1}
C.~Zhou, Q.~Li, C.~Li, J.~Yu, Y.~Liu, G.~Wang, K.~Zhang, C.~Ji, Q.~Yan, L.~He \emph{et~al.}, ``A comprehensive survey on pretrained foundation models: A history from bert to chatgpt,'' \emph{International Journal of Machine Learning and Cybernetics}, pp. 1--65, 2024.

\bibitem{downt1}
W.-L. Chiang, Z.~Li, Z.~Lin, Y.~Sheng, Z.~Wu, H.~Zhang, L.~Zheng, S.~Zhuang, Y.~Zhuang, J.~E. Gonzalez \emph{et~al.}, ``Vicuna: An open-source chatbot impressing gpt-4 with 90\%* chatgpt quality,'' \emph{See https://vicuna. lmsys. org (accessed 14 April 2023)}, vol.~2, no.~3, p.~6, 2023.

\bibitem{downt2}
X.~Geng, A.~Gudibande, H.~Liu, E.~Wallace, P.~Abbeel, S.~Levine, and D.~Song, ``Koala: A dialogue model for academic research,'' \emph{Blog post, April}, vol.~1, p.~6, 2023.

\bibitem{downt3}
K.~Singhal, S.~Azizi, T.~Tu, S.~S. Mahdavi, J.~Wei, H.~W. Chung, N.~Scales, A.~Tanwani, H.~Cole-Lewis, S.~Pfohl \emph{et~al.}, ``Large language models encode clinical knowledge,'' \emph{Nature}, vol. 620, no. 7972, pp. 172--180, 2023.

\bibitem{mfttasks1}
R.~K. Mahabadi, S.~Ruder, M.~Dehghani, and J.~Henderson, ``Parameter-efficient multi-task fine-tuning for transformers via shared hypernetworks,'' \emph{arXiv preprint arXiv:2106.04489}, 2021.

\bibitem{mfttasks2}
X.~Liu, P.~He, W.~Chen, and J.~Gao, ``Multi-task deep neural networks for natural language understanding,'' \emph{arXiv preprint arXiv:1901.11504}, 2019.

\bibitem{msearch1}
D.~Duong and B.~D. Solomon, ``Analysis of large-language model versus human performance for genetics questions,'' \emph{European Journal of Human Genetics}, vol.~32, no.~4, pp. 466--468, 2024.

\bibitem{msearch2}
Y.~Dubois, C.~X. Li, R.~Taori, T.~Zhang, I.~Gulrajani, J.~Ba, C.~Guestrin, P.~S. Liang, and T.~B. Hashimoto, ``Alpacafarm: A simulation framework for methods that learn from human feedback,'' \emph{NeurIPS}, 2024.

\bibitem{massist1}
Z.~Wang, R.~Li, B.~Dong, J.~Wang, X.~Li, N.~Liu, C.~Mao, W.~Zhang, L.~Dong, J.~Gao \emph{et~al.}, ``Can llms like gpt-4 outperform traditional ai tools in dementia diagnosis? maybe, but not today,'' \emph{arXiv preprint arXiv:2306.01499}, 2023.

\bibitem{massist2}
A.~Gilson, C.~W. Safranek, T.~Huang, V.~Socrates, L.~Chi, R.~A. Taylor, D.~Chartash \emph{et~al.}, ``How does chatgpt perform on the united states medical licensing examination (usmle)? the implications of large language models for medical education and knowledge assessment,'' \emph{JMIR medical education}, vol.~9, no.~1, p. e45312, 2023.

\bibitem{mexam1}
T.~H. Kung, M.~Cheatham, A.~Medenilla, C.~Sillos, L.~De~Leon, C.~Elepa{\~n}o, M.~Madriaga, R.~Aggabao, G.~Diaz-Candido, J.~Maningo \emph{et~al.}, ``Performance of chatgpt on usmle: potential for ai-assisted medical education using large language models,'' \emph{PLoS digital health}, vol.~2, no.~2, p. e0000198, 2023.

\bibitem{visualbert}
L.~H. Li, M.~Yatskar, D.~Yin, C.-J. Hsieh, and K.-W. Chang, ``Visualbert: A simple and performant baseline for vision and language,'' \emph{arXiv preprint arXiv:1908.03557}, 2019.

\bibitem{flava}
A.~Singh, R.~Hu, V.~Goswami, G.~Couairon, W.~Galuba, M.~Rohrbach, and D.~Kiela, ``Flava: A foundational language and vision alignment model,'' in \emph{Proceedings of the IEEE/CVF Conference on Computer Vision and Pattern Recognition}, 2022, pp. 15\,638--15\,650.

\bibitem{vilbert}
J.~Lu, D.~Batra, D.~Parikh, and S.~Lee, ``Vilbert: Pretraining task-agnostic visiolinguistic representations for vision-and-language tasks,'' in \emph{NeurIPS}, 2019.

\bibitem{beit}
H.~Bao, L.~Dong, S.~Piao, and F.~Wei, ``Beit: Bert pre-training of image transformers,'' \emph{arXiv preprint arXiv:2106.08254}, 2021.

\bibitem{autopencoders}
K.~He, X.~Chen, S.~Xie, Y.~Li, P.~Doll{\'a}r, and R.~Girshick, ``Masked autoencoders are scalable vision learners,'' in \emph{Proceedings of the IEEE/CVF conference on computer vision and pattern recognition}, 2022, pp. 16\,000--16\,009.

\bibitem{masked}
G.~Kwon, Z.~Cai, A.~Ravichandran, E.~Bas, R.~Bhotika, and S.~Soatto, ``Masked vision and language modeling for multi-modal representation learning,'' \emph{arXiv preprint arXiv:2208.02131}, 2022.

\bibitem{t5}
C.~Raffel, N.~Shazeer, A.~Roberts, K.~Lee, S.~Narang, M.~Matena, Y.~Zhou, W.~Li, and P.~J. Liu, ``Exploring the limits of transfer learning with a unified text-to-text transformer,'' \emph{Journal of machine learning research}, vol.~21, no. 140, pp. 1--67, 2020.

\bibitem{vit}
A.~Dosovitskiy, ``An image is worth 16x16 words: Transformers for image recognition at scale,'' \emph{arXiv preprint arXiv:2010.11929}, 2020.

\bibitem{mae}
K.~He, X.~Chen, S.~Xie, Y.~Li, P.~Doll{\'a}r, and R.~Girshick, ``Masked autoencoders are scalable vision learners,'' in \emph{Proceedings of the IEEE/CVF conference on computer vision and pattern recognition}, 2022, pp. 16\,000--16\,009.

\bibitem{swin}
Z.~Liu, Y.~Lin, Y.~Cao, H.~Hu, Y.~Wei, Z.~Zhang, S.~Lin, and B.~Guo, ``Swin transformer: Hierarchical vision transformer using shifted windows,'' in \emph{Proceedings of the IEEE/CVF international conference on computer vision}, 2021, pp. 10\,012--10\,022.

\bibitem{bubo}
Y.~Zhao, Z.~Lin, D.~Zhou, Z.~Huang, J.~Feng, and B.~Kang, ``Bubogpt: Enabling visual grounding in multi-modal llms,'' \emph{arXiv preprint arXiv:2307.08581}, 2023.

\bibitem{sam}
A.~Kirillov, E.~Mintun, N.~Ravi, H.~Mao, C.~Rolland, L.~Gustafson, T.~Xiao, S.~Whitehead, A.~C. Berg, W.-Y. Lo \emph{et~al.}, ``Segment anything,'' in \emph{Proceedings of the IEEE/CVF International Conference on Computer Vision}, 2023, pp. 4015--4026.

\bibitem{videoec}
S.~Yu, J.~Cho, P.~Yadav, and M.~Bansal, ``Self-chained image-language model for video localization and question answering,'' \emph{NeurIPS}, 2024.

\bibitem{cformer}
F.~Chen, M.~Han, H.~Zhao, Q.~Zhang, J.~Shi, S.~Xu, and B.~Xu, ``X-llm: Bootstrapping advanced large language models by treating multi-modalities as foreign languages,'' \emph{arXiv preprint arXiv:2305.04160}, 2023.

\bibitem{whisper}
A.~Radford, J.~W. Kim, T.~Xu, G.~Brockman, C.~McLeavey, and I.~Sutskever, ``Robust speech recognition via large-scale weak supervision,'' in \emph{International conference on machine learning}.\hskip 1em plus 0.5em minus 0.4em\relax PMLR, 2023, pp. 28\,492--28\,518.

\bibitem{hubert}
W.-N. Hsu, B.~Bolte, Y.-H.~H. Tsai, K.~Lakhotia, R.~Salakhutdinov, and A.~Mohamed, ``Hubert: Self-supervised speech representation learning by masked prediction of hidden units,'' \emph{IEEE/ACM transactions on audio, speech, and language processing}, vol.~29, pp. 3451--3460, 2021.

\bibitem{llava}
H.~Liu, C.~Li, Q.~Wu, and Y.~J. Lee, ``Visual instruction tuning,'' in \emph{NeurIPS}, 2024.

\bibitem{mlp}
D.~E. Rumelhart, G.~E. Hinton, and R.~J. Williams, ``Learning representations by back-propagating errors,'' \emph{nature}, vol. 323, no. 6088, pp. 533--536, 1986.

\bibitem{qllama}
Z.~Chen, J.~Wu, W.~Wang, W.~Su, G.~Chen, S.~Xing, M.~Zhong, Q.~Zhang, X.~Zhu, L.~Lu \emph{et~al.}, ``Internvl: Scaling up vision foundation models and aligning for generic visual-linguistic tasks,'' in \emph{Proceedings of the IEEE/CVF Conference on Computer Vision and Pattern Recognition}, 2024, pp. 24\,185--24\,198.

\bibitem{chexpert}
J.~Irvin, P.~Rajpurkar, M.~Ko, Y.~Yu, S.~Ciurea-Ilcus, C.~Chute, H.~Marklund, B.~Haghgoo, R.~Ball, K.~Shpanskaya \emph{et~al.}, ``Chexpert: A large chest radiograph dataset with uncertainty labels and expert comparison,'' in \emph{Proceedings of the AAAI conference on artificial intelligence}, vol.~33, no.~01, 2019, pp. 590--597.

\bibitem{endovis}
{German Cancer Research Center}, ``{ENDOVIS Datasets and Publications},'' \url{https://opencas.dkfz.de/endovis/datasetspublications/}, 2024, accessed: 2024-12-18.

\bibitem{vqarad}
J.~J. Lau, S.~Gayen, A.~Ben~Abacha, and D.~Demner-Fushman, ``A dataset of clinically generated visual questions and answers about radiology images,'' \emph{Scientific data}, vol.~5, no.~1, pp. 1--10, 2018.

\bibitem{mimiccxrjpg}
A.~E. Johnson, T.~J. Pollard, N.~R. Greenbaum, M.~P. Lungren, C.-y. Deng, Y.~Peng, Z.~Lu, R.~G. Mark, S.~J. Berkowitz, and S.~Horng, ``Mimic-cxr-jpg, a large publicly available database of labeled chest radiographs,'' \emph{arXiv preprint arXiv:1901.07042}, 2019.

\bibitem{swissprotclap}
S.~Liu, Y.~Li, Z.~Li, A.~Gitter, Y.~Zhu, J.~Lu, Z.~Xu, W.~Nie, A.~Ramanathan, C.~Xiao \emph{et~al.}, ``A text-guided protein design framework,'' \emph{arXiv preprint arXiv:2302.04611}, 2023.

\bibitem{medicat}
S.~Subramanian, L.~L. Wang, S.~Mehta, B.~Bogin, M.~van Zuylen, S.~Parasa, S.~Singh, M.~Gardner, and H.~Hajishirzi, ``Medicat: A dataset of medical images, captions, and textual references,'' \emph{arXiv preprint arXiv:2010.06000}, 2020.

\bibitem{pathvqa}
X.~He, Y.~Zhang, L.~Mou, E.~Xing, and P.~Xie, ``Pathvqa: 30000+ questions for medical visual question answering,'' \emph{arXiv preprint arXiv:2003.10286}, 2020.

\bibitem{satds}
Z.~Zhao, Y.~Zhang, C.~Wu, X.~Zhang, Y.~Zhang, Y.~Wang, and W.~Xie, ``One model to rule them all: Towards universal segmentation for medical images with text prompts,'' \emph{arXiv preprint arXiv:2312.17183}, 2023.

\bibitem{medtrinity}
Y.~Xie, C.~Zhou, L.~Gao, J.~Wu, X.~Li, H.-Y. Zhou, S.~Liu, L.~Xing, J.~Zou, C.~Xie \emph{et~al.}, ``Medtrinity-25m: A large-scale multimodal dataset with multigranular annotations for medicine,'' \emph{arXiv preprint arXiv:2408.02900}, 2024.

\bibitem{chexinstruct}
Z.~Chen, M.~Varma, J.-B. Delbrouck, M.~Paschali, L.~Blankemeier, D.~Van~Veen, J.~M.~J. Valanarasu, A.~Youssef, J.~P. Cohen, E.~P. Reis \emph{et~al.}, ``Chexagent: Towards a foundation model for chest x-ray interpretation,'' \emph{arXiv preprint arXiv:2401.12208}, 2024.

\bibitem{slake}
B.~Liu, L.-M. Zhan, L.~Xu, L.~Ma, Y.~Yang, and X.-M. Wu, ``Slake: A semantically-labeled knowledge-enhanced dataset for medical visual question answering,'' in \emph{ISBI}, 2021.

\bibitem{mimicnle}
M.~Kayser, C.~Emde, O.-M. Camburu, G.~Parsons, B.~Papiez, and T.~Lukasiewicz, ``Explaining chest x-ray pathologies in natural language,'' in \emph{MICCAI}, 2022.

\bibitem{chimed}
Q.~Ye, J.~Liu, D.~Chong, P.~Zhou, Y.~Hua, F.~Liu, M.~Cao, Z.~Wang, X.~Cheng, Z.~Lei \emph{et~al.}, ``Qilin-med: Multi-stage knowledge injection advanced medical large language model,'' \emph{arXiv preprint arXiv:2310.09089}, 2023.

\bibitem{cxrpro}
V.~Ramesh, N.~Chi, and P.~Rajpurkar, ``Cxr-pro: Mimic-cxr with prior references omitted,'' \emph{arXiv preprint arXiv:2311.16863}, 2023.

\bibitem{pubmedvisionalignment}
J.~Chen, C.~Gui, R.~Ouyang, A.~Gao, S.~Chen, G.~Chen, X.~Wang, Z.~Cai, K.~Ji, X.~Wan \emph{et~al.}, ``Towards injecting medical visual knowledge into multimodal llms at scale,'' in \emph{EMNLP}, 2024.

\bibitem{medmd}
C.~Wu, X.~Zhang, Y.~Zhang, Y.~Wang, and W.~Xie, ``Towards generalist foundation model for radiology,'' \emph{arXiv preprint arXiv:2308.02463}, 2023.

\bibitem{pathcap}
Y.~Sun, C.~Zhu, S.~Zheng, K.~Zhang, Z.~Shui, X.~Yu, Y.~Zhao, H.~Li, Y.~Zhang, R.~Zhao \emph{et~al.}, ``Pathasst: Redefining pathology through generative foundation ai assistant for pathology,'' \emph{arXiv preprint arXiv:2305.15072}, vol.~2, 2023.

\bibitem{quilt1m}
W.~Ikezogwo, S.~Seyfioglu, F.~Ghezloo, D.~Geva, F.~Sheikh~Mohammed, P.~K. Anand, R.~Krishna, and L.~Shapiro, ``Quilt-1m: One million image-text pairs for histopathology,'' in \emph{NeurIPS}, 2024.

\bibitem{openpath}
Z.~Huang, F.~Bianchi, M.~Yuksekgonul, T.~J. Montine, and J.~Zou, ``A visual--language foundation model for pathology image analysis using medical twitter,'' \emph{Nature medicine}, vol.~29, no.~9, pp. 2307--2316, 2023.

\bibitem{pmc15m}
S.~Zhang, Y.~Xu, N.~Usuyama, J.~Bagga, R.~Tinn, S.~Preston, R.~Rao, M.~Wei, N.~Valluri, C.~Wong \emph{et~al.}, ``Large-scale domain-specific pretraining for biomedical vision-language processing,'' \emph{arXiv preprint arXiv:2303.00915}, vol.~2, no.~3, p.~6, 2023.

\bibitem{mscxr}
S.~Bannur, S.~Hyland, Q.~Liu, F.~P{\'e}rez-Garc{\'\i}a, M.~Ilse, D.~C. de~Castro, B.~Boecking, H.~Sharma, K.~Bouzid, A.~Schwaighofer \emph{et~al.}, ``Ms-cxr-t: Learning to exploit temporal structure for biomedical vision-language processing,'' 2023.

\bibitem{iu}
D.~Demner-Fushman \emph{et~al.}, ``Preparing a collection of radiology examinations for distribution and retrieval,'' \emph{Journal of the American Medical Informatics Association}, vol.~23, no.~2, pp. 304--310, 2015.

\bibitem{roco}
O.~Pelka, S.~Koitka, J.~R{\"u}ckert, F.~Nensa, and C.~M. Friedrich, ``Radiology objects in context (roco): a multimodal image dataset,'' in \emph{MICCAI}, 2018.

\bibitem{llavamedalignment}
C.~Li, C.~Wong, S.~Zhang, N.~Usuyama, H.~Liu, J.~Yang, T.~Naumann, H.~Poon, and J.~Gao, ``Llava-med: Training a large language-and-vision assistant for biomedicine in one day,'' in \emph{NeurIPS}, 2024.

\bibitem{chimedvl}
J.~Liu, Z.~Wang, Q.~Ye, D.~Chong, P.~Zhou, and Y.~Hua, ``Qilin-med-vl: Towards chinese large vision-language model for general healthcare,'' \emph{arXiv preprint arXiv:2310.17956}, 2023.

\bibitem{apollo}
X.~Wang, N.~Chen, J.~Chen, Y.~Hu, Y.~Wang, X.~Wu, A.~Gao, X.~Wan, H.~Li, and B.~Wang, ``Apollo: Lightweight multilingual medical llms towards democratizing medical ai to 6b people,'' \emph{arXiv preprint arXiv:2403.03640}, 2024.

\bibitem{m3ddata}
F.~Bai, Y.~Du, T.~Huang, M.~Q.-H. Meng, and B.~Zhao, ``M3d: Advancing 3d medical image analysis with multi-modal large language models,'' \emph{arXiv preprint arXiv:2404.00578}, 2024.

\bibitem{mimiccxr}
A.~E. Johnson, T.~J. Pollard, S.~J. Berkowitz, N.~R. Greenbaum, M.~P. Lungren, C.-y. Deng, R.~G. Mark, and S.~Horng, ``Mimic-cxr, a de-identified publicly available database of chest radiographs with free-text reports,'' \emph{Scientific data}, vol.~6, no.~1, p. 317, 2019.

\bibitem{nci_tcga}
{National Cancer Institute}, ``{The Cancer Genome Atlas},'' \url{https://www.cancer.gov/ccg/research/genome-sequencing/tcga}, 2024, accessed: 2024-12-18.

\bibitem{medicalmeadow}
T.~Han, L.~C. Adams, J.-M. Papaioannou, P.~Grundmann, T.~Oberhauser, A.~L{\"o}ser, D.~Truhn, and K.~K. Bressem, ``Medalpaca--an open-source collection of medical conversational ai models and training data,'' \emph{arXiv preprint arXiv:2304.08247}, 2023.

\bibitem{cprd}
E.~Herrett, A.~M. Gallagher, K.~Bhaskaran, H.~Forbes, R.~Mathur, T.~Van~Staa, and L.~Smeeth, ``Data resource profile: clinical practice research datalink (cprd),'' \emph{International journal of epidemiology}, vol.~44, no.~3, pp. 827--836, 2015.

\bibitem{mimicthree}
A.~E. Johnson, T.~J. Pollard, L.~Shen, L.-w.~H. Lehman, M.~Feng, M.~Ghassemi, B.~Moody, P.~Szolovits, L.~Anthony~Celi, and R.~G. Mark, ``Mimic-iii, a freely accessible critical care database,'' \emph{Scientific data}, vol.~3, no.~1, pp. 1--9, 2016.

\bibitem{cmedqa2}
S.~Zhang, X.~Zhang, H.~Wang, L.~Guo, and S.~Liu, ``Multi-scale attentive interaction networks for chinese medical question answer selection,'' \emph{IEEE Access}, vol.~6, pp. 74\,061--74\,071, 2018.

\bibitem{pubmedcausal}
B.~Yu, Y.~Li, and J.~Wang, ``Detecting causal language use in science findings,'' in \emph{EMNLP}, 2019.

\bibitem{meqsum}
A.~B. Abacha and D.~Demner-Fushman, ``On the summarization of consumer health questions,'' in \emph{Proceedings of the 57th Annual Meeting of the Association for Computational Linguistics}, 2019, pp. 2228--2234.

\bibitem{medmentions}
S.~Mohan and D.~Li, ``Medmentions: A large biomedical corpus annotated with umls concepts,'' \emph{arXiv preprint arXiv:1902.09476}, 2019.

\bibitem{medquad}
A.~Ben~Abacha and D.~Demner-Fushman, ``A question-entailment approach to question answering,'' \emph{BMC bioinformatics}, vol.~20, pp. 1--23, 2019.

\bibitem{cmekg}
O.~Byambasuren, Y.~Yang, Z.~Sui, D.~Dai, B.~Chang, S.~Li, and H.~Zan, ``Preliminary study on the construction of chinese medical knowledge graph,'' \emph{Journal of Chinese Information Processing}, vol.~33, no.~10, pp. 1--9, 2019.

\bibitem{webmedqa}
J.~He, M.~Fu, and M.~Tu, ``Applying deep matching networks to chinese medical question answering: a study and a dataset,'' \emph{BMC medical informatics and decision making}, vol.~19, pp. 91--100, 2019.

\bibitem{pubmedqa}
Q.~Jin, B.~Dhingra, Z.~Liu, W.~W. Cohen, and X.~Lu, ``Pubmedqa: A dataset for biomedical research question answering,'' \emph{arXiv preprint arXiv:1909.06146}, 2019.

\bibitem{mediqa}
M.~Savery, A.~B. Abacha, S.~Gayen, and D.~Demner-Fushman, ``Question-driven summarization of answers to consumer health questions,'' \emph{Scientific Data}, vol.~7, no.~1, p. 322, 2020.

\bibitem{thepile}
L.~Gao, S.~Biderman, S.~Black, L.~Golding, T.~Hoppe, C.~Foster, J.~Phang, H.~He, A.~Thite, N.~Nabeshima \emph{et~al.}, ``The pile: An 800gb dataset of diverse text for language modeling,'' \emph{arXiv preprint arXiv:2101.00027}, 2020.

\bibitem{cometa}
M.~Basaldella, F.~Liu, E.~Shareghi, and N.~Collier, ``Cometa: A corpus for medical entity linking in the social media,'' \emph{arXiv preprint arXiv:2010.03295}, 2020.

\bibitem{cord19}
L.~L. Wang, K.~Lo, Y.~Chandrasekhar, R.~Reas, J.~Yang, D.~Burdick, D.~Eide, K.~Funk, Y.~Katsis, R.~Kinney \emph{et~al.}, ``Cord-19: The covid-19 open research dataset,'' \emph{ArXiv}, 2020.

\bibitem{mimic4}
A.~Johnson, L.~Bulgarelli, T.~Pollard, S.~Horng, L.~Celi, and R.~Mark, ``Mimic-iv. physionet,'' 2021.

\bibitem{rct}
B.~C. Wallace, S.~Saha, F.~Soboczenski, and I.~J. Marshall, ``Generating (factual?) narrative summaries of rcts: Experiments with neural multi-document summarization,'' \emph{AMIA Summits on Translational Science Proceedings}, vol. 2021, p. 605, 2021.

\bibitem{ms2}
J.~DeYoung, I.~Beltagy, M.~van Zuylen, B.~Kuehl, and L.~L. Wang, ``Ms2: Multi-document summarization of medical studies,'' \emph{arXiv preprint arXiv:2104.06486}, 2021.

\bibitem{cdsr}
Y.~Guo, W.~Qiu, Y.~Wang, and T.~Cohen, ``Automated lay language summarization of biomedical scientific reviews,'' in \emph{Proceedings of the AAAI Conference on Artificial Intelligence}, vol.~35, no.~1, 2021, pp. 160--168.

\bibitem{sumpudmed}
V.~Gupta, P.~Bharti, P.~Nokhiz, and H.~Karnick, ``Sumpubmed: Summarization dataset of pubmed scientific articles,'' in \emph{Proceedings of the 59th Annual Meeting of the Association for Computational Linguistics and the 11th International Joint Conference on Natural Language Processing: Student Research Workshop}, 2021.

\bibitem{medqausmile}
D.~Jin, E.~Pan, N.~Oufattole, W.-H. Weng, H.~Fang, and P.~Szolovits, ``What disease does this patient have? a large-scale open domain question answering dataset from medical exams,'' \emph{Applied Sciences}, vol.~11, no.~14, p. 6421, 2021.

\bibitem{s2orc}
J.~Bishop, Q.~Xie, and S.~Ananiadou, ``Gencomparesum: a hybrid unsupervised summarization method using salience,'' in \emph{Proceedings of the 21st workshop on biomedical language processing}, 2022.

\bibitem{chqsumm}
S.~Yadav, D.~Gupta, and D.~Demner-Fushman, ``Chq-summ: A dataset for consumer healthcare question summarization,'' \emph{arXiv preprint arXiv:2206.06581}, 2022.

\bibitem{medmcqa}
A.~Pal, L.~K. Umapathi, and M.~Sankarasubbu, ``Medmcqa: A large-scale multi-subject multi-choice dataset for medical domain question answering,'' in \emph{Conference on health, inference, and learning}.\hskip 1em plus 0.5em minus 0.4em\relax PMLR, 2022, pp. 248--260.

\bibitem{psych8k}
J.~M. Liu, D.~Li, H.~Cao, T.~Ren, Z.~Liao, and J.~Wu, ``Chatcounselor: A large language models for mental health support,'' \emph{arXiv preprint arXiv:2309.15461}, 2023.

\bibitem{huatuo26m}
J.~Li, X.~Wang, X.~Wu, Z.~Zhang, X.~Xu, J.~Fu, P.~Tiwari, X.~Wan, and B.~Wang, ``Huatuo-26m, a large-scale chinese medical qa dataset,'' \emph{arXiv preprint arXiv:2305.01526}, 2023.

\bibitem{medinstruct52k}
X.~Zhang, C.~Tian, X.~Yang, L.~Chen, Z.~Li, and L.~R. Petzold, ``Alpacare: Instruction-tuned large language models for medical application,'' \emph{arXiv preprint arXiv:2310.14558}, 2023.

\bibitem{gapreplay}
Z.~Chen, A.~H. Cano, A.~Romanou, A.~Bonnet, K.~Matoba, F.~Salvi, M.~Pagliardini, S.~Fan, A.~K{\"o}pf, A.~Mohtashami \emph{et~al.}, ``Meditron-70b: Scaling medical pretraining for large language models,'' \emph{arXiv preprint arXiv:2311.16079}, 2023.

\bibitem{alpaca}
R.~Taori, I.~Gulrajani, T.~Zhang, Y.~Dubois, X.~Li, C.~Guestrin, P.~Liang, and T.~B. Hashimoto, ``Stanford alpaca: An instruction-following llama model,'' 2023.

\bibitem{cmtmedqa}
S.~Yang, H.~Zhao, S.~Zhu, G.~Zhou, H.~Xu, Y.~Jia, and H.~Zan, ``Zhongjing: Enhancing the chinese medical capabilities of large language model through expert feedback and real-world multi-turn dialogue,'' in \emph{Proceedings of the AAAI Conference on Artificial Intelligence}, vol.~38, no.~17, 2024, pp. 19\,368--19\,376.

\bibitem{tcmcorpus1b}
G.~Yang, X.~Liu, J.~Shi, Z.~Wang, and G.~Wang, ``Tcm-gpt: Efficient pre-training of large language models for domain adaptation in traditional chinese medicine,'' \emph{Computer Methods and Programs in Biomedicine Update}, p. 100158, 2024.

\bibitem{pubmed2024}
{PubMed}, ``Pubmed database,'' \url{https://pubmed.ncbi.nlm.nih.gov/}, 2024, accessed: 2024-12-18.

\bibitem{pmc}
\BIBentryALTinterwordspacing
{PubMed Central}, ``{PubMed Central},'' 2024, accessed: 2024-12-18. [Online]. Available: \url{https://pmc.ncbi.nlm.nih.gov/}
\BIBentrySTDinterwordspacing

\bibitem{huatuollm}
H.~Wang, C.~Liu, N.~Xi, Z.~Qiang, S.~Zhao, B.~Qin, and T.~Liu, ``Huatuo: Tuning llama model with chinese medical knowledge,'' \emph{arXiv preprint arXiv:2304.06975}, 2023.

\bibitem{medalpaca}
T.~Han, L.~C. Adams, J.-M. Papaioannou, P.~Grundmann, T.~Oberhauser, A.~L{\"o}ser, D.~Truhn, and K.~K. Bressem, ``Medalpaca--an open-source collection of medical conversational ai models and training data,'' \emph{arXiv preprint arXiv:2304.08247}, 2023.

\bibitem{mellama}
Q.~Xie, Q.~Chen, A.~Chen, C.~Peng, Y.~Hu, F.~Lin, X.~Peng, J.~Huang, J.~Zhang, V.~Keloth \emph{et~al.}, ``Me llama: Foundation large language models for medical applications,'' \emph{arXiv preprint arXiv:2402.12749}, 2024.

\bibitem{med-palm2}
K.~Singhal, T.~Tu, J.~Gottweis, R.~Sayres, E.~Wulczyn, M.~Amin, L.~Hou, K.~Clark, S.~R. Pfohl, H.~Cole-Lewis \emph{et~al.}, ``Toward expert-level medical question answering with large language models,'' \emph{Nature Medicine}, pp. 1--8, 2025.

\bibitem{palm2}
R.~Anil, A.~M. Dai, O.~Firat, M.~Johnson, D.~Lepikhin, A.~Passos, S.~Shakeri, E.~Taropa, P.~Bailey, Z.~Chen \emph{et~al.}, ``Palm 2 technical report,'' \emph{arXiv preprint arXiv:2305.10403}, 2023.

\bibitem{clinicalcammel}
A.~Toma, P.~R. Lawler, J.~Ba, R.~G. Krishnan, B.~B. Rubin, and B.~Wang, ``Clinical camel: An open-source expert-level medical language model with dialogue-based knowledge encoding,'' \emph{CoRR}, 2023.

\bibitem{bloomz}
N.~Muennighoff, T.~Wang, L.~Sutawika, A.~Roberts, S.~Biderman, T.~L. Scao, M.~S. Bari, S.~Shen, Z.-X. Yong, H.~Schoelkopf \emph{et~al.}, ``Crosslingual generalization through multitask finetuning,'' \emph{arXiv preprint arXiv:2211.01786}, 2022.

\bibitem{gatortrongpt}
C.~Peng, X.~Yang, A.~Chen, K.~E. Smith, N.~PourNejatian, A.~B. Costa, C.~Martin, M.~G. Flores, Y.~Zhang, T.~Magoc \emph{et~al.}, ``A study of generative large language model for medical research and healthcare,'' \emph{NPJ digital medicine}, vol.~6, no.~1, p. 210, 2023.

\bibitem{clinicalgpt}
G.~Wang, G.~Yang, Z.~Du, L.~Fan, and X.~Li, ``Clinicalgpt: large language models finetuned with diverse medical data and comprehensive evaluation,'' \emph{arXiv preprint arXiv:2306.09968}, 2023.

\bibitem{bloom}
T.~Le~Scao, A.~Fan, C.~Akiki, E.~Pavlick, S.~Ili{\'c}, D.~Hesslow, R.~Castagn{\'e}, A.~S. Luccioni, F.~Yvon, M.~Gall{\'e} \emph{et~al.}, ``Bloom: A 176b-parameter open-access multilingual language model,'' 2023.

\bibitem{zhongjing}
S.~Yang, H.~Zhao, S.~Zhu, G.~Zhou, H.~Xu, Y.~Jia, and H.~Zan, ``Zhongjing: Enhancing the chinese medical capabilities of large language model through expert feedback and real-world multi-turn dialogue,'' in \emph{Proceedings of the AAAI Conference on Artificial Intelligence}, vol.~38, no.~17, 2024, pp. 19\,368--19\,376.

\bibitem{cpllm}
O.~Ben~Shoham and N.~Rappoport, ``Cpllm: Clinical prediction with large language models,'' \emph{PLOS Digital Health}, vol.~3, no.~12, p. e0000680, 2024.

\bibitem{chatcounselor}
J.~M. Liu, D.~Li, H.~Cao, T.~Ren, Z.~Liao, and J.~Wu, ``Chatcounselor: A large language models for mental health support,'' \emph{arXiv preprint arXiv:2309.15461}, 2023.

\bibitem{qwen}
J.~Bai, S.~Bai, Y.~Chu, Z.~Cui, K.~Dang, X.~Deng, Y.~Fan, W.~Ge, Y.~Han, F.~Huang \emph{et~al.}, ``Qwen technical report,'' \emph{arXiv preprint arXiv:2309.16609}, 2023.

\bibitem{qilinmed}
Q.~Ye, J.~Liu, D.~Chong, P.~Zhou, Y.~Hua, F.~Liu, M.~Cao, Z.~Wang, X.~Cheng, Z.~Lei \emph{et~al.}, ``Qilin-med: Multi-stage knowledge injection advanced medical large language model,'' \emph{arXiv preprint arXiv:2310.09089}, 2023.

\bibitem{baichuan}
B.~Inc., ``Baichuan-7b: A large-scale pretrained language model,'' \url{https://github.com/baichuan-inc/Baichuan-7B}, 2023, accessed: 2025-04-13.

\bibitem{alpacare}
X.~Zhang, C.~Tian, X.~Yang, L.~Chen, Z.~Li, and L.~R. Petzold, ``Alpacare: Instruction-tuned large language models for medical application,'' \emph{arXiv preprint arXiv:2310.14558}, 2023.

\bibitem{tcmgpt}
G.~Yang, X.~Liu, J.~Shi, Z.~Wang, and G.~Wang, ``Tcm-gpt: Efficient pre-training of large language models for domain adaptation in traditional chinese medicine,'' \emph{Computer Methods and Programs in Biomedicine Update}, p. 100158, 2024.

\bibitem{pediatricsgpt}
D.~Yang, J.~Wei, D.~Xiao, S.~Wang, T.~Wu, G.~Li, M.~Li, S.~Wang, J.~Chen, Y.~Jiang \emph{et~al.}, ``Pediatricsgpt: Large language models as chinese medical assistants for pediatric applications,'' \emph{arXiv preprint arXiv:2405.19266}, 2024.

\bibitem{baichuan2}
A.~Yang, B.~Xiao, B.~Wang, B.~Zhang, C.~Bian, C.~Yin, C.~Lv, D.~Pan, D.~Wang, D.~Yan \emph{et~al.}, ``Baichuan 2: Open large-scale language models,'' \emph{arXiv preprint arXiv:2309.10305}, 2023.

\bibitem{huatuogpt2}
J.~Chen, X.~Wang, K.~Ji, A.~Gao, F.~Jiang, S.~Chen, H.~Zhang, D.~Song, and W.~o. Xie, ``Huatuogpt-ii, one-stage training for medical adaption of llms,'' \emph{arXiv preprint arXiv:2311.09774}, 2023.

\bibitem{doctorglm}
H.~Xiong, S.~Wang, Y.~Zhu, Z.~Zhao, Y.~Liu, L.~Huang, Q.~Wang, and D.~Shen, ``Doctorglm: Fine-tuning your chinese doctor is not a herculean task,'' \emph{arXiv preprint arXiv:2304.01097}, 2023.

\bibitem{meditron}
Z.~Chen, A.~H. Cano, A.~Romanou, A.~Bonnet, K.~Matoba, F.~Salvi, M.~Pagliardini, S.~Fan, A.~K{\"o}pf, A.~Mohtashami \emph{et~al.}, ``Meditron-70b: Scaling medical pretraining for large language models,'' \emph{arXiv preprint arXiv:2311.16079}, 2023.

\bibitem{bianque}
Y.~Chen, Z.~Wang, X.~Xing, Z.~Xu, K.~Fang, J.~Wang, S.~Li, J.~Wu, Q.~Liu, X.~Xu \emph{et~al.}, ``Bianque: Balancing the questioning and suggestion ability of health llms with multi-turn health conversations polished by chatgpt,'' \emph{arXiv preprint arXiv:2310.15896}, 2023.

\bibitem{biomistral}
Y.~Labrak, A.~Bazoge, E.~Morin, P.-A. Gourraud, M.~Rouvier, and R.~Dufour, ``Biomistral: A collection of open-source pretrained large language models for medical domains,'' \emph{arXiv preprint arXiv:2402.10373}, 2024.

\bibitem{mistral}
A.~Q. Jiang, A.~Sablayrolles, A.~Mensch, C.~Bamford, D.~S. Chaplot, D.~de~Las~Casas, F.~Bressand, G.~Lengyel, G.~Lample, L.~Saulnier, L.~R. Lavaud, M.-A. Lachaux, P.~Stock, T.~Le~Scao, T.~Lavril, T.~Wang, T.~Lacroix, and W.~El~Sayed, ``Mistral: A powerful dense decoder for open-weight language models,'' \emph{arXiv preprint arXiv:2310.06825}, 2023.

\bibitem{aquilamed}
L.~Zhao, W.~Zeng, X.~Shi, H.~Zhou, D.~Hao, and Y.~Lin, ``Aqulia-med llm: Pioneering full-process open-source medical language models,'' \emph{arXiv preprint arXiv:2406.12182}, 2024.

\bibitem{iimedgpt}
Y.~Zhang, Z.~Chang, W.~Cai, M.~Ren, K.~Yuan, Y.~Sun, and Z.~Ding, ``Iimedgpt: Promoting large language model capabilities of medical tasks by efficient human preference alignment,'' \emph{arXiv preprint arXiv:2501.02869}, 2025.

\end{thebibliography}
}
\twocolumn[
\appendix
\vspace{0.4cm} 
\section*{A. Preliminaries}
\vspace{6pt} 
]

\begin{figure}[htbp]  
    \centering

    \begin{minipage}{\textwidth} 
        \subsection*{Pre-train and Fine-tune}

        \subsubsection*{Pre-training}
        \setlength{\parindent}{2em}       
        Model training is a critical phase that enables the model to effectively handle specific tasks. Under the pre-training and fine-tuning learning paradigm, pre-training represents the first step in the LLM training process. During the pre-training phase, the model implements various pre-training tasks utilizing a large amount of unlabeled text data, such as predicting or generating the next word or sentence. This procedure endows the model with the ability to comprehend and generate natural language initially.    
    \vspace{0.8cm}
    \end{minipage}
    \begin{minipage}{\textwidth}
        \centering
        \includegraphics[width=0.8\textwidth, keepaspectratio]{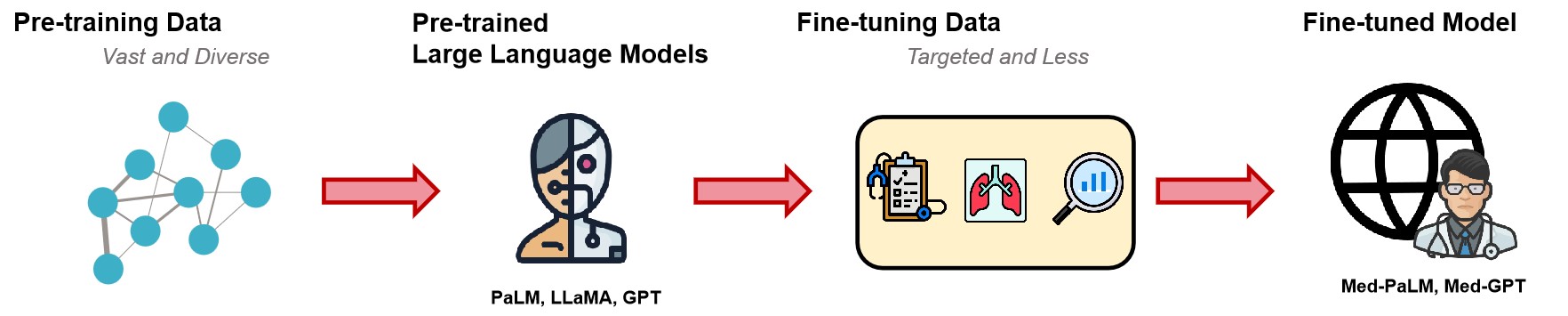}
        \caption{\footnotesize The evolution of large pre-trained language models through fine-tuning to domain-specific models.}
        \label{fig:finetune}
    \end{minipage}
\end{figure}
\begin{figure}[htbp]  
    \centering
    \begin{minipage}{\textwidth}  
\vspace{0.2cm} 
        \subsubsection*{Fine-tuning}
        \vspace{0.1cm}
        \setlength{\parindent}{2em}
        After finishing the pre-training phase, although the model possesses the ability to comprehend and generate natural language, it still does not completely meet the specific requirements of various tasks. By utilizing labeled data for supervised fine-tuning, these models can acquire the capability to handle particular downstream tasks \cite{downt1}\cite{downt2}\cite{downt3}. Fine-tuning is not limited to a single task; rather, it can be effectively performed across multiple tasks \cite{mfttasks1}\cite{mfttasks2}.    

        By engaging in multi-task training and fine-tuning interactions, the applicability of the model is broadened, thereby enhancing its flexibility. By fine-tuning on domain-specific datasets, large language models (LLMs) have demonstrated versatility across various domains, as illustrated in Figure~\ref{fig:finetune}. In the field of medical healthcare, LLMs have multiple applications, such as medical searching\cite{msearch1} \cite{msearch2} and medical assistance\cite{massist1} \cite{massist2}. By being fine-tuned appropriately, some LLMs can even pass the medical licensing examination\cite{massist2}\cite{mexam1}. These examples demonstrate that LLMs possess valuable potential in the medical healthcare domain.

        Though breakthroughs have been made by LLMs in the field of medical healthcare, the type of data utilized remains limited to a single type. The restriction hinders the development of deep learning applications in medical healthcare, as such settings typically require processing various types of data, including images, omics, and audio.
   \vspace{0.26cm}      
        \subsubsection*{Transformer}
The core of the Transformer is the self-attention mechanism, which dynamically assigns varying weights to different parts of the input, enhancing the model’s ability to capture contextual relationships. Additionally, positional encoding is employed to integrate the relative or absolute positional information of tokens within the sequence, preserving the order of the sequence, which is crucial for understanding the flow of the text.
The Transformer architecture is primarily composed of two critical components: the encoder and the decoder. \cite{1} mentions that each encoder layer consists of a stack of six identical layers, where one sub-layer is a multi-head self-attention mechanism, and the other is a position-wise fully connected feed-forward network. Similarly, the decoder also possesses a stack of six identical layers. Additionally, the decoder includes a third sub-layer designed to perform multi-head attention over the encoder stack's outputs.

An attention function can be considered a mapping that utilizes a set of queries along with corresponding sets of keys and values to generate outputs, all of which are represented as vectors. The outputs are computed as a weighted sum of the values, with the weights determined by a compatibility function that assesses the relationship between each query and its 
    \end{minipage}
\end{figure}

\clearpage
\begin{figure}[htbp]  
    \vspace{0.3cm} % 增加垂直间距
    \begin{minipage}{\textwidth}
        \subsection*{Pre-training tasks} 
        \setlength{\parindent}{2em}
        Currently, mainstream pre-training tasks in MLLMs include a combination of Masked Language Modeling (MLM) and Masked Image Modeling (MIM) \cite{visualbert,flava,vilbert}, and Image-Text Matching (ITM) \cite{7,8}. Figure~\ref{fig:pretraintask1} illustrates the schematic diagram of the two main pre-training tasks.
        
        \setlength{\parskip}{1em}  
        \subsubsection*{ITC} 
        \setlength{\parindent}{2em}
        The ITC task encourages the learning of meaningful representations through the comparison of positive and negative pairs of visual and textual data, which is widely applied in pre-training tasks for vision-language models. The ITC task trains the model by minimizing the distance between matching image-text pairs while maximizing the distance between non-matching pairs. A classic example is the CLIP model, which employs cosine distance as a metric to assess similarity between text and image embeddings. It utilizes a contrastive loss function to jointly train both text encoders and image encoders, thereby bridging the two modalities of vision and language.
    \end{minipage}
    \vspace{20pt}  % 增加垂直间距
    \begin{minipage}{\textwidth}
        \centering
        \vspace{0.8cm}
        \includegraphics[width=0.9\columnwidth]{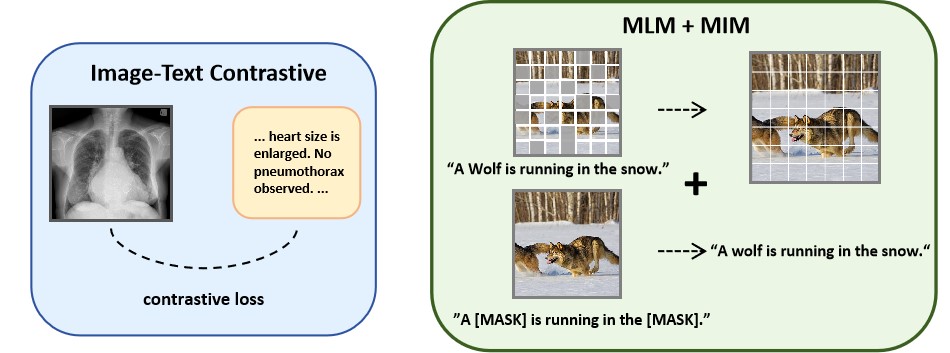}  % 使用 0.9\columnwidth 来确保图像在单栏内
        \vspace{15pt}
        \caption{\footnotesize Illustration of common pretraining tasks: Image-Text Matching (ITM) for cross-modal alignment, and MIM+MLM for understanding masked image and text features.}
        \label{fig:pretraintask1}
    \end{minipage}
    \vspace{60pt}  % 增加垂直间距
    \begin{minipage}{\textwidth}
        \setlength{\parskip}{1em}  
        \subsubsection*{MLM + MIM} 
        \setlength{\parindent}{2em}
        The combination of Masked Language Modeling (MLM) and Masked Image Modeling (MIM) enables the learning of potential connections between language and visual data by virtue of reconstructing hidden linguistic tokens through the integration of linguistic knowledge and visual cues. MLM is one of the widely used tasks in natural language processing (NLP) pre-training that involves randomly selecting a certain proportion of tokens from text data and replacing them with a special token, typically denoted as MASK. The model predicts these masked tokens by considering the context on both sides, thereby allowing it to grasp detailed contextual information \cite{2}. Inspired by MLM and the introduction of Vision Transformers (ViT), some researchers have begun to extend the methods of MLM to images—this concept is now referred to as MIM \cite{beit,autopencoders}. Similar to MLM, MIM masks portions of image patches for prediction purposes, enabling the model to comprehend image information effectively. By combining Masked Language Modeling (MLM) with Masked Image Modeling (MIM), where partial information is masked in one modality while utilizing another modality for reconstruction, models can better understand and relate complex relationships between images and text, thereby enhancing performance in multimodal tasks \cite{masked}.
    \end{minipage}
\end{figure}

\clearpage
\begin{figure}[htbp]  
    \vspace{0.8cm} % 增加垂直间距
    \begin{minipage}{\textwidth}
    \subsection*{MLLMs Main Components}
    \vspace{20pt}
    \subsubsection*{LLM}
    \setlength{\parindent}{2em}
      MLLM generally utilizes the LLM, which has been pre-trained and fine-tuned, as the core of the model. LLM receives the other aligned modality’s input and uses its capability of comprehension, inference, and generation to produce text outputs or command tokens within the textual feature space. These command tokens respond to prompts provided by users, and guide other components in executing complex cross-modal tasks. LLaMA, T5 \cite{t5}, Vicuna, Palm are commonly utilized as foundational models for large language models.
    \vspace{0.3cm}     
    \subsubsection*{Image Encoder}
    \setlength{\parindent}{2em}
    Image encoder, as the encoder for the majority of the modality, has seen mature development. The commonly used models include ViT\cite{vit} and its variants, as well as MAE\cite{mae} and Swin Transformer\cite{swin}. Among them, ViT is widely utilized for image feature extraction by employing a series of transformer blocks. Each block is composed of multi-head self-attention layers and feedforward networks, allowing ViT to focus on different image features. The specific feature that is concentrated depends on the specific training tasks. For instance, the pre-trained ViT used in CLIP is typically applied for general image understanding, whereas the pre-trained ViT employed in BuboGPT \cite{bubo}, particularly within SAM \cite{sam}, is more suited for detailed and fine-grained image analysis.

    \vspace{0.3cm}          
    \subsubsection*{Video Encoder}
    \setlength{\parindent}{2em}
    LMMs typically adapt the image encoder as a video encoder, applying downsampling preprocessing to extract representative frames. Subsequently, the processing is the same as that of static images, which preserves the core visual information of the video and reduces the computational requirements \cite{videoec}.

    \vspace{0.3cm}     
    \subsubsection*{Audio Encoder}
    \setlength{\parindent}{2em}
    Common audio encoders include transformer-based C-Former\cite{cformer}, Whisper\cite{whisper}, and HuBERT\cite{hubert}. A common approach to audio feature extraction is to progressively leverage convolutional layers and transformers. Generally, there are two primary methods for encoding audio signals. One method directly encodes the audio waveform, while the other transforms the audio signal into a spectrogram before encoding.Specifically, \cite{hubert} utilizes a convolutional waveform encoder to directly extract features from the audio waveform. After this initial processing, the features are further processed by a transformer that employs BERT-style training techniques, such as masked prediction, to obtain the higher-level features. On the other hand, \cite{whisper} computes the Mel spectrogram representation of the audio signal, then, processes these features through convolutional layers and a transformer to meet the requirements for multilingual and multitask audio processing.

    \vspace{0.3cm}     
    \subsubsection*{Alignment Module}
    \setlength{\parindent}{2em}
    Alignment module is designed to align feature vectors from different modals by transforming the feature vectors generated by encoders of different modes into LLM-compatible feature representations. For example, LLaVA\cite{llava} uses the multi-layer perceptron (MLP) \cite{mlp} to map the image visual features extracted with ViT to the word embedding space of the language model. In addition, a common and feasible solution is to extract information in a query-based manner using a couple of learnable queries, which was first implemented in Flamingo and BLIP-2. Q-Former in BLIP-2, a lightweight Transformer, extracts image feature information suitable for LLM understanding from the visual encoder by generating queries. Also, the research has found that expanding the parameters of the alignment module can make the information synthesis between different modules more complete. QLLaMA\cite{qllama}, which is regarded as a larger version of QFormer, was developed on the basis of multilingual LLaMA, adding 96 learnable queries and randomly initialized cross-attention. With this large parameter alignment module, InternVL can achieve satisfactory performance on multimodal dialogue tasks even with frozen LLM decoders.

    \end{minipage}
   
\end{figure}

\twocolumn[ 
\vspace{0.4cm} 
\section*{B. Interpretability-enhancing optimization}
\vspace{6pt} 
]  

\begin{figure}[htbp]  
    \begin{minipage}{\textwidth} 
        \setlength{\parindent}{2em}
         \renewcommand{\baselinestretch}{7.5}
        \vspace{0.3cm}     
      Some rate mechanisms have been published to evaluate text generation from medical MLLMs. For example, \cite{gptformat} shown in Figure 1, has designed an evaluation mechanism that rates answers based on three aspects: key points, inference process, and evidence, using GPT-4O. The criticism is provided by human doctors. Specifically, GPT-4O checks whether the answers generated by the models provide evidence supporting the conclusion or diagnosis, whether the models follow the correct inference process, and whether they ignore key content.

        \vspace{0.7cm}
        \centering
        \includegraphics[width=0.8\textwidth, keepaspectratio]{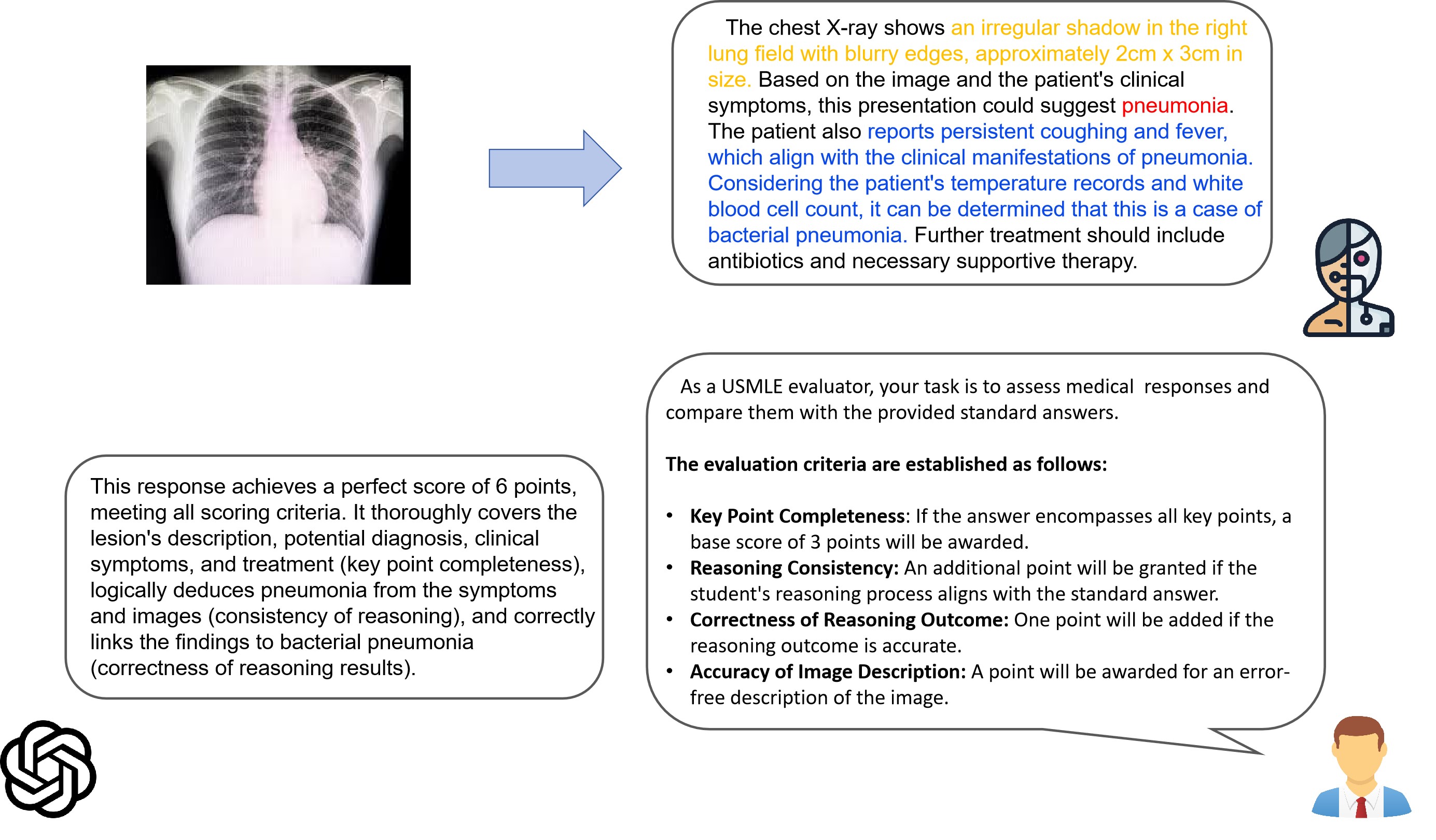}
        \caption{\footnotesize Utilizing GPT-4o in the role of a doctor to evaluate the diagnostic content of medical MLLMs.}
        \label{fig:evaluate}
        \vspace{-0.4cm}
    \end{minipage}
\end{figure}

\twocolumn[
\vspace{0.7cm}
\noindent
\begin{minipage}{\textwidth}
    \section*{C. Medical Multimodal Large Language Model Training Data}
    \vspace{1cm}
    
    \begin{center}
    \renewcommand{\arraystretch}{1.2}
    \captionof{table}{Medical Multimodal Large Language Model Training Data.}
    \label{tab:mllmtrainingdata1}
    \resizebox{\textwidth}{!}{%
    \begin{tabular}{p{1.5cm} p{3cm} p{10cm} p{4cm}}
        \toprule
        \textbf{Year} & \textbf{Dataset} & \textbf{Description} & \textbf{Type} \\
        \midrule
    2015 & CheXpert\cite{chexpert} & A dataset includes labeled chest X-ray images, which are utilized for disease classification and detection tasks. & Image-Report \\
2017 & EndoVis 2017\cite{endovis} & A dataset of endoscopic surgical images and videos is designed for the analysis of surgical scenarios. & Image-Caption \\
2018 & EndoVis 2018\cite{endovis} & A dataset for the segmentation and tracking tasks of endoscopic images. & Image-Caption \\
2018 & VQA-RAD\cite{vqarad} & A dataset comprises radiological images accompanied by relevant question-and-answer pairs for visual question answering. & QA \\
2019 & MIMIC-CXR-JPG\cite{mimiccxrjpg} & A dataset includes chest X-ray images in JPEG format from the MIMIC database. & Image-Report \\
2019 & SwissProtCLAP\cite{swissprotclap} & A dataset comprises biomedical literature and data related to protein functions. & Scientific Literature \\
2020 & MediCaT\cite{medicat} & A multimodal dataset comprising medical images and their corresponding captions. & Image-Caption \\
2020 & PathVQA\cite{pathvqa} & A dataset specifically designed for visual question answering tasks in the field of pathology images. & QA \\
2020 & SAT-DS\cite{satds} & A The dataset provides question-and-answer and testing tasks in the field of medicine. & QA \\
2020 & MedTrinity-25M\cite{medtrinity} & A large-scale medical image caption dataset & Image-Caption \\
2020 & CheXInstruct\cite{chexinstruct} & A directive for the chest X-ray data follows the task dataset, facilitating image comprehension. & Instruction-Following Data \\
2021 & SLAKE\cite{slake} & A multimodal dataset designed for medical visual question answering supports interaction between images and text. & QA \\
2021 & MIMIC-NLE\cite{mimicnle} & A dataset for natural language explanations of medical imaging. & Image-Report \\
2022 & ChiMed-VL-Instruction \cite{chimed} & Chinese Medical Visual-Language Instruction Following Task Dataset. & Instruction-Following Data \\
2022 & CXR-PRO\cite{cxrpro} & A professional-grade report and image pairing dataset of chest X-ray images. & Image-Report \\
2022 & PubMedVision-Alignment\cite{pubmedvisionalignment} & A dataset for the task of aligning medical visuals with text from PubMed. & Image-Caption \\
2023 & PMC-VQA\cite{pmcvqa} & A dataset comprises medical images and corresponding question-and-answer pairs specifically designed for visual question answering tasks. & QA \\
2023 & MedMD\cite{medmd} & A dataset for matching medical images with their corresponding descriptions supports image annotation tasks. & Image-Caption \\
2023 & PathCap\cite{pathcap} & A multimodal dataset of pathological images and text annotations. & Image-Caption \\
2023 & Quilt-1M\cite{quilt1m} & A large-scale dataset for the annotation of medical images through visual and textual pairing. & Image-Caption \\
2023 & OpenPath\cite{openpath} & The provided medical pathology images and labels are suitable for image annotation and classification tasks. & Image-Caption \\
2023 & PMC-15M\cite{pmc15m} & A dataset comprises a vast collection of medical images and associated textual data sourced from PMC. & Image-Caption \\
2023 & MS-CXR\cite{mscxr} & Specifically designed for the analysis of chest X-ray images. & Image-Report \\
2023 & IU-Xray or Open-I\cite{iu} & A dataset of medical X-ray images and reports is intended for the analysis of medical imaging. & Image-Report \\
2024 & ROCO\cite{roco} & A multimodal dataset pairing medical images with their corresponding descriptions. & Image-Caption \\
2024 & LLaVA-Med-Alignment\cite{llavamedalignment} & A dataset for aligning medical visuals and language supports the training of multimodal models. & Image-Caption \\

        \bottomrule
    \end{tabular}}
    \end{center}
\end{minipage}
\vspace{0.2cm}
]

\twocolumn[
\vspace{0.4cm}
\noindent
\begin{minipage}{\textwidth}  
    \begin{center}
    \renewcommand{\arraystretch}{1.2}
    \label{tab:mllmtrainingdata2}
    \resizebox{\textwidth}{!}{%
    \begin{tabular}{p{1.5cm} p{3cm} p{10cm} p{4cm}}
        \toprule
        \textbf{Year} & \textbf{Dataset} & \textbf{Description} & \textbf{Type} \\
        \midrule
2024 & ChiMed-VL-Alignment\cite{chimedvl} & Chinese Medical Visual-Language Alignment Dataset. & Image-Caption \\
2024 & ApolloCorpora\cite{apollo} & A multimodal medical dataset that integrates text and images. & Hybrid \\
2024 & M3D-Data\cite{m3ddata} & A medical 3D image dataset supports three-dimensional medical imaging analysis tasks. & Hybrid \\
2024 & MIMIC-CXR\cite{mimiccxr} & A medical imaging dataset comprises chest X-ray images along with their corresponding reports. & Image-Report \\
\hspace*{6.5pt}- & TCGA\cite{nci_tcga} & A cancer genomics dataset encompasses genetic and clinical data from a variety of tumor samples. & Scientific Literature \\ 
        \bottomrule
    \end{tabular}}
    \end{center}

\vspace{4.5cm}
\noindent
\begin{minipage}{\textwidth}
    \section*{D. Medical Large Language Model Training Data}
    \vspace{6pt}
    
    \begin{center}
    \renewcommand{\arraystretch}{1.2}
    \captionof{table}{Medical Multimodal Large Language Model Training Data.}
    \label{tab:llmtrainingdata1}
    \resizebox{\textwidth}{!}{%
    \begin{tabular}{p{1.5cm} p{3cm} p{10cm} p{4cm}}
        \toprule
        \textbf{Year} & \textbf{Dataset} & \textbf{Description} & \textbf{Type} \\
        \midrule
    2002 & Medical Meadow \cite{medicalmeadow} & A dataset focused on medical Q\&A for natural language processing tasks in healthcare. & QA \\
    2015 & CPRD \cite{cprd} & Source from the UK primary care clinical practice research database, covering patient treatment data. & EHR \\
    2016 & MIMIC-III \cite{mimicthree} & Health data from over forty thousand critical care patients, including detailed clinical information. & EHR \\
    2018 & cMedQA2 \cite{cmedqa2} & An community medical Q\&A dataset, featuring numerous medical-related questions and answers & QA \\
    2019 & PubMed Causal \cite{pubmedcausal} & Medical text dataset focused on causal reasoning, sourced from PubMed. & Dialogue \\
    2019 & MeQSum \cite{meqsum} & A dataset of medical problem summaries for training and evaluating automatic summary generation models. & Medical Question Summarization \\
    2019 & MedMentions \cite{medmentions} & A resource of medically-annotated scientific literature for training NLP models to identify and categorize medical concepts from text. & Scientific Literature \\
    2019 & MedQuAD \cite{medquad} & A dataset includes pairs of medical questions and answers extracted from the NIH website, covering various medical topics suitable for Q\&A tasks. & QA \\
    2019 & CMeKG \cite{cmekg} & A Chinese medical knowledge graph containing rich medical entities and relationships, designed to support medical information retrieval and Q\&A systems. & Knowledge Base \\
    2019 & CMeKG \cite{cmekg} & A Chinese medical knowledge graph containing rich medical entities and relationships, designed to support medical information retrieval and Q\&A systems. & Knowledge Base \\
    2019 & webMedQA \cite{webmedqa} & A dataset for question answering, derived from online medical Q\&A. & QA \\
    2019 & PubMedQA \cite{pubmedqa} & A dataset consisting of question and answer pairs derived from PubMed abstracts, focused on biomedical research & QA \\
    2020 & MEDIQA \cite{mediqa} & A medical Q\&A and text reasoning dataset focused on natural language understanding tasks in the medical field. & Dialogue \\
  2020 & The Pile \cite{thepile} & A large text dataset containing diverse textual data from various sources, including biomedical literature. & Scientific Literature \\

           \bottomrule
    \end{tabular}}
    \end{center}
\end{minipage}
\vspace{0.2cm}

\end{minipage}
\vspace{0.2cm}
]

\twocolumn[
\vspace{0.4cm}
\noindent
\begin{minipage}{\textwidth}
    \vspace{6pt}
    
    \begin{center}
    \renewcommand{\arraystretch}{1.2}
    \label{tab:llmtrainingdata12}
    \resizebox{\textwidth}{!}{%
    \begin{tabular}{p{1.5cm} p{3cm} p{10cm} p{4cm}}
        \toprule
        \textbf{Year} & \textbf{Dataset} & \textbf{Description} & \textbf{Type} \\
        \midrule
      2020 & COMETA \cite{cometa} & A medical entity annotation dataset collected from social media, aimed at studying the representation of medical concepts in informal text. & Web Data (social media) \\
 2020 & MedDialog \cite{meddialog} & A dataset includes multi-turn medical dialogues between doctors and patients, suitable for training and evaluating medical dialogue systems. & Dialogue \\
    2020 & CORD-19 \cite{cord19} & An open dataset focused on COVID-19 and coronavirus research, containing a wealth of related academic articles and literature. & Scientific Literature \\  
   
    2021 & MIMIC-IV \cite{mimic4} & The extended version of MIMIC-III includes updated clinical data for ICU patients. & EHR \\
    2021 & RCT \cite{rct} & A dataset of randomized controlled trials for medical research and evidence-based practice. & Scientific Literature \\

    2021 & MS\^{}2 \cite{ms2} & A dataset of multi-document summaries in medical research for training and evaluating models that generate medical literature abstracts. & Scientific Literature \\
    2021 & CDSR \cite{cdsr} & A database of systematic reviews providing high-quality medical evidence to support clinical decisions. & Scientific Literature \\
    2021 & SumPubMed \cite{sumpudmed} & A dataset of PubMed abstracts for research on automatic summarization of biomedical literature. & Scientific Literature \\
    2021 & MedQA-USMLE \cite{medqausmile} & A dataset for question answering based on typical questions from the United States Medical Licensing Examination (USMLE). & QA \\
    2022 & S2ORC \cite{s2orc} & An open research corpus containing millions of academic papers across various disciplines, including medicine. & Scientific Literature \\
    2022 & CHQ-Summ \cite{chqsumm} & A summary dataset of children's health issues, featuring questions from parents on online forums and summarized responses from professional doctors. & Medical Question Summarization \\
    2022 & MedMCQA \cite{medmcqa} & A multiple-choice medical Q\&A dataset covering medical knowledge assessment tasks. & QA \\
    2023 & Psych8k \cite{psych8k} & A Q\&A dataset for mental health, featuring questions and expert responses. & QA \\
  2023 & Huatuo-26M \cite{huatuo26m} & Chinese medical Q\&A dataset, featuring a large collection of medical questions and answers. & QA \\
    2023 & MedInstruct-52k \cite{medinstruct52k} & A dataset of 52,000 medical instructions for training and evaluating adherence models in the medical field. & Instruction-Following Data \\
    2023 & GAP-Replay \cite{gapreplay} & Integrating clinical and non-clinical data for generalized medical natural language understanding tasks. & Hybrid \\
    2023 & Alpaca-EN-AN \cite{alpaca} & A dataset containing English instructions for instruction-following tasks. & English Instructions \\
    2024 & CMtMedQA \cite{cmtmedqa} & A large-scale Chinese medical multi-turn dialogue dataset suitable for language model training, covering 14 medical departments and various scenarios, featuring numerous proactive questions from doctors. & Dialogue \\
    2024 & TCM-Corpus-1B \cite{tcmcorpus1b} & A 1 billion-word corpus of traditional Chinese medicine to support natural language processing research in the field. & Web Data \\
    \hspace*{6.5pt}- & PubMed \cite{pubmed2024} & Providing abstracts and citation data for biomedical literature is a widely used resource in biomedical information. & Scientific Literature \\
    \hspace*{6.5pt}- & PMC \cite{pmc} & The full-text dataset of PubMed Central compiles open-access journal literature in biomedical and life sciences. & Scientific Literature \\

           \bottomrule
    \end{tabular}}
    \end{center}
\end{minipage}
\vspace{0.2cm}
]

\begin{table*}
\vspace{0.5cm}
    \section*{E. Medical Large Language Model}
    \centering
    \vspace{1cm}
    \caption{Details of Medical Large Language Models.}
    \label{tab:largelanguagemodel1}
    \resizebox{\linewidth}{!}{%
    \begin{tabular}{p{1.5cm} p{3cm} p{2cm} p{11cm}}
    \toprule
    \textbf{Year} & \textbf{Model} & \textbf{Base Model} & \textbf{Description} \\
    \midrule
        2022/12 & Med-PaLM\cite{medpalm}& PaLM\cite{palm}&Posses exceptional performance particularly in areas such as clinical note summarization, radiological image analysis, and drug interaction prediction.  \\\\
        2023/03 & ChatDoctor\cite{chatdoctor}& LLaMA\cite{5}& Refine LLaMA by utilizing a comprehensive dataset comprising 100,000 doctor-patient dialogues obtained from widely used online medical consultation platforms.\\\\
        2023/04 & PMC-LLaMA\cite{pmcllama}& LLaMA\cite{5}&By integrating 4.8 million biomedical academic papers and 30,000 medical textbooks for data-centric knowledge injection into LLAMA.\\\\
        2023/04 & Baize-Healthcare\cite{baize}& LLaMA\cite{5}&Leveraging 100,000 dialogue segments produced through self-dialogue by ChatGPT, which is specifically tailored to improve performance in medical-related conversations. \\\\
        2023/04 & HuaTuo\cite{huatuollm}& LLaMA\cite{5}&Based on LLaMA ,  specifically optimized and finely tuned for traditional Chinese medicine knowledge.Has been trained using a medical knowledge graph that includes entries related to traditional Chinese medicine and relevant medical literature data. \\\\
        2023/04 & MedAlpaca\cite{medalpaca} & LLaMA\cite{5}&Enhance performance in healthcare-related question-and-answer tasks by optimizing the training process on a diverse array of medical texts including medical flashcards, Wikipedia entries, and dialogue datasets, based on the LLaMA architecture. \\\\
        2023/04 & Me-LLaMA\cite{mellama} & LLaMA2\cite{llama2} &Optimized on the LLaMA architecture through pre-training and instruction fine-tuning for medical, clinical, and general data to excel in tasks like question answering, entity recognition, and text summarization. \\\\
        2023/05 & Med-PaLM 2\cite{med-palm2}& PaLM 2\cite{palm2} & Base the development of Med-PaLM 2 on the updated PaLM 2 architecture, and achieve improved accuracy on medical examination questions like those in the USMLE, surpassing the original Med-PaLM.\\\\
        2023/05 & Clinical Camel\cite{clinicalcammel}& LLaMA2\cite{llama2}&Fine-tune LLaMA-2 using QLoRA to implement the Dialogue-Based Knowledge Encoding (DBKE) strategy, transforming dense medical texts into dialogue data and enhancing its applicability in medical dialogue systems.\\\\
        2023/06 & HuatuoGPT\cite{huatuo}& BLOOMZ\cite{bloomz} &Integrate real physician data with refined datasets and optimize performance through reinforcement learning to enable the system to diagnose like a qualified doctor.\\\\
        2023/06 & GatorTronGPT\cite{gatortrongpt}& GPT-3\cite{gpt3}&Leverage a substantial corpus of approximately 277 billion mixed clinical and English words to train GatorTronGPT, which generates synthetic clinical texts closely resembling real clinical notes authored by medical professionals.\\\\
        2023/08 & ClinicalGPT\cite{clinicalgpt} & BLOOM\cite{bloom}& Integrate extensive real-world medical records with specialized knowledge to optimize ClinicalGPT's capabilities for handling multi-turn dialogue consultations, medical knowledge Q\&A, patient consultations, medical examinations, and diagnostic analysis of medical records.\\\\
        2023/08 & Zhongjing\cite{zhongjing}& Ziya-LLaMA\cite{5}& Utilize various forms of medical data for fine-tuning and construct the large-scale Chinese medical multi-turn question-answer dataset, CMtMedQA, to enhance conversational capabilities. \\\\
        2023/09 & Radiology-Llama2\cite{radiologyllama2} & LLaMA2\cite{llama2} &Optimized for generating coherent and clinically relevant radiological impressions from extensive datasets of radiology reports, enhanced through a specific instruction-tuning process to increase accuracy and practicality in clinical applications.\\\\
        % 2023/09 & MedChatZH\cite{medchatzh}& Baichuan\cite{baichu} &Continue pre-training on a corpus of traditional Chinese medicine literature using the LLaMA framework, followed by fine-tuning with a carefully curated dataset of medical guidance. \\\\
        2023/10 & CPLLM\cite{cpllm}& LLaMA2\cite{llama2} &Specifically optimized for predicting clinical diseases and readmissions by utilizing patients' historical diagnostic records to forecast whether a patient will be diagnosed with the target disease during their next visit or subsequent periods.\\\\
        2023/10 & ChatCounselor\cite{chatcounselor}& Vicuna\cite{4}&Designed for psychological counseling,  leverages  LLaMA-7B  and fine-tuned with a dataset from one-on-one interviews with 260 individuals, each lasting an hour and exclusively conducted by licensed psychological counselors, ensuring the model's proficiency and reliability in providing expert-level counseling guidance. \\\\       
        2023/10 & Apollo\cite{apollo}& Qwen\cite{qwen}& Optimized for precise multilingual medical tasks, this model encompasses multilingual healthcare information retrieval, case analysis, and clinical decision support.   It excels particularly in providing diagnostic recommendations, addressing symptom inquiries, and enhancing medical education and training.\\\\
        2023/11 & Qilin-Med\cite{qilinmed}& Baichuan\cite{baichuan} & A  Chinese medicine LLM employs a multi-stage training approach, utilizing medical question-answering, pure text, knowledge graphs, and dialogue data has demonstrated  exceptional performance in various medical domain tasks, including medical knowledge question-answering, case analysis, and understanding of medical texts.\\\\
 2023/11 & AlpaCare\cite{alpacare}& LLaMA\cite{5}&Fine-tune LLaMA by using the diverse, machine-generated medical IFT dataset, MedInstruct-52k, which integrates GPT-4 and ChatGPT while leveraging a high-quality, expert-curated seed subset. The model has demonstrated exceptional performance across various medical applications, free-form teaching assessments, and multiple general domain benchmark tests.\\\\      
    \bottomrule
    \end{tabular}}
    \vspace{-0.4cm}
\end{table*}

\begin{table*}
    \vspace{-1.4cm}
    \centering
    \label{tab:largelanguagemodel2}
    \resizebox{\linewidth}{!}{%
    \begin{tabular}{p{1.5cm} p{3cm} p{2cm} p{11cm}}
    \toprule
    \textbf{Year} & \textbf{Model} & \textbf{Base Model} & \textbf{Description} \\
    \midrule

 2023/11 & BioMedLM\cite{biomedlm}& Transformer & Trained on PubMed abstracts and full texts. After fine-tuning can generate robust biomedical question-answering results that are comparable to larger models, effectively addressing computational cost issues. Concentrates on medical genetics and multiple-choice biomedical question answering, achieving excellent scores in MedMCQA and MMLU medical genetics examinations.\\\\
        2024/01 & TCM-GPT\cite{tcmgpt}& BLOOM\cite{bloom}&The first large-scale language model of 7 billion parameters specifically pre-trained on Traditional Chinese Medicine (TCM) data has significantly enhanced the model's performance across various downstream tasks in the TCM domain,underscores the necessity and effectiveness of domain-specific pre-training.\\\\
        2024/01 & PediatricsGPT\cite{pediatricsgpt} & Baichuan 2\cite{baichuan2}& Built on a robust training process using systematic approaches, the first Chinese pediatric LLM assistant leverages data from pediatric textbooks, guidelines, and knowledge graphs for fine-tuning.  Specifically optimized for tasks such as pediatric disease diagnosis, treatment recommendations, and symptom inquiries, it outperforms previous Chinese medical LLMs.  This achievement underscores the significance and effectiveness of in-depth fine-tuning within specific medical domains.\\\\
        2024/02 & HuatuoGPT-II\cite{huatuogpt2}& Baichuan 2\cite{baichuan2}& Extensively using traditional Chinese medicine textual data for pre-training, which includes both classical literature and modern medical research, has led to outstanding performance in the latest round of the Chinese Medical Licensing Examination. This not only demonstrates effectiveness but also highlights generalization capabilities. \\\\
        2024/02 & DoctorGLM\cite{doctorglm}& ChatGLM-6B& Designed specifically for Chinese medical dialogue, the large language model is based on ChatGLM-6B Trained using a diverse range of Chinese medical dialogue data from various fields, including surgery, obstetrics and gynecology, pediatrics, and internal medicine, the model is optimized for bilingual environments in both Chinese and English.\\\\
        2023/11 & MEDITRON\cite{meditron} & LLaMA2\cite{llama2}& Built on Llama2 and fine-tuned using selected PubMed articles, abstracts, and internationally recognized medical guidelines, demonstrates performance surpassing GPT-3.5 and Med-PaLM. Approaching the effectiveness of GPT-4 and Med-PaLM-2 with only a slight gap, robust capabilities in the medical domain are showcased.\\\\
        2024/01 & AMIE\cite{amie}& PaLM 2\cite{palm2}& Employing an autoregressive architecture optimized for medical conversations and diagnostic tasks, the diagnostic dialogue system based on Palm2 integrates real-world data—including medical Q\&A, note summaries, and dialogue datasets—with simulated dialogue data through a multi-stage fine-tuning process. Diagnostic capabilities and clinical communication performance are further enhanced through reinforcement learning cycles.\\\\
        2024/03 & BianQue\cite{bianque}& ChatGLM\cite{chatglm}& Designed to enhance multi-turn inquiries and health advice in medical dialogues, the large language model based on Chat-6B incorporates a variety of data, including medical Q\&A instructions, drug information guidelines, medical encyclopedia knowledge directives, and Chat GPT distillation instructions.  Strengthened capabilities in providing recommendations and conducting knowledge queries are achieved through this comprehensive approach. \\\\
        2024/05 & BioMistral\cite{biomistral}& Mistral\cite{mistral}& Optimized based on the Mistral 7B Instruct model and primarily pre-trained on articles and abstracts from PubMed Central, the model employs multilingual optimization and lightweight techniques. Particularly suitable for multilingual medical question-answering tasks, it is designed for efficient operation in resource-constrained environments.\\\\
        2024/05 & SoulChat\cite{llmchatbot1}& ChatGLM\cite{chatglm} & Enhance empathy, listening, and comforting abilities in mental health dialogues through fine-tuning with a substantial dataset of multi-turn empathetic conversations, significantly improving human-like expression.\\\\
        2024/06 & Aqulia-Med \cite{aquilamed}& Aquila2\cite{aquila2}&Supporting bilingual communication in Chinese and English, the first medical large language model features a fully open-source training process. Utilizing a three-stage training workflow—Continued Pre-training (CPT), Supervised Fine-tuning (SFT), and Reinforcement Learning from Human Feedback (RLHF)—it demonstrates wide applicability in scenarios such as medical question answering, diagnostic support, and medical education.\\\\
        2025/01 & IIMedGPT\cite{iimedgpt}&Qwen-14B-base\cite{qwen}&Integrate medical instruction data with the Direct Preference Optimization (DPO) method, significantly enhancing performance in medical tasks. Broad applicability is demonstrated across scenarios such as medical question answering, clinical reasoning, and medical education. \\\\
    \bottomrule
    \end{tabular}}
    \vspace{-0.4cm}
\end{table*}

\end{document}